\documentclass{article}
\usepackage[table,xcdraw]{xcolor}
\usepackage{amssymb}
\usepackage{amsmath}
\usepackage[preprint]{corl_2026}
\usepackage{graphicx}
\usepackage{booktabs}
\usepackage{array}
\usepackage{adjustbox}
\usepackage{microtype}
\DeclareMathOperator{\argmax}{argmax}

\usepackage{tikz}
\usepackage{setspace}
\usetikzlibrary{decorations.text}
\usetikzlibrary{decorations.markings}
\usetikzlibrary{calc}
\usetikzlibrary{shapes}
\usetikzlibrary{positioning}
\usetikzlibrary{arrows}
\usetikzlibrary{shapes.arrows}
\usetikzlibrary{arrows.meta}
\usetikzlibrary{fit}
\usetikzlibrary{shapes.geometric}
\usetikzlibrary{backgrounds}

\newlength{\minimumsizeaicon}
\newlength{\minimumheightaicon}
\newlength{\minimumsizeinterconnection}
\newlength{\minimumheightinterconnection}

\newlength{\minimumsizegoal}

\definecolor{aicRed}{HTML}{e78284}   
\definecolor{aicGreen}{HTML}{a6d189} 
\definecolor{aicBlue}{HTML}{8caaee}  
\definecolor{aicOrange}{HTML}{e5c890}  
\definecolor{aicGray}{HTML}{dce0e8}

\newcommand{\figcell}[2]{%
  \adjustbox{valign=m}{\includegraphics[#1]{#2}}%
}

\newcommand{\panellabel}[1]{%
  \adjustbox{valign=m}{\small\textbf{(#1)}}%
}

\tikzset{
	aicon/state/.style={
		rectangle,
		fill=aicBlue,
		rounded corners=4pt,
		minimum width=\minimumsizeaicon,
		minimum height=\minimumheightaicon,
		inner sep=2pt,
		text=white,
		align=center,
		font=\linespread{0.8}\selectfont\scriptsize
	},
	aicon/active/.style={
		rectangle,
		rounded corners=4pt,
		fill=aicGreen,
		minimum width=\minimumsizeinterconnection,
		minimum height=\minimumheightinterconnection,
		inner sep=2pt,
		text=white,
		align=center,
		font=\linespread{0.8}\selectfont\scriptsize
	},
	aicon/sensor/.style={
		rectangle,
		rounded corners=4pt,
		fill=aicOrange,
		minimum width=\minimumsizeaicon,
		minimum height=\minimumheightaicon,
		inner sep=2pt,
		text=white,
		align=center,
		font=\linespread{0.8}\selectfont\scriptsize
	},
	aicon/action/.style={
		rectangle,
		rounded corners=4pt,
		fill=aicOrange,
		minimum width=\minimumsizeaicon,
		minimum height=\minimumheightaicon,
		inner sep=2pt,
		text=white,
		align=center,
		font=\linespread{0.8}\selectfont\scriptsize
	},
	aicon/goal/.style={
		diamond,
		fill=aicRed,
		minimum size=\minimumsizegoal,
		inner sep=2pt,
		text=white,
        align=center,
		font=\linespread{0.8}\selectfont\scriptsize
	},
	aicon/link/.style={
		draw=aicGray,
		line width=2.5pt,
	},
	aicon/gradient/.style={
		draw=aicRed,
		line width=2.5pt,
		->,
		>=latex
	}
}

\title{World--Task Factorization for Robot Learning}

\vspace{-0.3cm}
\author{
  Eduardo Sebasti\'an\\
  Department of Computer Science and Technology, University of Cambridge, United Kingdom\\
  \texttt{es2121@cam.ac.uk} \vspace{-0.3cm}\\
  \AND
  Adrian Pfisterer\\
  Robotics and Biology Laboratory, Technische Universität Berlin, Germany\\
  \texttt{adrian.pfisterer@tu-berlin.de} \vspace{-0.3cm}\\
  \And
  Vito Mengers\\
  Robotics and Biology Laboratory, Technische Universität Berlin\\
  Science of Intelligence (SCIoI), Cluster of Excellence, Berlin, Germany\\
  \texttt{v.mengers@tu-berlin.de} \vspace{-0.3cm}\\
  \And
  Oliver Brock\\
  Robotics and Biology Laboratory, Technische Universität Berlin\\
  Science of Intelligence (SCIoI), Cluster of Excellence, Berlin, Germany\\
  Robotics Institute Germany\\
  \texttt{oliver.brock@tu-berlin.de} \vspace{-0.3cm}\\
  \And
  Amanda Prorok \\
  Department of Computer Science and Technology, University of Cambridge, United Kingdom\\
  \texttt{asp45@cam.ac.uk} \\
}

\begin{document}
\maketitle
\vspace{-1cm}
\begin{abstract}
Robot learning must produce policies that generalize to new combinations of constraints, teammates, and environments. To achieve this, we must structurally factor the policy, which is a choice that ultimately dictates what generalizes, what requires retraining, and what remains entangled. Existing methods span a wide spectrum, from expecting structure to emerge from data scaling, to hand-designing it via hierarchies, skill libraries or learned specializations. In this paper, we study what we argue is the simplest, most fundamental factorization in robotics: separating the world from the task. Specifically, we investigate the conditions under which this factorization is principled. World factors, such as kinematics, sensing, and geometry of the environment, are properties of the embodied system and the environment; they exist independently of intent. Task factors, such as goal priorities, short-term trade-offs, and role choices, are defined by the task's logic over what the world admits. We formalize this asymmetry through Bayesian model evidence: a world/task factorization aligns with the data-generating process, maintains high likelihood through an analytical world model, and reduces the Occam razor's penalty on task parameters. We instantiate this factorization by pairing AICON, a differentiable graph of recursive estimators and interconnections that is itself compositional, operates without task-specific data, and propagates cost gradients to actuators, with a compact, learned policy that modulates gradient paths. Gradients serve as the interface between the two factors: they carry world structure through the graph and task structure through costs, enabling low-dimensional learning while preserving structural generalization. 
We empirically study the world/task factorization across three problems that encompass heterogeneous robot dynamics, environment classes, task logic and sensorimotor modalities.
Our framework outperforms end-to-end learned baselines and matches or exceeds analytical heuristics in all settings, generalizes zero-shot to out-of-distribution configurations, and transfers to real hardware without retraining. 
\end{abstract}


\section{Introduction}\label{sec:intro}

A long-standing aspiration in robotics is that a policy trained once should remain useful as embodiments, teammates, and task conditions vary. Realizing this means choosing how to structure the policy---committing to a factorization. The chosen factorization is consequential: it determines what generalizes ``for free'', what requires retraining, and what remains entangled. Existing approaches span a broad spectrum. At one end, structure is expected to emerge from data scaling, with foundation-model and vision-language-action approaches as the prominent instantiation~\cite{brohan2023rt2, intelligence2025pi_}. At the other end, structure is built in through inductive biases matched to the domain, with evidence accumulating that this produces more reliable generalization~\cite{battaglia2018relational}. This paper adds to that evidence by addressing a question we argue is more consequential than usually acknowledged: when structure is built in, what should be factored from what?

Different families of methods commit to different factorizations: hierarchical methods factor along temporal scale, skill libraries along primitives, modular networks along discovered specializations~\cite{sutton1999between, devin2017learning, ahn2022saycan}. The premise of this paper is that, before any of these, robot policies admit a more fundamental factorization---the one between the constraints imposed by the world and those imposed by the task. We argue this is the simplest non-trivial factorization available in robotics, in the sense that it follows from an invariance that the data itself exhibits: world quantities---sensing geometry, contact, dynamics---characterize the embodied system and its environment and remain stable when the task changes; task quantities---which goal to pursue first, when to accept short-term cost, when to defer to a teammate---vary across episodes while world constraints are fixed. This invariance is what makes the world/task factorization non-arbitrary, and it is the conditional independence structure we will show is exploited for Bayesian model comparison. This insight articulates our contributions:
\begin{enumerate}
    \item We argue that the world/task factorization, despite (or because of) its simplicity, is a principled basis for structurally generalizing robot policies, and should precede other choices.
    \item Theoretically, we provide a Bayesian model-evidence argument showing that, under a stated premise about the data-generating process of embodied systems, this factorization maximizes marginal likelihood among tractable alternatives (Sec. \ref{sec:theory}).
    \item Empirically, we instantiate the factorization as a hybrid framework pairing AICON~\cite{mengers2025no} with a learned gradient modulator (Sec.~\ref{sec:instantiation}), and show that gradients constitute a natural interface between world and task: the learned modulator yields sample-efficient training and near-optimal behavior in the task factor, while AICON's graph structure contributes structural compositionality and out-of-distribution generalization in the world factor (Sec.~\ref{sec:results}).
\end{enumerate}
We test our framework on three diverse scenarios under distinct operational paradigms: heterogeneous multi-robot search (coordination), obstacle-aware bimanual handover (kinematics), and pressure-plate puzzles (temporal logic).  Across all, the framework is sample-efficient, outperforms the purely learned and at least performs on par with the purely analytical baselines, generalizes zero-shot to larger teams, targets and unseen obstacles, and transfers directly to real hardware. 

\section{Related Work}\label{sec:related}

The challenge of generalizable robot behavior sits at the intersection of two research problems: encoding the structure of the physical world into controllers, and learning the constraints of specific tasks from experience. The gap between them has motivated a range of hybrid and hierarchical approaches that impose some factorization on the policy. We review each in turn, with attention to what is factored from what and what that implies for generalization.

\textbf{Encoding world structure for robust control.} 
A family of geometric and control approaches builds physical constraints directly into the action generation process. Riemannian Motion Policies and Geometric Fabrics compose task-space accelerations under positive-definite metrics, producing reactive and physically consistent behavior~\cite{ratliff2018riemannian, li2021rmp2, pantic2023obstacle, van2022geometric, merva2025globally}. Like classical potential field methods~\cite{khatib1986real}, however, these approaches resolve conflicts between competing objectives myopically: they have no mechanism for non-myopic preference over world-consistent paths. Variants that learn the Riemannian metric from demonstrations or rewards~\cite{calinon2020gaussians, rana2020learning, gruffaz2025riemannian, braun2024riemannian, ding2025fast, tennenholtz2022uncertainty, wang2023hierarchical, alhousani2023geometric} address some of this rigidity but encode task geometry in a single monolithic network, losing the structural compositionality of the underlying representation. AICON~\cite{mengers2025no} differs in that its graph-based structure produces a dynamic manifold whose gradients evolve with the estimation process, making it adaptable without retraining and naturally composable: new regularities can be added as nodes and edges without modifying existing components. Our framework builds on these properties, adding a learned policy that resolves task constraints that world-structure encoding alone cannot.

\textbf{Learning task constraints from experience.} At the other end of the spectrum, deep reinforcement learning and imitation learning~\cite{tang2025deep, hoeller2024anymal, lin2025learning} learn task constraints with high-capacity networks directly from rewards or demonstrations. Vision-language-action models~\cite{brohan2023rt2, kim2025openvla} extend this to richer input modalities and broader task repertoires. The strength of these approaches is flexibility, while their limitation is that world structure must be rediscovered from data at a significant cost: it drives up sample complexity and produces policies that are often fragile under distribution shift. 

\textbf{Hybrid and hierarchical approaches.} Several families of methods impose some factorization on the policy. Hierarchical RL factors along temporal scale~\cite{sutton1999between, bacon2017option}, skill libraries factor along primitives~\cite{ahn2022saycan, brohan2023rt2}, and Mixture-of-Experts and modular networks factor along discovered specializations~\cite{shazeer2017outrageously, andreas2016neural}. Bi-level optimization~\cite{dempe2020bilevel, liu2021investigating, hu2024bi} factors along fast versus slow dynamics, though usually assuming time-scale separation~\cite{das2025latent} or recovering the structure only implicitly through end-to-end learning~\cite{schmied2023learning}. A closely related class adapts a robust controller with a high-level learned policy~\cite{lambert2019low, carlucho2020adaptive, yang2025cbf, zhang2025learning}, which is structurally similar to our approach. The key difference is that these methods adapt whatever parameters the controller exposes, whereas our approach locates learning at the gradient interface the world/task factorization identifies, which is theoretically motivated (Sec.~\ref{sec:theory}) and produces structural consequences not shared with those methods (Sec. \ref{sec:instantiation}).

\textbf{Task and motion planning (TAMP), and world models.} 
TAMP separates high-level task decisions from continuous motion feasibility~\cite{garrett2021integrated}, resembling our world/task factorization but serving a different purpose: it uses models of the world to guide symbolic search by sampling plans, checking feasibility, and backtracking, whereas our framework uses world structure as a differentiable substrate for gradient propagation. The literature on world models~\cite{wu2023daydreamer, lecun2022path, hou2026world} asks a related but distinct question: what to predict from observations and how to couple prediction with action generation. The factorization question we raise here is largely absent: learned world models encode a mixture of physical regularities and task preferences, an entanglement identified as a ``causal~conditioning~gap''~\cite{hou2026world}.

\section{A Theoretical Bayesian Perspective on World/Task Factorization}\label{sec:theory}

We formulate the search for a policy as a model selection problem. A policy with parameters $\theta$ defines a model class $\mathcal{M}$, and we seek the class that maximizes the Bayesian evidence among classes instantiating different factorizations of $\theta$. This framing is appropriate because factorization is precisely what Bayesian model comparison penalizes or rewards: a factorization that aligns with the conditional independence structure of the data concentrates posterior mass efficiently; one that does not spreads prior mass over parameter combinations the data never realizes.
 
\textbf{Bayesian model evidence and the Occam factor.} Let a policy be a probabilistic model class $\mathcal{M}$ with parameters $\theta$ and prior $P(\theta \mid \mathcal{M})$. Given data $\mathcal{D}$, the evidence (marginal likelihood) of $\mathcal{M}$ is
\begin{equation}\label{eq:occam_og}
    P(\mathcal{D} \mid \mathcal{M}) = \int P(\mathcal{D} \mid \theta, \mathcal{M}) \, P(\theta \mid \mathcal{M}) \, d\theta.
\end{equation}
This quantity is the unbiased Bayesian criterion for comparing model classes \cite{mackay1992bayesian,mackay2003information, bishop2006pattern}: it penalizes models that spread their prior mass over parameter regions the data does not constrain, without any manual complexity term. Under a Laplace approximation around the posterior mode $\theta^\star$,
\begin{equation}\label{eq:occam_approx}
    P(\mathcal{D} \mid \mathcal{M}) \approx {P(\mathcal{D} \mid \theta^\star, \mathcal{M})} \cdot {\frac{(2\pi)^{d/2} \, |\Sigma_\theta|^{1/2}}{V_\theta}},
\end{equation}
where the first term is the best-fit likelihood and the second is the Occam factor, with $V_\theta$ the prior parameter volume, $|\Sigma_\theta|^{1/2}$ the posterior parameter volume, and $d = \dim(\theta)$. A model class with the right inductive bias has small $V_\theta$ relative to its best-fit likelihood.
 
\textbf{The world/task factorization maximizes Bayesian evidence.} Computing~\eqref{eq:occam_og} requires that the prior $P(\theta \mid \mathcal{M})$ admits a factorization the posterior can exploit. We call a factorization of $\theta = (\theta_A, \theta_B)$ tractable if the prior factors as
\begin{equation}\label{eq:tractable}
    P(\theta_A, \theta_B \mid \mathcal{M}) = P(\theta_A \mid \mathcal{M}) \cdot P(\theta_B \mid \theta_A, \mathcal{M})
\end{equation}
in a way that each conditional factor admits a Laplace approximation around its posterior mode. Among tractable factorizations, the one that maximizes evidence is the one whose independence structure matches the data-generating process. A prior that imposes independences absent from the likelihood inflates posterior volume $\Sigma_\theta$, depressing the Occam factor; one that ignores independences present in the likelihood spreads mass over parameter combinations the data does not realize, increasing $V_\theta$ and depressing the Occam factor again. Other candidate factorizations---hierarchical RL along temporal scale, skill libraries along primitives, MoE along discovered specializations, foundation models along web-scale tokenization---entangle world and task within each factor or lack a tractable closed-form prior, and so cannot exploit the conditional independence structure of robot data.
 
The relevant question is therefore: which factorization of robot policy parameters aligns with the conditional independence structure of robot data? We argue that this is the world/task factorization. World parameters $\theta_{\mathrm{world}}$---kinematics, sensing, geometry---govern the data-generating process regardless of which task is being executed; a robot arm's joint limits do not change when its reward function does. Task parameters $\theta_{\mathrm{task}}$---goal priorities, trade-offs, role assignments---vary across episodes while $\theta_{\mathrm{world}}$ is fixed. This invariance is the conditional independence the data itself exhibits. The world/task factorization is the prior factorization that matches it:
\begin{equation}\label{eq:wt_factor}
    P(\theta_{\mathrm{world}}, \theta_{\mathrm{task}} \mid \mathcal{M}_{\text{w/t}}) = P(\theta_{\mathrm{world}} \mid \mathcal{M}_{\text{w/t}}) \cdot P(\theta_{\mathrm{task}} \mid \theta_{\mathrm{world}}, \mathcal{M}_{\text{w/t}}).
\end{equation}

Among tractable candidates, three remain: two unfactored extremes---\emph{structure-only} and \emph{end-to-end}---and the factored \emph{world/task} class. A structure-only model class encodes the world analytically and carries no learned task component; its prior volume is small (a large Occam factor), but it lacks the capacity to resolve trade-offs between conflicting goals or long-horizon preferences, so its best-fit likelihood is poor on tasks that demand non-myopic behavior. An end-to-end model class sits at the opposite: a single unstructured network with prior parameter volume $V_\theta \sim |\Theta_{\text{NN}}|$, exponentially larger than any structured prior, making the Occam factor small and hindering generalization even when the best-fit likelihood is high. The world/task model class factors these two apart: it pairs the analytic world model with a learned task component, keeping the Occam factor small while inheriting physical correctness from the world model, so the best-fit likelihood stays high. The log-evidence ratio of the world/task class over the end-to-end one is
\begin{equation}\label{eq:evidence_ratio}
    \log \frac{P(\mathcal{D} \mid \mathcal{M}_{\text{w/t}})}{P(\mathcal{D} \mid \mathcal{M}_{\text{e2e}})} \;\approx\; \log \frac{V_{\text{e2e}}}{V_{\mathrm{world}} \cdot V_{\mathrm{task}}} \;+\; \bigl[\ell^\star_{\text{w/t}} - \ell^\star_{\text{e2e}}\bigr],
\end{equation}
with $\ell^\star_\bullet$ the maximum-likelihood log-score. If world parameters are invariant across tasks, encoding them analytically converts a large learned prior into a structured one with $V_{\mathrm{world}} \sim 1$, bounding the evidence gap below by the ratio of prior volumes---a consequence we probe empirically in Sec.~\ref{sec:results}.
 
\section{Instantiating the World/Task Factorization via Task Gradients}
\label{sec:instantiation}
 
Having formalized the world/task factorization for robot learning, we now study its practical implications. Consider a team of $N$ heterogeneous robots solving a Decentralized Partially Observable Markov Decision Process. Each robot $r$ has access only to a local observation $o_r^t$, comprising sensor measurements $z_r^t$, local state estimates $x_r^t$, and messages from neighbors in a time-varying communication graph. The team aims to maximize a global objective $J = \mathbb{E}\big[\sum_{t=0}^{\infty} \gamma^t R^t\big]$, where $R^t$ aggregates task-defined cost functions. Under the world/task factorization, $\theta_{\mathrm{world}}$ encodes the physical structure of robots and environment and is unaffected by learning; $\theta_{\mathrm{task}}$ is what the policy~learns.
 
\textbf{Estimation and control with AICON.} We instantiate the world factor $\theta_{\mathrm{world}}$ using AICON (Active InterCONnect)~\cite{mengers2025no}, which represents the robot's world model as a directed graph $\mathcal{G} = (\mathcal{V}, \mathcal{E})$. Each node $i \in \mathcal{V}$ is a Recursive Estimator (RE) maintaining a differentiable estimate $x_i \in \mathbb{R}^{n_i}$ of a world quantity (object pose, contact mode, robot position). Each edge $j \in \mathcal{E}$ is an Active Interconnection (AI), an implicit differentiable function $h_j(x_{i_1}, \ldots, x_{i_\ell}) = 0$ encoding a regularity---kinematic chains, geometric constraints, sensor models---that holds whenever its incident estimates are consistent. New regularities can be added as nodes and edges without modifying existing components, making the graph itself compositional, and since each RE updates using only its incident AIs, execution is inherently decentralized and asynchronous across heterogeneous robots. Because the graph encodes only physical regularities whose structure is invariant across tasks, it satisfies the tractability condition of Sec.~\ref{sec:theory}: $\theta_{\mathrm{world}}$ admits a structured prior with small volume $V_{\mathrm{world}}$, and the posterior over it can be evaluated independently of $\theta_{\mathrm{task}}$. We write $x = (x_i)_{i \in \mathcal{V}}$ for the joint estimate. A task is specified by $p$ differentiable cost functions $g_k : \mathbb{R}^{|x|} \to \mathbb{R}$. Each cost function can induce many gradient paths through the graph: a gradient path $\mathrm{p} \in \mathbb{P}$ is a sequence of differentiable functions connecting a cost to an actuation signal $a$ via the chain rule,
\begin{equation}\label{eq:chain}
    \nabla^\mathrm{p}_a g_k = \frac{\partial g_k}{\partial x_{i_1}} \frac{\partial x_{i_1}}{\partial x_{i_2}} \cdots \frac{\partial x_{i_{N-1}}}{\partial x_{i_N}} \frac{\partial x_{i_N}}{\partial a}\,.
\end{equation}
Different gradient paths reveal distinct subgoals: some reduce cost directly, others steer the system into more favorable estimation regimes. The base AICON action rule selects the steepest path,
\begin{equation}\label{eq:aicon_og}
    a^{t+1} = a^{t} - w \cdot \nabla^\mathrm{p^\star}_a g_{k}(x^{t}), \qquad \mathrm{p}^\star = \argmax_{\mathrm{p} \in \mathbb{P}} \bigl\| \nabla^\mathrm{p}_a g_k(x^{t}) \bigr\|,
\end{equation}
which is myopic: it selects the steepest gradient without knowledge of which direction within the feasible solution space is task-optimal. Nullspace projections can make conflict resolution feasible for a broad class of simultaneous conflicts~\cite{mengers2025nullspace}, but feasibility and directness are different problems---the learned modulator addresses the latter, enabling task-optimal trajectories rather than exploratory ones.
 
\textbf{Gradients as the interface for the task factor.} Given the factorization~\eqref{eq:wt_factor}, the task factor \mbox{$P(\theta_{\mathrm{task}} \mid \theta_{\mathrm{world}})$} must be parameterized over some space. Each gradient path $\nabla^\mathrm{p}_a g_k(x)$ is the chain-ruled product of a task quantity and a set of world quantities, as expressed in Eq. \eqref{eq:chain}. Gradients therefore simultaneously reflect world structure and task structure, making them a natural interface. By the KKT characterization of Pareto-stationarity~\cite{desideri2012multiple, fliege2000steepest}, every direction that descends all active paths simultaneously is a convex combination $\sum_\mathrm{p} \kappa_\mathrm{p} \nabla^\mathrm{p}_a g_k$ for $\kappa \in \Delta^{|\mathbb{P}|-1}$. However, when conflicts between paths cannot be resolved by any such combination---as when all gradients point into or away from a non-convex obstacle and the required direction lies in their nullspace---a Riemannian preconditioner $\Xi \in \mathbb{S}_{++}^{|a|}$ provides access to those directions~\cite{mengers2025nullspace, ratliff2018riemannian}. The full admissible family of task-conditioned descent laws is therefore
\begin{equation}\label{eq:admissible}
    a^{t+1} = a^t - \Xi \sum\nolimits_{\mathrm{p} \in \mathbb{P}} \kappa_\mathrm{p} \, \nabla^\mathrm{p}_a g_k(x^t), \qquad (\kappa, \Xi) \in \Delta^{|\mathbb{P}|-1} \times \mathbb{S}_{++}^{|a|}.
\end{equation}
Any smaller parameterization excludes admissible descent directions; any larger one introduces redundancy. Thus~\eqref{eq:admissible} is the maximum-evidence parameterization of the task factor in~\eqref{eq:evidence_ratio}, and \eqref{eq:aicon_og} is recovered as the corner $\kappa = e_{\mathrm{p}^\star}, \Xi = I$.
 
\textbf{Learning the modulator and consequences of the design.} A small learned policy $\pi(\cdot; \theta_{\mathrm{task}})$ maps each robot $r$'s local context $y_r^t$ to $(\kappa_r^t, \Xi_r^t)$,
\begin{equation}\label{eq:policy}
    \pi(y_r^t; \theta_{\mathrm{task}}) = (\kappa_r^t, \Xi_r^t), \quad \kappa_r^t = \mathrm{softmax}(z_\kappa(y_r^t; \theta_{\mathrm{task}})), \quad \Xi_r^t = \mathrm{Exp}(z_\Xi(y_r^t; \theta_{\mathrm{task}})),
\end{equation}
with $\mathrm{Exp}$ the matrix exponential ensuring positive-definiteness. Training optimizes only $\theta_{\mathrm{task}}$ via PPO or cross-entropy on expert gradient-path selections; $\theta_{\mathrm{world}}$ is never updated. Architectural details are in Appendix~\ref{app:policy_details}. This design has three consequences that follow directly from the factorization. First, learning is low-dimensional: policy outputs live on $\Delta^{|\mathbb{P}|-1} \times \mathbb{S}_{++}^{|a|}$, independent of state dimensionality, explaining the sample-efficiency gap in Sec.~\ref{sec:results}. Second, compositionality is structural: adding a world regularity at deployment instantiates a new RE/AI pair and extends $\mathbb{P}$ without changing $\theta_{\mathrm{task}}$, so the policy generalizes zero-shot. Third, sim-to-real transfer reduces to swapping modules in $\theta_{\mathrm{world}}$; the learned policy transfers untouched provided the gradient interface is preserved.

\section{Results}\label{sec:results}
 
\begin{figure}[t]
  \centering
  
  \newlength{\toprowh}
  \newlength{\rowbh}
  \newlength{\rowch}
  \newlength{\rowdh}
  \newlength{\roweh}

  \setlength{\toprowh}{0.16\textheight}
  \setlength{\rowbh}{0.135\textheight}
  \setlength{\rowch}{0.135\textheight}
  \setlength{\rowdh}{0.135\textheight}
  \setlength{\roweh}{0.135\textheight}

  \newlength{\labelcolw}
  \newlength{\imgcolw}
  \newlength{\plotw}

  \setlength{\labelcolw}{0.035\linewidth}
  \setlength{\imgcolw}{0.295\linewidth}
  \setlength{\plotw}{0.94\linewidth}

  \setlength{\tabcolsep}{3pt}
  \renewcommand{\arraystretch}{1.0}

  \begin{tabular}{@{}>{\centering\arraybackslash}m{\labelcolw}
                  >{\centering\arraybackslash}m{\imgcolw}
                  >{\centering\arraybackslash}m{\imgcolw}
                  >{\centering\arraybackslash}m{\imgcolw}@{}}

    & \hspace{0.05\imgcolw}{\footnotesize Search}
    & \hspace{0.2\imgcolw}{\footnotesize Handover}
    & \hspace{0.1\imgcolw}{\footnotesize Pressure Plate}
    \\[2pt]

    \panellabel{a}
    &
    \hspace{0.13\imgcolw}\figcell{width=0.83\imgcolw,keepaspectratio}{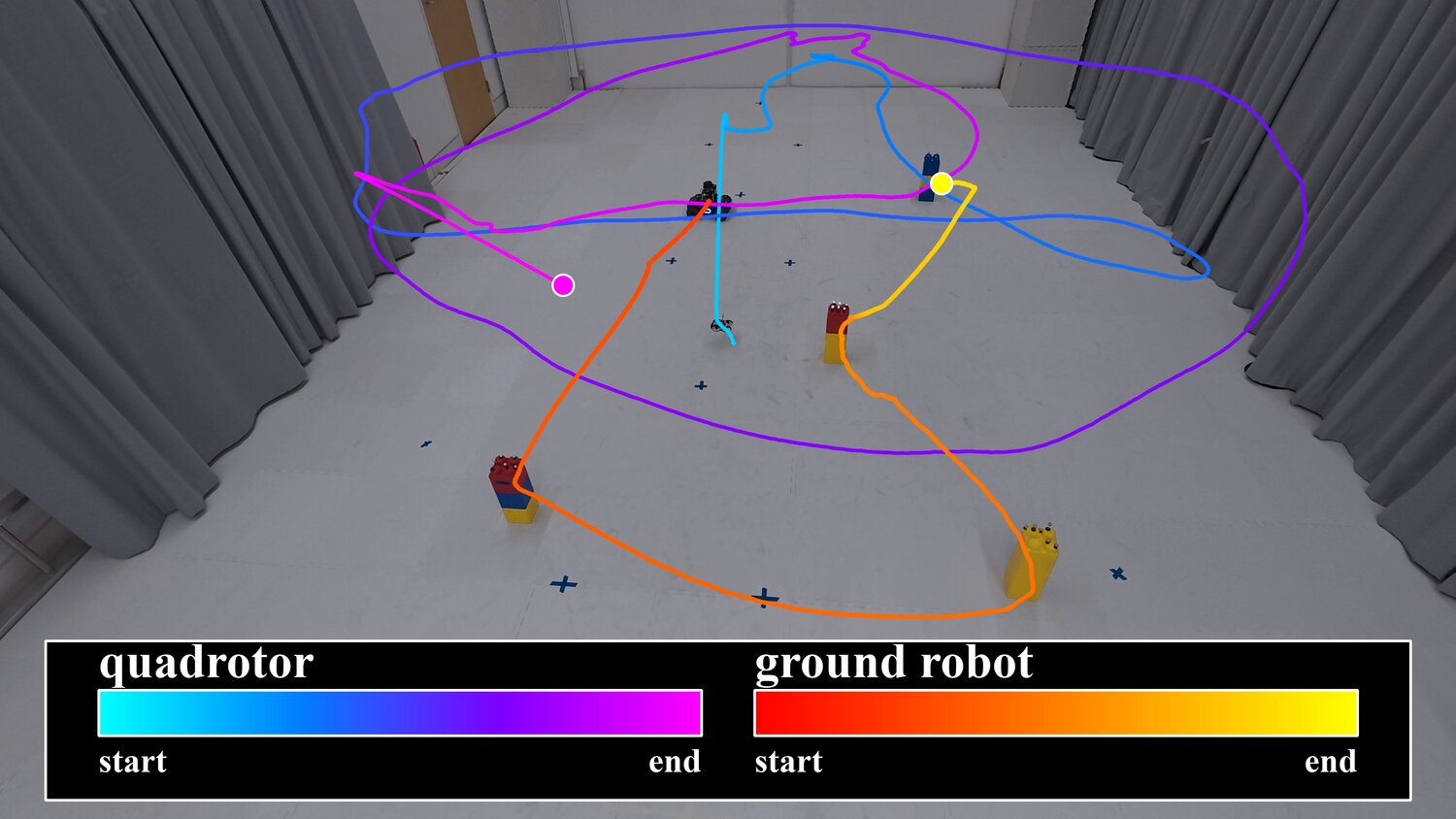}
    &
    \hspace{0.17\imgcolw}\figcell{width=0.83\imgcolw,keepaspectratio}{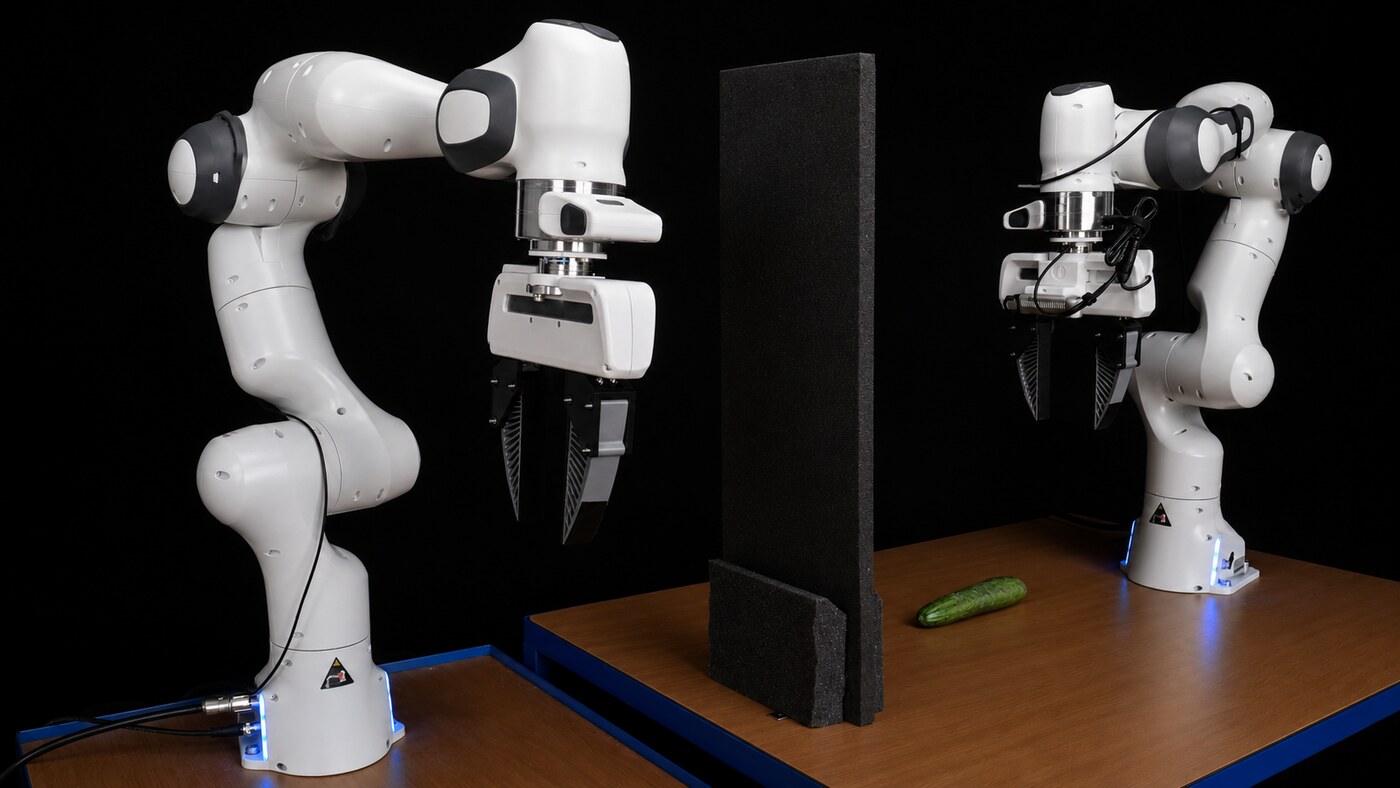}
    &
    \hspace{0.05\imgcolw}\figcell{width=0.83\imgcolw,keepaspectratio}{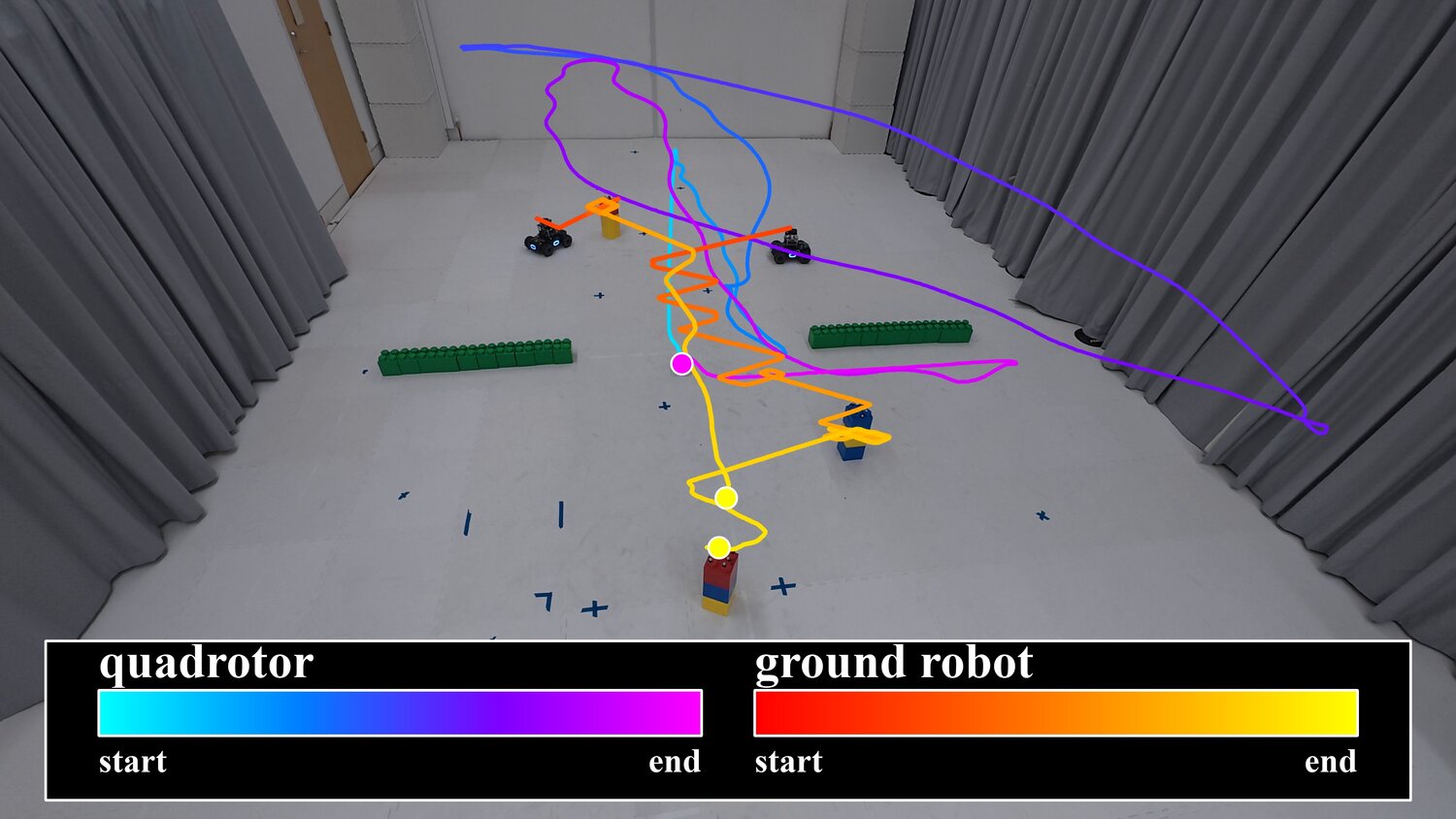}
    \\[4pt]

    \panellabel{b}
    &
    \multicolumn{3}{c@{}}{%
      \figcell{height=\rowbh,width=\plotw,keepaspectratio}{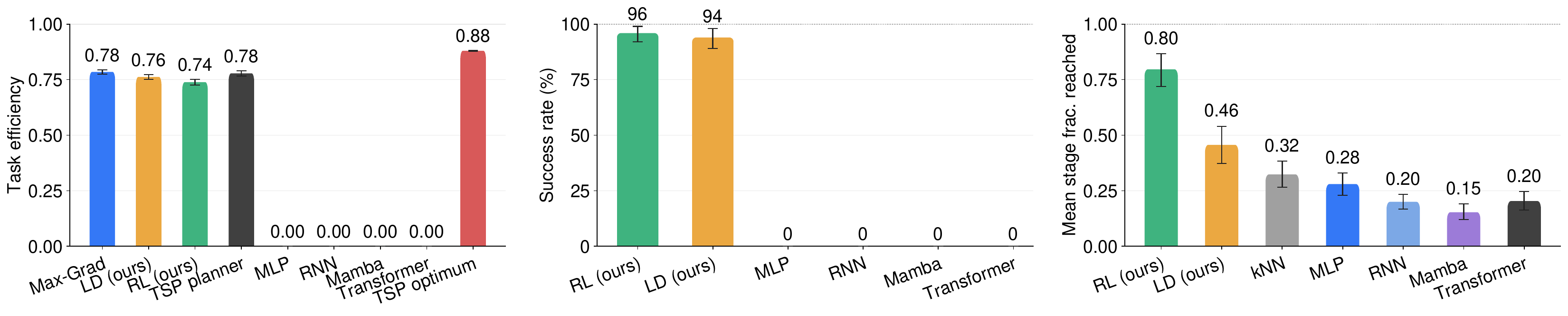}%
    }
    \\[3pt]

    \panellabel{c}
    &
    \multicolumn{3}{c@{}}{%
      \figcell{height=\rowch,width=\plotw,keepaspectratio}{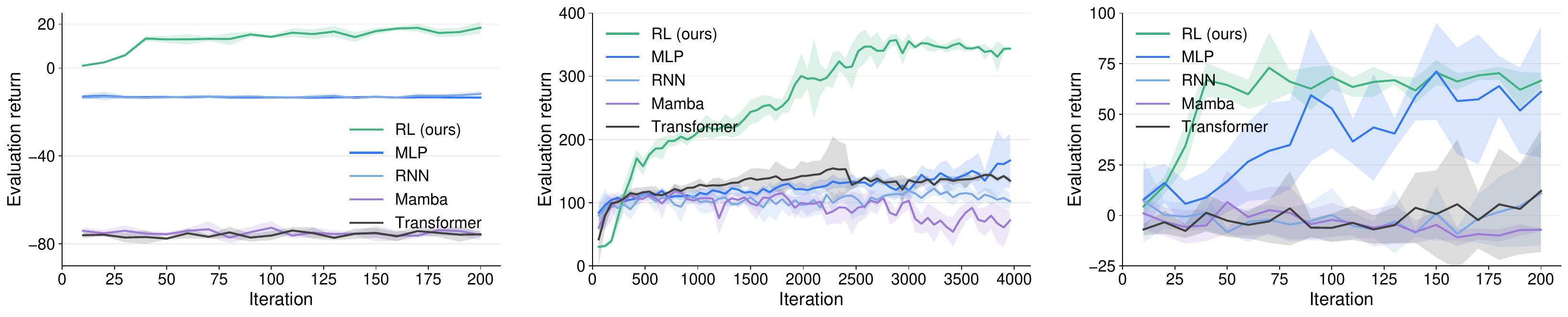}%
    }
    \\[3pt]

    \panellabel{d}
    &
    \multicolumn{3}{c@{}}{%
      \figcell{height=\rowdh,width=\plotw,keepaspectratio}{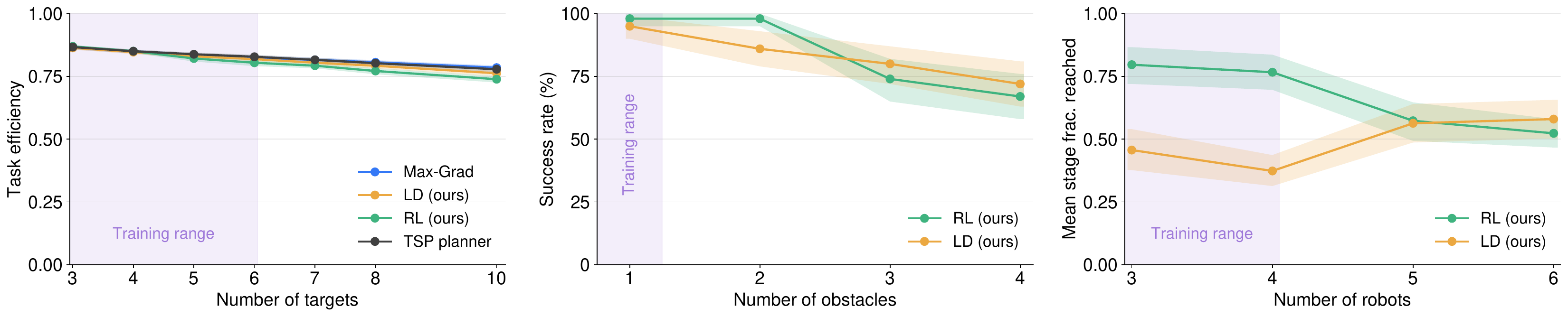}%
    }
    \\[3pt]

    \panellabel{e}
    &
    \multicolumn{3}{c@{}}{%
      \figcell{height=\roweh,width=\plotw,keepaspectratio}{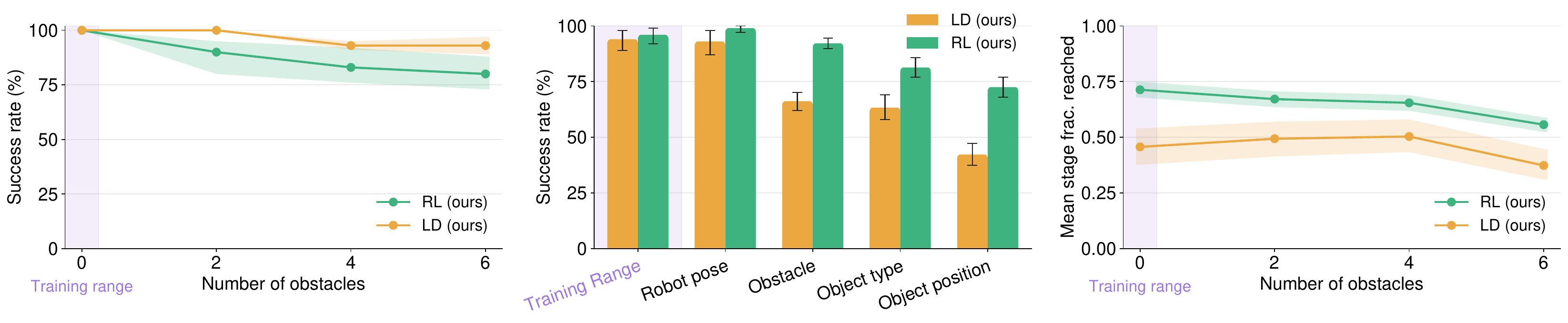}%
    }

  \end{tabular}
  \caption{{\small Task performance. (a) Experimental examples with real robots (time-colored trajectories for search and pressure plate, and a snapshot of the handover). (b) Performance vs. baselines over 5 seeds and 100 episodes (search: task efficiency as a fraction of remaining rollout steps when all targets are found; handover: success rate; pressure plate: mean stage fraction reached). (c) Training curves for our RL policy and learning baselines. (d) Zero-shot generalization beyond the training range (shaded): search vs. number of targets, handover vs. number of obstacles, and pressure plate vs. number of ground robots. (e) Zero-shot compositionality to out-of-distribution regimes: search vs. obstacles, handover vs. object types and poses, and pressure plate vs. obstacles}.}
  \label{fig:main_performance}
  \vspace{-0.5cm} 
\end{figure}

Our evaluation is organized around five questions (\textbf{Q1}--\textbf{Q5}) that together probe the two lines of evidence developed in Secs.~\ref{sec:theory} and~\ref{sec:instantiation}. \textbf{Q1} and \textbf{Q2} test the empirical prediction of the Bayesian argument: that locating learning at the gradient interface should yield sample efficiency gains over end-to-end baselines. \textbf{Q3} and \textbf{Q4} test the structural compositionality claim: that changes to the world factor at deployment should not require retraining the task factor. \textbf{Q5} tests sim-to-real transfer as module swaps. We answer these questions with three collaborative tasks, each chosen to probe a qualitatively different relationship between world structure and task difficulty (Fig.~\ref{fig:main_performance}a).
 
\textbf{Task 1 -- Search}: A 2D bounded arena contains $T$ static targets. The objective is for a quadrotor and a ground robot to locate and retrieve all targets as efficiently as possible. The quadrotor has a ``downward-facing'' sensor to maintain a discretized occupancy grid map of the environment. The ground robot is equipped with a LiDAR and is the only robot capable of reaching targets. The task is combinatorial, a partially observable TSP~\cite{applegate2011traveling}, but strong heuristics exist, letting us test whether learned gradient modulation recovers near-optimal behavior without hand-designed priorities.
 
\textbf{Task 2 -- Handover}: Two Franka Emika Panda manipulators, mounted on opposite sides of a table, must collaborate to hand over an object while avoiding an obstacle placed between them. The task constraint creates a local minimum: the gradient for reaching the object directly opposes the gradient for avoiding the obstacle, and gradient descent alone cannot escape it. In this task, success depends on precise inter-agent coordination, timing, and the continuous dynamics of high-DOF manipulators, where small deviations in motion can prevent a stable grasp or transfer. This requires the learned policy to express task-specific, long-term preference over world-consistent paths.
 
\textbf{Task 3 -- Pressure Plate}: A 2D arena is divided into two rooms by a central wall with a door. Two pressure plates are randomly placed, one in each room; the door opens only when a robot stands on either plate. A goal zone is located in the right room, and $N$ ground robots equipped with LiDARs must all reach it, aided by a quadrotor that maintains an occupancy grid map. This task requires robots with identical capabilities to implicitly differentiate roles---some holding plates, others crossing---under partial observability. No myopic heuristic resolves the combinatorial interdependence of these roles, making this the hardest test of the task factor.
 
\textbf{Baselines}: For all tasks, we consider \textbf{RL} and \textbf{LD} as our method trained with rewards and demonstrations respectively. We also compare with non-modulated \textbf{AICON}, which uses the same graph of REs and AIs but only applies Eq. \eqref{eq:aicon_og}. As purely learning-based methods we use \textbf{MLP}~\cite{rumelhart1986learning}, \textbf{RNN}~\cite{elman1990finding}, \textbf{Mamba}~\cite{gu2024mamba}, and \textbf{Transformer}~\cite{vaswani2017attention} as representative architectures spanning memoryless, recurrent, and attention-based approaches to partial observability. For the search task we additionally compare with \textbf{Max-Grad}, \textbf{TSP planner}, and \textbf{TSP optimum}; for the pressure plate task with \textbf{KNN}. Policy, environment, and training details are in Appendices~\ref{app:policy_details},~\ref{app:environments}, and~\ref{app:training}; AICON graphs are in Appendix~\ref{app:aicon_graphs}.

\textbf{(Q1) Does our approach outperform purely learned and purely analytical baselines?} 
Figure~\ref{fig:main_performance}b reports performance across the three tasks, aggregated over 5 seeds and 100 episodes. Both variants of our method consistently outperform every end-to-end learned baseline, which collapse to near-zero in all tasks---confirming that confining learning to the low-dimensional space of gradient paths is far more effective than learning motor commands from scratch. The comparison against analytical baselines depends on how well the task is understood: in search, decades of TSP research yield heuristics (Max-Grad) already near the brute-force optimum, and all informed methods cluster together. We read this as a positive sanity check---when a near-optimal heuristic exists, our method recovers comparable performance with no hand-designed priorities. In handover and pressure plate, where no such heuristic exists, RL and LD dominate every baseline by a wide margin, supporting our central claim that the world/task factorization enables sample-efficient learning of complex policies.

\textbf{(Q2) How do reward-based and demonstration-based supervision compare?} 
Across all tasks, LD matches or outperforms RL in final cost using only 10 demonstrations per task (Fig.~\ref{fig:main_performance}b--c): demonstrations provide a dense, targeted signal on which gradient paths to prioritize, whereas PPO must discover this from reward. Still, the RL variant converges with few environment interactions in every task (less than 100, 3000 and 75 respectively), despite partial observability, raw high-dimensional sensor inputs, and multi-robot interaction.

\textbf{(Q3) Does our approach generalize with team size and task complexity?}
Generalization (Fig.~\ref{fig:main_performance}d) in our framework operates through two complementary mechanisms: AICON's graph absorbs new entities as RE/AI pairs, extending the gradient path set without modifying existing components; and the learned modulator is permutation-invariant over gradient paths, so its output dimensionality does not grow with team size or task complexity. In the search task, a single model trained with 2--6 targets transfers without retraining to up to 10 targets: performance degrades by only 5\% at 10 targets in search, matching the trend of the brute-force TSP optimum. In the pressure plate task, stage fraction stays above 50\% when transferring a model for 3--4 robots to up to 6, improving with more robots since larger teams create more opportunities for stage transitions. A model trained with 1 obstacle in the handover task similarly performs well with up to 4 obstacles. Performance degrades gracefully rather than catastrophically throughout.

\textbf{(Q4) Do our policies compose zero-shot to out-of-distribution
conditions?} 
While Q3 tests scaling within known configuration types, Q4 tests transfer to qualitatively novel ones (Fig.~\ref{fig:main_performance}e). In search and pressure plate, we insert static obstacles never present during training: each enters the AICON graph as a new RE/AI pair without altering the task factor, and both RL and LD retain high success with increasing obstacle density. In handover, we vary robot pose, obstacle size and shape, object type, and object pose one factor at a time: LD degrades gracefully while RL retains high success for the same reason---the world factor absorbs each new constraint as graph structure and the task factor can be reused verbatim. Together these results show that the factorization supports genuine out-of-distribution generalization, not just scaling.

\textbf{(Q5) Do our policies transfer to real robots?} 
We deploy all our policies on hardware with no retraining: because AICON isolates sensing and estimation behind the RE/AI abstraction, transfer reduces to swapping simulated sensor drivers in the world factor for their physical counterparts, while the learned modulator is used untouched. In search (Fig.~\ref{fig:main_performance}), RL and LD each succeed in 8/8 deployments (100\%), despite a real quadrotor subject to true flight dynamics and localization drift, sensor noise that alters the exploration pattern, a LiDAR of different range, and a variable number of targets across runs. In handover, the policy succeeds in 9/10 deployments (90\%), robust even though the object's shape and weight, the arms' placement, orientation, and gripper morphology are all out of training distribution; the only failure occurs when the trajectory demands motion beyond the arm's unmodelled actuator limits---a regime that is not physically realizable in any case. In pressure plate, under the same quadrotor and LiDAR changes and a variable number of ground robots (4 deployments per method), RL succeeds 4/4 (100\%) and LD 3/4 (75\%). The single LD failure is not a sim-to-real artifact: it is a failed stage transition when there are only two ground robots---which is out-of-distribution and a failure mode already visible in simulation (Fig.~\ref{fig:main_performance}b). The error therefore lives in the learned task factor and is preserved across the reality gap rather than introduced by it, direct evidence that the world factor transfers cleanly and residual errors are attributable to task learning, not world modeling. Extended results are in Appendix \ref{app:experiments} and the supplementary video.

\section{Limitations}\label{sec:limitations}

The central conceptual claim, that world and task parameters are asymmetrically independent, and that this independence is what makes the factorization productive, rests on a premise that holds clearly for kinematics, geometry, and sensor models, but becomes less sharp at the boundary. Consider grasp selection: the optimal grasp on a hammer depends jointly on the world properties of geometry and mass distribution and on what the robot intends to do with it: whether to drive a nail or pass it to a teammate. Neither side fully determines the answer. Constraints of this kind could in principle be modeled on either side, and the choice affects sample efficiency, generalization, and interpretability in ways we do not yet fully characterize. A principled methodology for deciding what belongs in each factor remains an open problem. The Bayesian argument in Sec.~\ref{sec:theory} is a theoretical motivation under stated assumptions, not a proof in full generality. Three assumptions are load-bearing: the Laplace approximation, the tractability condition on the prior, and the assumption that the world admits a complete analytical description so that $V_{\mathrm{world}} \sim 1$. In particular, the last assumption might hold cleanly in our setting but not extend to domains where the world model must itself be partially learned from data, a case we leave to future work. Our empirical evaluation was deliberately biased toward tasks for which strong heuristics or expert policies are available, giving us a gold-standard reference against which to isolate the contribution of the learned modulation. This makes the factorization's benefits interpretable but leaves open how the framework performs in less structured settings where an analytical AICON model, heuristics, reward functions, or demonstration data are not easily obtained. Extending the framework to such settings is the most important direction for future work.

\section{Conclusions}\label{sec:conclusion} 
The question posed in this paper is not whether to build structure into robot policies but which structure, and along which factorization. We argued that the world/task factorization should precede other structural choices, because it aligns with an invariance the data itself exhibits: world parameters remain stable across tasks while task parameters vary. We formalized this through a Bayesian model-evidence argument showing that this factorization maximizes marginal likelihood among tractable alternatives. We instantiated it by pairing AICON's compositional, graph-based world model with a small learned policy modulating gradient paths---an interface spanning the full admissible family of task-conditioned descent laws. This design yields three consequences by construction: low-dimensional learning, structural compositionality inherited from AICON's graph structure, and sim-to-real transfer as module swaps. Across three collaborative tasks probing qualitatively different relationships between world structure and task difficulty, both lines of evidence held: the framework was sample-efficient, generalized zero-shot to larger teams and unseen configurations, and transferred to real hardware without retraining. The broader claim is that when choosing how to structure a robot policy, the right question to ask first is whether the factorization respects the conditional independence structure of the problem---and that getting this right is what makes the resulting policy composable.

\acknowledgments{E. Sebasti\'an and A. Prorok gratefully acknowledge the support provided by a Leverhulme Trust Research Project Grant. A. Pfisterer, V. Mengers and O. Brock gratefully acknowledge funding by the Deutsche Forschungsgemeinschaft (DFG, German Research Foundation) under Germany's Excellence Strategy -- EXC 2002/1 ``Science of Intelligence'' -- project number 390523135, and by the German Federal Ministry of Research, Technology and Space (BMFTR) under the Robotics Institute Germany (RIG).}

\bibliography{references}

\begin{thebibliography}{54}
\providecommand{\natexlab}[1]{#1}
\providecommand{\url}[1]{\texttt{#1}}
\expandafter\ifx\csname urlstyle\endcsname\relax
  \providecommand{\doi}[1]{doi: #1}\else
  \providecommand{\doi}{doi: \begingroup \urlstyle{rm}\Url}\fi

\bibitem[Brohan et~al.(2023)Brohan, Brown, Carbajal, Chebotar, Chen, Choromanski, Ding, Driess, Dubey, Finn, et~al.]{brohan2023rt2}
A.~Brohan, N.~Brown, J.~Carbajal, Y.~Chebotar, X.~Chen, K.~Choromanski, T.~Ding, D.~Driess, A.~Dubey, C.~Finn, et~al.
\newblock {RT-2}: Vision-language-action models transfer web knowledge to robotic control.
\newblock In \emph{Conference on Robot Learning}, 2023.

\bibitem[Intelligence et~al.(2025)Intelligence, Black, Brown, Darpinian, Dhabalia, Driess, Esmail, Equi, Finn, Fusai, et~al.]{intelligence2025pi_}
P.~Intelligence, K.~Black, N.~Brown, J.~Darpinian, K.~Dhabalia, D.~Driess, A.~Esmail, M.~Equi, C.~Finn, N.~Fusai, et~al.
\newblock $\pi_{0.5}$: a vision-language-action model with open-world generalization.
\newblock \emph{arXiv preprint arXiv:2504.16054}, 2025.

\bibitem[Battaglia et~al.(2018)Battaglia, Hamrick, Bapst, Sanchez-Gonzalez, Zambaldi, Malinowski, Tacchetti, Raposo, Santoro, Faulkner, et~al.]{battaglia2018relational}
P.~W. Battaglia, J.~B. Hamrick, V.~Bapst, A.~Sanchez-Gonzalez, V.~Zambaldi, M.~Malinowski, A.~Tacchetti, D.~Raposo, A.~Santoro, R.~Faulkner, et~al.
\newblock Relational inductive biases, deep learning, and graph networks.
\newblock \emph{arXiv preprint arXiv:1806.01261}, 2018.

\bibitem[Sutton et~al.(1999)Sutton, Precup, and Singh]{sutton1999between}
R.~S. Sutton, D.~Precup, and S.~Singh.
\newblock Between {MDPs} and semi-{MDPs}: A framework for temporal abstraction in reinforcement learning.
\newblock \emph{Artificial Intelligence}, 112\penalty0 (1-2):\penalty0 181--211, 1999.

\bibitem[Devin et~al.(2017)Devin, Gupta, Darrell, Abbeel, and Levine]{devin2017learning}
C.~Devin, A.~Gupta, T.~Darrell, P.~Abbeel, and S.~Levine.
\newblock Learning modular neural network policies for multi-task and multi-robot transfer.
\newblock In \emph{IEEE International Conference on Robotics and Automation}, pages 2169--2176, 2017.

\bibitem[Brohan et~al.(2023)Brohan, Chebotar, Finn, Hausman, Herzog, Ho, Ibarz, Irpan, Jang, Julian, et~al.]{ahn2022saycan}
A.~Brohan, Y.~Chebotar, C.~Finn, K.~Hausman, A.~Herzog, D.~Ho, J.~Ibarz, A.~Irpan, E.~Jang, R.~Julian, et~al.
\newblock Do as {I} can, not as {I} say: Grounding language in robotic affordances.
\newblock In \emph{Conference on Robot Learning}, pages 287--318, 2023.

\bibitem[Mengers and Brock(2025)]{mengers2025no}
V.~Mengers and O.~Brock.
\newblock No plan but everything under control: Robustly solving sequential tasks with dynamically composed gradient descent.
\newblock In \emph{IEEE International Conference on Robotics and Automation}, pages 90--96, 2025.

\bibitem[Ratliff et~al.(2018)Ratliff, Issac, Kappler, Birchfield, and Fox]{ratliff2018riemannian}
N.~D. Ratliff, J.~Issac, D.~Kappler, S.~Birchfield, and D.~Fox.
\newblock Riemannian motion policies.
\newblock \emph{arXiv preprint arXiv:1801.02854}, 2018.

\bibitem[Li et~al.(2021)Li, Cheng, Rana, Xie, Van~Wyk, Ratliff, and Boots]{li2021rmp2}
A.~Li, C.-A. Cheng, M.~A. Rana, M.~Xie, K.~Van~Wyk, N.~Ratliff, and B.~Boots.
\newblock {RMP2}: A structured composable policy class for robot learning.
\newblock \emph{Robotics: Science and Systems}, 2021.

\bibitem[Pantic et~al.(2023)Pantic, Meijer, B{\"a}hnemann, Alatur, Andersson, Cadena, Siegwart, and Ott]{pantic2023obstacle}
M.~Pantic, I.~Meijer, R.~B{\"a}hnemann, N.~Alatur, O.~Andersson, C.~Cadena, R.~Siegwart, and L.~Ott.
\newblock Obstacle avoidance using raycasting and {Riemannian} motion policies at khz rates for mavs.
\newblock In \emph{IEEE International Conference on Robotics and Automation}, pages 1666--1672, 2023.

\bibitem[Van~Wyk et~al.(2022)Van~Wyk, Xie, Li, Rana, Babich, Peele, Wan, Akinola, Sundaralingam, Fox, et~al.]{van2022geometric}
K.~Van~Wyk, M.~Xie, A.~Li, M.~A. Rana, B.~Babich, B.~Peele, Q.~Wan, I.~Akinola, B.~Sundaralingam, D.~Fox, et~al.
\newblock Geometric fabrics: Generalizing classical mechanics to capture the physics of behavior.
\newblock \emph{IEEE Robotics and Automation Letters}, 7\penalty0 (2):\penalty0 3202--3209, 2022.

\bibitem[Merva et~al.(2025)Merva, Bakker, Spahn, Zhao, Virgala, and Alonso-Mora]{merva2025globally}
T.~Merva, S.~Bakker, M.~Spahn, D.~Zhao, I.~Virgala, and J.~Alonso-Mora.
\newblock Globally-guided geometric fabrics for reactive mobile manipulation in dynamic environments.
\newblock \emph{IEEE Robotics and Automation Letters}, 2025.

\bibitem[Khatib(1986)]{khatib1986real}
O.~Khatib.
\newblock Real-time obstacle avoidance for manipulators and mobile robots.
\newblock \emph{The International Journal of Robotics Research}, 5\penalty0 (1):\penalty0 90--98, 1986.

\bibitem[Calinon(2020)]{calinon2020gaussians}
S.~Calinon.
\newblock Gaussians on {Riemannian} manifolds: Applications for robot learning and adaptive control.
\newblock \emph{IEEE Robotics \& Automation Magazine}, 27\penalty0 (2):\penalty0 33--45, 2020.

\bibitem[Rana et~al.(2020)Rana, Li, Ravichandar, Mukadam, Chernova, Fox, Boots, and Ratliff]{rana2020learning}
M.~A. Rana, A.~Li, H.~Ravichandar, M.~Mukadam, S.~Chernova, D.~Fox, B.~Boots, and N.~Ratliff.
\newblock Learning reactive motion policies in multiple task spaces from human demonstrations.
\newblock In \emph{Conference on Robot Learning}, pages 1457--1468. PMLR, 2020.

\bibitem[Gruffaz and Sassen(2025)]{gruffaz2025riemannian}
S.~Gruffaz and J.~Sassen.
\newblock Riemannian metric learning: Closer to you than you imagine.
\newblock \emph{arXiv preprint arXiv:2503.05321}, 2025.

\bibitem[Braun et~al.(2024)Braun, Jaquier, Rozo, and Asfour]{braun2024riemannian}
M.~Braun, N.~Jaquier, L.~Rozo, and T.~Asfour.
\newblock Riemannian flow matching policy for robot motion learning.
\newblock In \emph{IEEE/RSJ International Conference on Intelligent Robots and Systems}, pages 5144--5151, 2024.

\bibitem[Ding et~al.(2025)Ding, Jaquier, Peters, and Rozo]{ding2025fast}
H.~Ding, N.~Jaquier, J.~Peters, and L.~Rozo.
\newblock Fast and robust visuomotor riemannian flow matching policy.
\newblock \emph{IEEE Transactions on Robotics}, pages 5327--5343, 2025.

\bibitem[Tennenholtz and Mannor(2022)]{tennenholtz2022uncertainty}
G.~Tennenholtz and S.~Mannor.
\newblock Uncertainty estimation using riemannian model dynamics for offline reinforcement learning.
\newblock \emph{Advances in Neural Information Processing Systems}, 35:\penalty0 19008--19021, 2022.

\bibitem[Wang et~al.(2023)Wang, Sagawa, and Yoshiyasu]{wang2023hierarchical}
Y.~Wang, R.~Sagawa, and Y.~Yoshiyasu.
\newblock A hierarchical robot learning framework for manipulator reactive motion generation via multi-agent reinforcement learning and riemannian motion policies.
\newblock \emph{IEEE Access}, 11:\penalty0 126979--126994, 2023.

\bibitem[Alhousani et~al.(2023)Alhousani, Saveriano, Sevinc, Abdulkuddus, Kose, and Abu-Dakka]{alhousani2023geometric}
N.~Alhousani, M.~Saveriano, I.~Sevinc, T.~Abdulkuddus, H.~Kose, and F.~J. Abu-Dakka.
\newblock Geometric reinforcement learning for robotic manipulation.
\newblock \emph{IEEE Access}, 11:\penalty0 111492--111505, 2023.

\bibitem[Tang et~al.(2025)Tang, Abbatematteo, Hu, Chandra, Mart{\'\i}n-Mart{\'\i}n, and Stone]{tang2025deep}
C.~Tang, B.~Abbatematteo, J.~Hu, R.~Chandra, R.~Mart{\'\i}n-Mart{\'\i}n, and P.~Stone.
\newblock Deep reinforcement learning for robotics: A survey of real-world successes.
\newblock \emph{Annual Review of Control, Robotics, and Autonomous Systems}, 8\penalty0 (1):\penalty0 153--188, 2025.

\bibitem[Hoeller et~al.(2024)Hoeller, Rudin, Sako, and Hutter]{hoeller2024anymal}
D.~Hoeller, N.~Rudin, D.~Sako, and M.~Hutter.
\newblock {ANYmal} parkour: Learning agile navigation for quadrupedal robots.
\newblock \emph{Science Robotics}, 9\penalty0 (88):\penalty0 eadi7566, 2024.

\bibitem[Lin et~al.(2025)Lin, Zhang, Li, Qi, Yi, Levine, and Malik]{lin2025learning}
T.~Lin, Y.~Zhang, Q.~Li, H.~Qi, B.~Yi, S.~Levine, and J.~Malik.
\newblock Learning visuotactile skills with two multifingered hands.
\newblock In \emph{IEEE International Conference on Robotics and Automation}, pages 5637--5643, 2025.

\bibitem[Kim et~al.(2024)Kim, Pertsch, Karamcheti, Xiao, Balakrishna, Nair, Rafailov, Foster, Lam, Sanketi, et~al.]{kim2025openvla}
M.~J. Kim, K.~Pertsch, S.~Karamcheti, T.~Xiao, A.~Balakrishna, S.~Nair, R.~Rafailov, E.~Foster, G.~Lam, P.~Sanketi, et~al.
\newblock {OpenVLA}: An open-source vision-language-action model.
\newblock In \emph{Conference on Robot Learning}, 2024.

\bibitem[Bacon et~al.(2017)Bacon, Harb, and Precup]{bacon2017option}
P.-L. Bacon, J.~Harb, and D.~Precup.
\newblock The option-critic architecture.
\newblock In \emph{Proceedings of the {AAAI} Conference on Artificial Intelligence}, volume~31, 2017.

\bibitem[Shazeer et~al.(2017)Shazeer, Mirhoseini, Maziarz, Davis, Le, Hinton, and Dean]{shazeer2017outrageously}
N.~Shazeer, A.~Mirhoseini, K.~Maziarz, A.~Davis, Q.~V. Le, G.~E. Hinton, and J.~Dean.
\newblock Outrageously large neural networks: The sparsely-gated mixture-of-experts layer.
\newblock In \emph{International Conference on Learning Representations}, 2017.

\bibitem[Andreas et~al.(2016)Andreas, Rohrbach, Darrell, and Klein]{andreas2016neural}
J.~Andreas, M.~Rohrbach, T.~Darrell, and D.~Klein.
\newblock Neural module networks.
\newblock In \emph{{IEEE} Conference on Computer Vision and Pattern Recognition}, pages 39--48, 2016.

\bibitem[Dempe(2020)]{dempe2020bilevel}
S.~Dempe.
\newblock Bilevel optimization: theory, algorithms, applications and a bibliography.
\newblock In \emph{Bilevel optimization: advances and next challenges}, pages 581--672. Springer, 2020.

\bibitem[Liu et~al.(2021)Liu, Gao, Zhang, Meng, and Lin]{liu2021investigating}
R.~Liu, J.~Gao, J.~Zhang, D.~Meng, and Z.~Lin.
\newblock Investigating bi-level optimization for learning and vision from a unified perspective: A survey and beyond.
\newblock \emph{IEEE Transactions on Pattern Analysis and Machine Intelligence}, 44\penalty0 (12):\penalty0 10045--10067, 2021.

\bibitem[Hu et~al.(2024)Hu, Shishika, Xiao, and Wang]{hu2024bi}
Z.~Hu, D.~Shishika, X.~Xiao, and X.~Wang.
\newblock Bi-cl: A reinforcement learning framework for robots coordination through bi-level optimization.
\newblock In \emph{IEEE/RSJ International Conference on Intelligent Robots and Systems}, pages 581--586, 2024.

\bibitem[Das et~al.(2025)Das, Chiu, Huang, Lindemann, and Sukhatme]{das2025latent}
S.~Das, D.~Chiu, Z.~Huang, L.~Lindemann, and G.~S. Sukhatme.
\newblock Latent activation editing: Inference-time refinement of learned policies for safer multirobot navigation.
\newblock \emph{arXiv preprint arXiv:2509.20623}, 2025.

\bibitem[Schmied et~al.(2023)Schmied, Hofmarcher, Paischer, Pascanu, and Hochreiter]{schmied2023learning}
T.~Schmied, M.~Hofmarcher, F.~Paischer, R.~Pascanu, and S.~Hochreiter.
\newblock Learning to modulate pre-trained models in {RL}.
\newblock \emph{Advances in Neural Information Processing Systems}, 36:\penalty0 38231--38265, 2023.

\bibitem[Lambert et~al.(2019)Lambert, Drew, Yaconelli, Levine, Calandra, and Pister]{lambert2019low}
N.~O. Lambert, D.~S. Drew, J.~Yaconelli, S.~Levine, R.~Calandra, and K.~S. Pister.
\newblock Low-level control of a quadrotor with deep model-based reinforcement learning.
\newblock \emph{IEEE Robotics and Automation Letters}, 4\penalty0 (4):\penalty0 4224--4230, 2019.

\bibitem[Carlucho et~al.(2020)Carlucho, De~Paula, and Acosta]{carlucho2020adaptive}
I.~Carlucho, M.~De~Paula, and G.~G. Acosta.
\newblock An adaptive deep reinforcement learning approach for {MIMO PID} control of mobile robots.
\newblock \emph{ISA Transactions}, 102:\penalty0 280--294, 2020.

\bibitem[Yang et~al.(2025)Yang, Werner, de~Sa, and Ames]{yang2025cbf}
L.~Yang, B.~Werner, M.~de~Sa, and A.~D. Ames.
\newblock {CBF-RL}: Safety filtering reinforcement learning in training with control barrier functions.
\newblock \emph{arXiv preprint arXiv:2510.14959}, 2025.

\bibitem[Zhang et~al.(2025)Zhang, Loquercio, Tang, Wang, Malik, and Mueller]{zhang2025learning}
D.~Zhang, A.~Loquercio, J.~Tang, T.-H. Wang, J.~Malik, and M.~W. Mueller.
\newblock A learning-based quadcopter controller with extreme adaptation.
\newblock \emph{IEEE Transactions on Robotics}, 41:\penalty0 3948--3964, 2025.

\bibitem[Garrett et~al.(2021)Garrett, Chitnis, Holladay, Kim, Silver, Kaelbling, and Lozano-P{\'e}rez]{garrett2021integrated}
C.~R. Garrett, R.~Chitnis, R.~Holladay, B.~Kim, T.~Silver, L.~P. Kaelbling, and T.~Lozano-P{\'e}rez.
\newblock Integrated task and motion planning.
\newblock \emph{Annual review of control, robotics, and autonomous systems}, 4\penalty0 (1):\penalty0 265--293, 2021.

\bibitem[Wu et~al.(2023)Wu, Escontrela, Hafner, Abbeel, and Goldberg]{wu2023daydreamer}
P.~Wu, A.~Escontrela, D.~Hafner, P.~Abbeel, and K.~Goldberg.
\newblock Daydreamer: World models for physical robot learning.
\newblock In \emph{Conference on robot learning}, pages 2226--2240. PMLR, 2023.

\bibitem[LeCun et~al.(2022)]{lecun2022path}
Y.~LeCun et~al.
\newblock A path towards autonomous machine intelligence version 0.9. 2, 2022-06-27.
\newblock \emph{Open Review}, 62\penalty0 (1):\penalty0 1--62, 2022.

\bibitem[Hou et~al.(2026)Hou, Li, Jia, An, Guo, Leng, Geng, Ze, Harada, Torr, et~al.]{hou2026world}
B.~Hou, G.~Li, J.~Jia, T.~An, X.~Guo, S.~Leng, H.~Geng, Y.~Ze, T.~Harada, P.~Torr, et~al.
\newblock World model for robot learning: A comprehensive survey.
\newblock \emph{arXiv preprint arXiv:2605.00080}, 2026.

\bibitem[MacKay(1992)]{mackay1992bayesian}
D.~J. MacKay.
\newblock Bayesian interpolation.
\newblock \emph{Neural Computation}, 4\penalty0 (3):\penalty0 415--447, 1992.

\bibitem[MacKay(2003)]{mackay2003information}
D.~J. MacKay.
\newblock \emph{Information theory, inference, and learning algorithms}.
\newblock Cambridge University Press, 2003.

\bibitem[Bishop(2006)]{bishop2006pattern}
C.~M. Bishop.
\newblock \emph{Pattern recognition and machine learning}.
\newblock Springer, 2006.

\bibitem[Mengers and Brock(2026)]{mengers2025nullspace}
V.~Mengers and O.~Brock.
\newblock Riding the shifting potential: When reactive control suffices for multi-goal behavior.
\newblock \emph{arXiv preprint arXiv:2605.27314}, 2026.

\bibitem[D{\'e}sid{\'e}ri(2012)]{desideri2012multiple}
J.-A. D{\'e}sid{\'e}ri.
\newblock Multiple-gradient descent algorithm ({MGDA}) for multiobjective optimization.
\newblock \emph{Comptes Rendus Math{\'e}matique}, 350\penalty0 (5--6):\penalty0 313--318, 2012.

\bibitem[Fliege and Svaiter(2000)]{fliege2000steepest}
J.~Fliege and B.~F. Svaiter.
\newblock Steepest descent methods for multicriteria optimization.
\newblock \emph{Mathematical Methods of Operations Research}, 51\penalty0 (3):\penalty0 479--494, 2000.

\bibitem[Applegate et~al.(2011)Applegate, Bixby, Chv{\'a}tal, and Cook]{applegate2011traveling}
D.~L. Applegate, R.~E. Bixby, V.~Chv{\'a}tal, and W.~J. Cook.
\newblock The traveling salesman problem: a computational study.
\newblock In \emph{The traveling salesman problem}. Princeton university press, 2011.

\bibitem[Rumelhart et~al.(1986)Rumelhart, Hinton, and Williams]{rumelhart1986learning}
D.~E. Rumelhart, G.~E. Hinton, and R.~J. Williams.
\newblock Learning representations by back-propagating errors.
\newblock \emph{Nature}, 323\penalty0 (6088):\penalty0 533--536, 1986.

\bibitem[Elman(1990)]{elman1990finding}
J.~L. Elman.
\newblock Finding structure in time.
\newblock \emph{Cognitive Science}, 14\penalty0 (2):\penalty0 179--211, 1990.

\bibitem[Gu and Dao(2024)]{gu2024mamba}
A.~Gu and T.~Dao.
\newblock Mamba: Linear-time sequence modeling with selective state spaces.
\newblock \emph{arXiv preprint arXiv:2312.00752}, 2024.

\bibitem[Vaswani et~al.(2017)Vaswani, Shazeer, Parmar, Uszkoreit, Jones, Gomez, Kaiser, and Polosukhin]{vaswani2017attention}
A.~Vaswani, N.~Shazeer, N.~Parmar, J.~Uszkoreit, L.~Jones, A.~N. Gomez, {\L}.~Kaiser, and I.~Polosukhin.
\newblock Attention is all you need.
\newblock In \emph{Advances in Neural Information Processing Systems}, volume~30, 2017.

\bibitem[Bettini et~al.(2022)Bettini, Kortvelesy, Blumenkamp, and Prorok]{bettini2022vmas}
M.~Bettini, R.~Kortvelesy, J.~Blumenkamp, and A.~Prorok.
\newblock {VMAS}: A vectorized multi-agent simulator for collective robot learning.
\newblock In \emph{International Symposium on Distributed Autonomous Robotic Systems}, pages 42--56. Springer, 2022.

\bibitem[Bou et~al.(2024)Bou, Bettini, Dittert, Kumar, Sodhani, Yang, De~Fabritiis, and Moens]{bou2024torchrl}
A.~Bou, M.~Bettini, S.~Dittert, V.~Kumar, S.~Sodhani, X.~Yang, G.~De~Fabritiis, and V.~Moens.
\newblock {TorchRL}: A data-driven decision-making library for {PyTorch}.
\newblock In \emph{International Conference on Learning Representations}, volume 2024, pages 1778--1811, 2024.

\end{thebibliography}

\newpage
\appendix

\section{AICON's background}\label{app:aicon}

AICON (Active InterCONnect)~\cite{mengers2025no} is the framework we use to instantiate the world factor $\theta_\mathrm{world}$. Beyond serving as the world side of the factorization, AICON has three properties that make it particularly well-suited for this role. First, its graph structure is compositional by construction: new world regularities can be added as nodes and edges without modifying existing components, which is what enables the zero-shot generalization demonstrated in Section~\ref{sec:results}. Second, it operates without task-specific data: the graph encodes known physical regularities directly, so $\theta_\mathrm{world}$ admits a structured prior with small volume $V_\mathrm{world}$ and can be evaluated independently of $\theta_\mathrm{task}$, satisfying the tractability condition of Section~\ref{sec:theory}. Third, its decentralized structure requires no global clock or centralized computation, admitting asynchronous execution across heterogeneous robots. This appendix provides technical background on its construction, complementing the description in Section~\ref{sec:instantiation}.

Formally, an Active Interconnection (AI) $j$ establishes a relationship between a set of estimates $\{\mathbf{x}_1, \dots, \mathbf{x}_{l_j}\}$, represented as an implicit differentiable function $\mathbf{h}_j(\bullet)$:
\begin{equation}\label{eq:ai}
    \mathbf{h}_j(\mathbf{x}_1^t, \mathbf{x}_2^t, \dots, \mathbf{x}_l^t) = \mathbf{0}.
\end{equation}
Equation \eqref{eq:ai} encodes a world regularity that must evaluate $\mathbf{0}$ if all the estimates are accurate. However, during the estimation process, $\mathbf{h}_j(\bullet)$ is usually a non-zero value. For readability, we will denote this residual as $\mathbf{h}_j^t$. The residual plays a dual role: it provides an informative prior for the connected REs and, through its derivatives, contributes the gradient signal that drives action selection. Because $h_j$ is evaluated on the current beliefs of the REs---whose uncertainties evolve with incoming observations---the information exchanged between nodes adapts over time. This is the sense in which the interconnections are \emph{active}, in contrast with the static constraints used by classical potential-field methods~\cite{khatib1986real}. Because each AI is defined only over its incident estimators, new regularities can be added to the graph without modifying existing ones.

A Recursive Estimator (RE) recursively computes its state estimate $\mathbf{x}_i^t$ based on its previous estimate $\mathbf{x}_i({t-1})$ and a set of $m_i$ informative priors $\{\mathbf{h}_1^t, \mathbf{h}_2^t, \dots, \mathbf{h}_{m_i}^t\}$ provided by its AIs:
\begin{equation}\label{eq:re}
    \mathbf{x}_i^t = \mathbf{f}(\mathbf{x}_i^{t-1}, \mathbf{h}_1^t, \mathbf{h}_2^t, \dots, \mathbf{h}_{m_i}^t),
\end{equation}
where $\mathbf{f}(\bullet)$ is a differentiable function representing probabilistic inference, typically instantiated as a Kalman Filter or one of its non-linear variants. This recursive structure allows each estimator to maintain a belief state while accounting for
sensory and process noise, and to recover quantities that are not directly observable by exploiting the regularities propagated by the AIs. Because each RE is updated using only the residuals of its incident AIs, estimation is inherently local and can proceed asynchronously across the graph.

Each cost function $g_k(\mathbf{x}^t)$ depends on a subset of the $M$ estimates of the graph, with \mbox{$\mathbf{x} = \{\mathbf{x}_1, \ldots, \mathbf{x}_M\}$}. The gradient of the cost is propagated back through the interconnections---via the chain rule across the differentiable functions $\mathbf{h}_j$ and $\mathbf{f}$---to determine the required change in actuation signals $\textbf{a}$, denoted $\nabla^\mathrm{p}_a g_{k}(x^{t})$. Action selection then follows Eq. \eqref{eq:aicon_og}: at each time step, AICON selects the gradient path of steepest descent among the set $\mathbb{P}$ of available paths. The union of these paths implicitly defines a world-grounded yet task-specific manifold on the estimated state, which is the object that our learned policy subsequently modulates.

A distinctive property of the construction above is that the resulting optimization landscape is \emph{dynamic}: as the beliefs of the REs evolve, so do the residuals $\mathbf{h}_j$ and their derivatives, producing a time-varying potential field that continuously adapts to the current estimated state of the world. This contrasts with static energy-based formulations and with learned Riemannian metrics, which encode the geometry of the task once and for all. The time-varying landscape is the mechanism by which behaviors such as interactive perception, error recovery, and reactive re-targeting emerge without explicit planning: whenever the state of the world changes, the gradient paths reshape accordingly, and the steepest-descent rule of Eq.~\eqref{eq:aicon_og} produces the corresponding behavior.

The graph structure of AICON does not require a global clock, a centralized computer, nor a single action-selection process. Each RE only needs access to the residuals of its incident AIs, and each gradient path is defined locally by differentiating along the subset of
the graph that connects a cost to an actuation signal. As a consequence, AICON admits decentralized execution across heterogeneous robots and asynchronous updates driven by the availability of sensor measurements. 

\section{Policy details}\label{app:policy_details}
\subsection{2D Search and Pressure Plate Tasks}

The policy $\Pi({y}_r^t,\theta_{\mathrm{task}})$ maps each robot's local observation and the available AICON gradient paths to two outputs: (i) a discrete \emph{gradient-path selector} $\Xi^t_r$ that chooses which goal gradient the robot should follow, and (ii) a scalar \emph{gain} $\kappa^t_r$ that modulates the magnitude of the selected gradient. The architecture is designed to be \emph{permutation-invariant} with respect to both the number of robots and the number of gradient paths, so that a single trained model generalizes to team sizes and graph topologies not seen during training.

At every time step each robot~$r$ receives a feature vector composed of five blocks:
\begin{enumerate}
    \item \textbf{Robot identifier}.  A normalized index $\text{id}_r = (2r+1)/(2N)$ that provides a unique, order-agnostic label in $(0,1)$.  The identifiers are \emph{randomly permuted} at the start of every episode so that the policy cannot memorize a fixed mapping between identifiers and roles; instead it must learn to coordinate based on positions.

    \item \textbf{Proprioceptive state}.  The robot's own position~$(x,y)$ and velocity~$(v_x,v_y)$.

    \item \textbf{Shared context}.  Global task state visible to all robots. In pressure plate, it encompasses the positions and status of both pressure plates, the door position and open/closed flag, and the goal position.  In the cooperative-search task the shared context instead contains the ground-robot position, the occupancy and target grids (flattened), and a sensor-noise scalar.

    \item \textbf{Teammate state}.  Positions and velocities of the neighboring robots. 

    \item \textbf{AICON gradient features}. For each of the~$\mathbb{P}$ gradient paths (goals) in the AICON graph the first two components of the Jacobian $\nabla^{\mathrm{p}}_{\mathbf{a}} g_k$ are included, providing the direction of steepest cost decrease for each candidate sub-goal.
\end{enumerate}
The policy processes the stacked robot features through a Transformer encoder followed by two task-specific heads:
\begin{itemize}
    \item \textbf{Robot branch.}  A two-hidden-layer MLP ($d_{\text{in}} \to 64 \to 64$, ReLU) encodes each robot's feature vector independently.

    \item \textbf{Context branch.}  A separate MLP ($11 \to 128 \to 64$, ReLU) encodes the shared context into a 64-dim vector that is broadcast to every robot slot.

    \item \textbf{Fusion.}  The robot and context embeddings are concatenated (128-dim) and projected to a hidden dimension $d_h=128$ via a linear layer.

    \item \textbf{Self-attention.}  A single \texttt{TransformerEncoderLayer} (4~heads, feed-forward dim 256, no dropout) with a key-padding mask over inactive robot slots.  This layer allows robots to exchange information about each other's embeddings, enabling coordination.

    \item \textbf{Gradient-path selector} $\Xi_r$ (actor head).  A two-layer MLP ($128\!\to\!64\!\to\!T$, Tanh hidden) produces per-robot logits over the $T$~targets.  Actions are sampled from a \texttt{OneHotCategorical} distribution during training and selected greedily at evaluation.

    \item \textbf{Gain head} $\kappa_r$.  A two-layer MLP ($128 \to 64 \to 2$, Tanh hidden) outputs two scalars: a location and a raw scale parameter. The scale is passed through a softplus activation to ensure positivity. The final gain applied to the selected gradient path is $\kappa_r
   = \kappa_{\text{base}} + \Delta\kappa_r$, where $\kappa_{\text{base}}$ is a task-dependent constant and $\Delta\kappa_r$ is the prediction of the gain head. 

\end{itemize}

For the search task, where strong heuristics exist from the TSP literature, we also compare with \textbf{Max-Grad} (selects the target whose AICON gradient has the largest magnitude), and \textbf{TSP planner} and \textbf{TSP optimum} (TSP brute force methods, the former only considers the targets currently observed and the latter all of them from the start). For the pressure plate task, we compare with \textbf{KNN} (k-nearest-neighbors classifier).

\subsection{Handover Task}

The handover task uses the same principle as the 2-D tasks---a learned selector chooses among AICON gradient paths---but the action space is adapted to bimanual manipulation. We use a shared policy for both arms. At every time step each arm receives a feature vector composed of four blocks:

\begin{enumerate}
    \item \textbf{Robot identifier}. A scalar index $\text{id} \in \{0, 1\}$ providing a unique identifier for each robot, respectively. 
    
    \item \textbf{AICON graph state}. Local and teammate end-effector pose estimates, object and obstacle pose estimates and distances w.r.t. end-effectors poses, grasp likelihoods, and in-reach likelihoods.

    \item \textbf{AICON gradient features}. Per-candidate gradient features and their null spaces for the paths exposed by the handover graph.

    \item \textbf{Validity mask}.  A binary mask that suppresses unavailable candidate directions before action selection.
\end{enumerate}

The graph exposes seven candidate directions per arm: three real AICON gradient paths and four nullspace directions derived from the current steepest gradient. The joint selector receives the two arm-wise inputs and outputs an action in $\{1,\ldots,7\}^2$; the selected directions are converted into Cartesian end-effector displacement commands by the handover controller.

The handover selectors use small MLP architectures:
\begin{itemize}
    \item \textbf{RL selector}.  The two arm-wise feature vectors are flattened into a joint observation and passed through a two-hidden-layer MLP with width 64.  The actor outputs two categorical distributions, one per arm, over the seven candidate directions.  During training, one direction is sampled for each arm; at evaluation, the most likely direction is selected.

    \item \textbf{LD selector}. The supervised selector scores the seven candidate directions for one arm at a time using a compact two-hidden-layer MLP with width 32. Invalid candidates are masked before the softmax, and the same selector is applied independently to both arms.
\end{itemize}
Handover-specific training details are given in Appendix~\ref{app:training}.

\section{Environment details}\label{app:environments}

Search and pressure plate tasks are implemented in VMAS~\cite{bettini2022vmas}, a vectorized 2-D multi-agent simulator built on PyTorch.  The AICON graph for each task is constructed using batched Extended Kalman Filters (EKFs) as recursive estimators and differentiable cost functions as goals.

\subsection{Multi-robot heterogeneous search}

\begin{figure}
    \centering
    \includegraphics[width=0.5\linewidth]{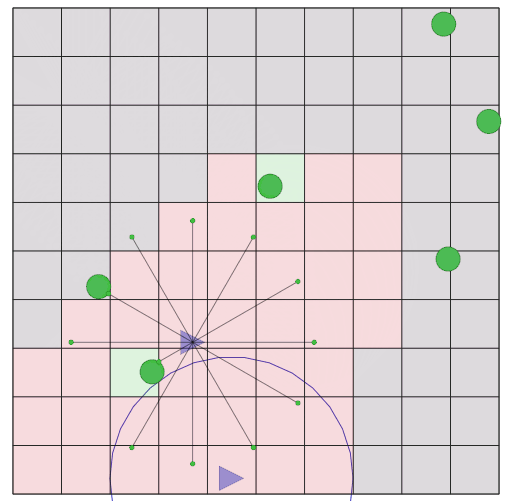}
    \caption{Example of the simulation environment for search. The ground robot is represented as a blue triangle, with LiDAR rays in black ended in a small green circle. The quadrotor is shown as a blue triangle surrounded by a blue circle that represents the observation radius. The targets are represented as green circles. The map shows a grid cell of the grid map estimate of the quadrotor, where gray means that the cell is unexplored, green means that the quadrotor found a target, and red means unoccupied.}
    \label{fig:examples_search}
\end{figure}

A 2-D bounded arena of size $[-1,1]^2$ contains $M$ static targets (default $M\!=\!5$) placed at random positions with a minimum inter-entity separation of 0.2\,m.  The team consists of one ground robot and one quadrotor. Fig.~\ref{fig:examples_search} shows an example of the simulation environment used to train the search task.

\paragraph{Agents.} The \emph{ground robot} (sphere, radius 0.1\,m) is the only agent capable of physically collecting targets; it is equipped with a LiDAR sensor (12~rays, range 0.3\,m) that returns per-ray distances and target-ID vectors.  The \emph{quadrotor} (sphere, radius 0.05\,m) is a non-colliding holonomic agent that maintains a downward-facing occupancy grid of size $G\!\times\! G$ (default $G\!=\!10$), discovering target locations when they fall within its sensing radius~$R_{\text{quadrotor}}=0.5$\,m.

\paragraph{Observations.}
The quadrotor observes its own position and velocity, the ground robot's position, and the current occupancy and target grids flattened to vectors of size~$G^2$.  It also receives a scalar indicating the sensor noise level.  The ground robot observes its own position and velocity along with the LiDAR distance and target-ID readings.

\paragraph{Reward.}
The team receives a collective reward of $+10$ for each target found by the ground robot and a bonus of $+100$ when all targets are cleared.  A time penalty of $-0.1$ is applied at each step to encourage efficiency.  An optional potential-based distance-shaping term rewards reduction in the shortest-path distance to unvisited targets, computed via brute-force TSP over the currently discovered set.

\paragraph{Termination.}
The episode ends when all~$M$ targets are collected or the maximum number of steps is reached.

\subsection{Collaborative handover with obstacles}

The task is implemented in robosuite using two single-arm Franka Emika Panda manipulators mounted in an opposed configuration around a table.  A graspable hammer object is initialized in the workspace of the first robot (the ``giver''), while the second robot (the ``receiver'') starts on the opposite side.  A static box obstacle is placed between the arms, so that the direct path from the giver to the receiver conflicts with obstacle avoidance.  The global objective is for the receiver to securely grasp the object.

Episodes run for at most 500 control steps at 20\,Hz.  Both arms use operational-space Cartesian position control with 3-D end-effector displacement commands and a binary gripper command.  The training distribution uses a hammer object sampled on the table within $x \in [-0.1,0.1]$\,m and $y \in [-0.05,0.05]$\,m.  The trained policies use one active box obstacle during training and in the in-distribution reference condition; its position and length are randomized around a nominal box between the two robots.

\paragraph{Agents.}
Both robots are controlled through Cartesian displacement commands applied to the end-effector positions.  The AICON graph tracks end-effector poses, gripper activations, object pose, grasp likelihoods, reach likelihoods, and distances to the obstacle.  The robosuite simulator executes the corresponding operational-space control command for each arm.

\paragraph{Observations.}
The AICON graph receives end-effector pose measurements from robosuite proprioception, object-pose measurements, axis-aligned obstacle bounding boxes, and gripper force magnitudes.  The learned selector does not observe raw images.  Its inputs are the graph states and candidate gradient features described in Appendix~\ref{app:policy_details}.

\paragraph{Reward.}
The RL selector uses a shaped reward with the following components:
\begin{itemize}
    \item \textbf{Giver progress:} reward for reaching, grasping, and lifting the object.
    \item \textbf{Transfer progress:} after lift-off, reward for moving the object to the transfer region and bringing the receiver to its grasp point.
    \item \textbf{Completion bonus:} terminal bonus when the receiver successfully grasps the object.
    \item \textbf{Obstacle penalty:} proximity and contact penalties for both arms.
\end{itemize}

\paragraph{Termination.}
The episode ends in success when the receiver grasps the object, or in failure when the 500-step limit is reached. 

\begin{figure}
    \centering
    \includegraphics[width=0.7\linewidth]{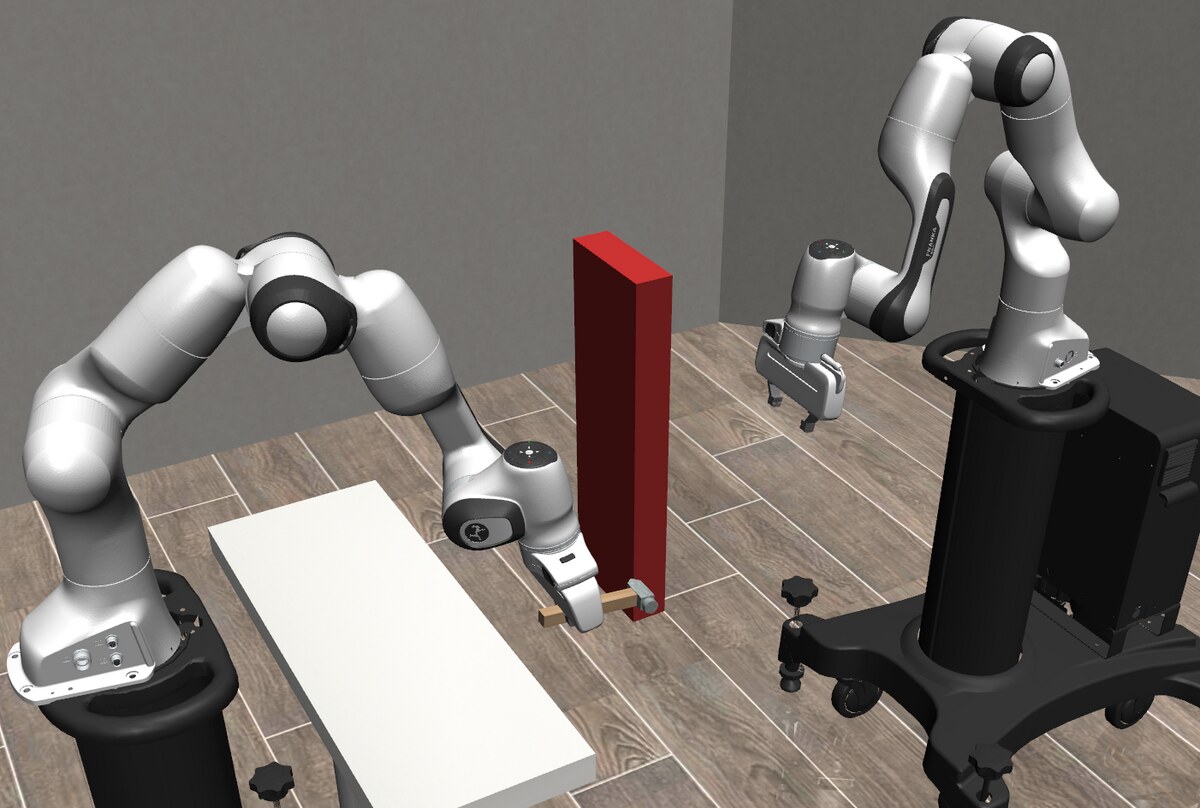}
    \caption{Example of the simulation environment for handover.}
    \label{fig:examples_handover}
\end{figure}

\subsection{Multi-robot pressure plate game}

\begin{figure}
    \centering
    \includegraphics[width=0.5\linewidth]{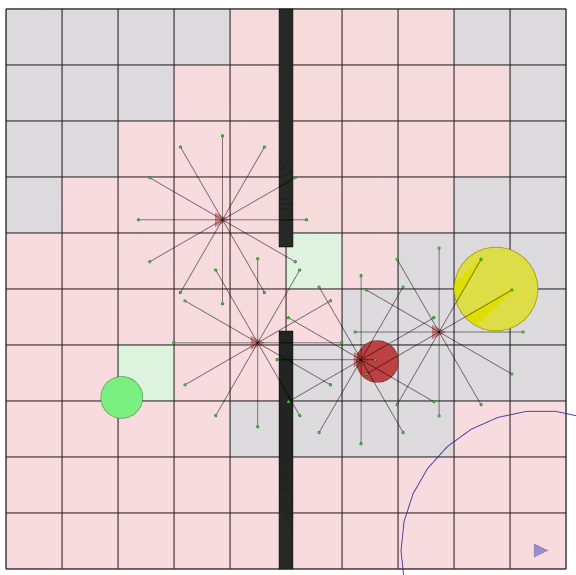}
    \caption{Example of the simulation environment for pressure plate. Ground robots are represented as a red triangles, with LiDAR rays in black ended in a small green circle. The quadrotor is shown as a blue triangle surrounded by a blue circle that represents the observation radius. The pressure plates are represented as green circles that turn red if they are being pressured by a ground robot. The goal is represented as a yellow circle. The map shows a grid cell of the grid map estimate of the quadrotor, where gray means that the cell is unexplored, green means that the quadrotor found a pressure plate, and red means unoccupied. The door is orange, but it does not appear in this example because one of the ground robots is pressuring the plate on the right.}
    \label{fig:examples_pp}
\end{figure}

The arena is a 2-D bounded space of size $[-2,2]^2$ divided into a left and right room by a central wall with a door gap.  Two pressure plates are randomly placed, one in each room.  The door opens only when a ground robot stands on \emph{either} pressure plate.  A goal zone (sphere, radius~0.3\,m) is located in the right room.  The mission is complete when all ground robots reach the goal. Fig.~\ref{fig:examples_pp} shows an example of the simulation environment used to train the pressure plate task.

\paragraph{Agents.}
$N$ ground robots (spheres, radius 0.1\,m, holonomic dynamics) are equipped with LiDAR sensors (12~rays, range 0.6\,m) that return distance measurements and a target-ID vector identifying pressure plates from walls.  One quadrotor (sphere, radius 0.05\,m, non-colliding) maintains a $G\!\times\!G$ occupancy grid ($G\!=\!10$) covering the full arena.

\paragraph{Observations.}
The quadrotor observes its own position and velocity, the positions of all ground robots, the door, the goal, and a local occupancy grid that indicates the location of pressure plates within its sensing radius~$R_{\text{quadrotor}}=1.0$\,m.  Each ground robot observes its own state, the positions of the door and goal, and LiDAR readings.  Unlike the search task, the LiDAR target-ID vector allows robots to distinguish pressure plates from walls and other obstacles.

\paragraph{AICON graph.}
The graph contains four goal cost functions per ground robot, each defining a gradient path toward one of the four targets: left plate, right plate, door, and goal.  Each cost function takes the form
\begin{equation}
    g_k(\mathbf{x}_{\text{rel}}, {\Sigma}_{\text{rel}}) = \left(|\|\mathbf{x}_{\text{rel}}\| - d_{\text{target}}| + \tfrac{1}{2}\|\mathbf{x}_{\text{rel}}\|^2 + 0.01\operatorname{tr}({\Sigma}_{\text{rel}})\right) \cdot \mathbb{1}_{\text{avail}},
\end{equation}
where $\mathbf{x}_{\text{rel}} \in \mathbb{R}^2$ is the EKF estimate of the robot-to-target relative position, $d_{\text{target}}=0$ is the desired distance, ${\Sigma}_{\text{rel}}$ is the associated covariance, and $\mathbb{1}_{\text{avail}}$ gates inactive targets.  The $\|\mathbf{x}_{\text{rel}}\|$ term provides a constant-magnitude unit-direction gradient, while the $\frac{1}{2}\|\mathbf{x}_{\text{rel}}\|^2$ term adds a distance-proportional push that improves robustness when the EKF Jacobian chain is noisy.  The door target is pinned to a position slightly past the door gap (offset 0.05\,m into the right room) so that the gradient pulls robots \emph{through} the opening rather than stopping at it.

\paragraph{Reward.}
The task uses a reward function with the following components:
\begin{itemize}
    \item \textbf{Time penalty:} $-0.01$ per step (shared).
    \item \textbf{Goal reaching:} $+10$ per robot that newly reaches the goal (shared).
    \item \textbf{Completion bonus:} $+20$ when all robots are simultaneously at the goal (shared).
    \item \textbf{Plate holding:} $+0.5$ per step for each robot actively pressing a plate at the appropriate stage of the task (shared, gated by the internal stage machine; see below).
    \item \textbf{Expert bonus (LD only):} $+0.3$ per step per robot whose gradient-path selection matches the expert's assignment; $0$ otherwise (asymmetric, per-robot).
\end{itemize}

\paragraph{Internal stage machine.}
The reward function maintains an internal task stage that tracks progress and gates the plate-holding reward:
\begin{itemize}
    \item \emph{Stage~0:} One robot presses the left plate while others navigate through the door.  Transition to Stage~1 occurs when the left plate is pressed and all other robots have crossed to the right side.
    \item \emph{Stage~1:} The right plate must be activated so the left-plate robot can cross.  Transition to Stage~2 occurs when the right plate has been pressed and the left-plate robot has crossed.
    \item \emph{Stage~2:} All robots navigate to the goal.
\end{itemize}

\paragraph{Termination.}
The episode terminates when all robots are simultaneously within the goal zone (radius 0.45\,m, accounting for robot radius and a small margin) or when the maximum number of steps is reached.

\section{Training details}\label{app:training}

\subsection{RL: Reinforcement learning for search and pressure-plate tasks}\label{app:training_rl}

We train our RL policy using Proximal Policy Optimization (PPO) with per-robot advantages in the style of Multi-Agent PPO (MAPPO).  All experiments use TorchRL \cite{bou2024torchrl} for environment wrapping, data collection, and replay-buffer management.

\paragraph{Hyperparameters.}
Table~\ref{tab:hyperparams} summarizes the training hyperparameters shared across tasks.

\begin{table}[ht]
\centering
\caption{PPO hyperparameters.}\label{tab:hyperparams}
\begin{tabular}{lc}
\toprule
\textbf{Parameter} & \textbf{Value} \\
\midrule
Learning rate (policy) & $3 \times 10^{-4}$ \\
Learning rate (critic) & $9 \times 10^{-4}$ \\
Discount factor $\gamma$ & 0.99 \\
GAE parameter $\lambda$ & 0.95 \\
PPO clip ratio $\epsilon$ & 0.2 \\
Entropy coefficient & $10^{-3}$ \\
Critic loss coefficient & 0.5 \\
Epochs per batch & 5 \\
Minibatch size & 400 \\
Frames per batch & 6400 \\
Max episode steps & 400 \\
Gradient norm clip & 1.0 \\
AICON gain $\kappa$ & 100 \\
Expert bonus (LD) & 0.3 \\
\bottomrule
\end{tabular}
\end{table}

\paragraph{Per-robot advantages (MAPPO-style).}
Although the critic produces a single centralized value~$V(s)$, we compute \emph{per-robot} advantages.  The reward tensor has shape $[B, N_{\max}]$ (one scalar per robot per step), and GAE is computed independently along each robot dimension using the shared value baseline expanded to match:
\begin{equation}
    A^{t}_r = \sum_{\ell=0}^{T-t-1} (\gamma\lambda)^\ell \delta^{t+\ell}_{r}, \quad
    \delta^{t}_r = R^{t}_r + \gamma V(s^{t+1})(1-d^t) - V(s^t),
\end{equation}
where $R^{t}_r$ is robot~$r$'s reward at time~$t$ and $d^t$ is the shared done flag.  The value target for the critic is the mean return across robots: $\hat{V}^t = \frac{1}{N}\sum_r (A^{t}_r + V(s^t))$.

\paragraph{Per-robot PPO ratio.}
Each robot's log-probability is stored independently during rollout collection.  During the PPO update, each robot gets its own importance-sampling ratio:
\begin{equation}
    \rho_{r} = \frac{\pi_{\theta_{\mathrm{task}}}(a^{t}_r \mid s^t)}{\pi_{\theta_{\mathrm{task},\text{old}}}(a^{t}_r \mid s^t)},
\end{equation}
and the clipped surrogate loss is computed per robot and summed:
\begin{equation}
    \mathcal{L}_{\text{policy}} = -\frac{1}{B}\sum_{b}\sum_{r} \min\!\Big(\rho_{r} A_{r},\; \text{clip}(\rho_{r}, 1\!-\!\epsilon, 1\!+\!\epsilon)\, A_{r}\Big).
\end{equation}
This gives each robot its own credit-assignment signal rather than sharing a single ratio across the team.  Advantages are normalized (zero-mean, unit-variance) across the batch with masking for inactive robot slots.

\paragraph{Robot/Target count variation.}
During training, each batch contains environments with different numbers of ground robots (pressure plate, $N \in \{2, 3, 4\}$) or targets (search, $T \in \{2, 3, 4, 5\}$).  This exposes the policy to varying team sizes and encourages learning count-invariant coordination strategies.

\paragraph{AICON warmup.}
At the start of each episode, 5 ``warmup'' steps are executed with zero ground-robot gains and a high drone gain ($\kappa_{\text{quadrotor}}=50$), allowing the EKF estimates to converge before the policy takes control.  Episode counters are reset after warmup so that stage-success detection is not contaminated.

\subsection{RL: Reinforcement learning for the handover }\label{app:training_rl_handover}

The handover RL policy is trained separately as a joint selector over the two robot arms.  Unlike the 2-D tasks, it is not trained with the TorchRL/MAPPO pipeline, does not use per-robot advantages, and does not learn a scalar gain head.  Instead, PPO receives the flattened two-arm selector observation, outputs one categorical gradient-candidate choice per arm, and estimates a single value for the joint handover state.  The selected candidates are then smoothed and scaled by a fixed Cartesian controller.  Table~\ref{tab:handover_rl_settings} lists the handover-specific settings. The pipeline is implemented in Stable-Baselines3.

\begin{table}[ht]
\centering
\caption{Handover-specific PPO settings.}\label{tab:handover_rl_settings}
\begin{tabular}{lc}
\toprule
\textbf{Parameter} & \textbf{Value} \\
\midrule
Parallel environments & 8 \\
Environment steps & $5\times 10^6$ \\
Learning rate & $3 \times 10^{-4}$ \\
Discount factor $\gamma$ & 0.99 \\
GAE parameter $\lambda$ & 0.95 \\
PPO clip ratio $\epsilon$ & 0.2 \\
Entropy coefficient & 0.005 \\
Critic loss coefficient & 0.5 \\
Epochs per batch & 5 \\
Minibatch size & 400 \\
Gradient norm clip & 1.0 \\
\bottomrule
\end{tabular}
\end{table}

\paragraph{End-to-end handover baselines.}
The MLP, RNN, Mamba, and Transformer baselines for handover are trained as direct-action PPO policies on flattened robosuite observations.  Their observation contains the object state and both robots' proprioceptive states, and their action is the full Cartesian displacement and gripper command for both arms. All baseline are trained using the same reward function and termination conditions as our RL and LD selectors.

\paragraph{Handover evaluation conditions.}
In-distribution evaluation uses the same one-obstacle, hammer-object distribution as training.  OOD evaluation varies one factor at a time, as summarized in Table~\ref{tab:handover_eval_conditions}. No OOD condition retrains the learned selector or the AICON graph.

\begin{table}[ht]
\centering
\caption{Handover evaluation conditions.}\label{tab:handover_eval_conditions}
\begin{tabular}{@{}p{0.28\linewidth}p{0.62\linewidth}@{}}
\toprule
\textbf{Condition} & \textbf{Variation} \\
\midrule
In distribution & One box obstacle, hammer object \\
Robot yaw & Per-reset yaw perturbation in $[-0.5,0.5]$\,rad per robot \\
Obstacle & Obstacle position shifted outside the training support, or larger obstacle footprint \\
Object pose & Object pickup position shifted outside the training support along $x$ or $y$ \\
Object type & Cylinder, hollow cylinder, or ratcheting wrench \\
Number of obstacles & Aactivates one to four obstacles: the reference obstacle, then side obstacles on the right and left of the workspace, and finally an overhead obstacle. For all obstacles position and size are randomized. \\
\bottomrule
\end{tabular}
\end{table}

\subsection{LD: Learning from demonstrations}\label{app:training_ld}

Our LD policy replaces the reward-based PPO signal with a supervised expert-bonus that provides a dense, per-robot teaching signal at every step.

\paragraph{Expert oracles for search and pressure-plate tasks.}
An analytical expert assigns each robot to one of the $P$ gradient paths based on the current state.  For the pressure-plate task, the expert considers robot positions relative to the door (left/right side), distances to each plate, and whether the door is currently open.  The assignment logic is:
\begin{itemize}
    \item \emph{All robots on the right side:} all assigned to Goal.
    \item \emph{Mixed (some left, some right):} the closest right-side robot is assigned to PlateR; remaining right-side robots to Goal.  If the door is closed, the closest left-side robot is assigned to PlateL (to open the door); if the door is open, left-side robots are assigned to Door (to cross).
    \item \emph{All robots on the left side:} if the door is closed, the closest robot to PlateL is assigned to PlateL, others to Door; if the door is open, all assigned to Door.
\end{itemize}
For the search scenario, the expert follows the target selected by a brute-force TSP solver.

\paragraph{Demonstrations for handover task.}
For handover, expert demonstrations are generated by a scripted state machine rather than by a symbolic target-assignment oracle.  The giver is commanded to approach and grasp the object, then moves to a sampled handover point beside the obstacle while the receiver approaches the same region.  Near the transfer point, both arms are commanded to reduce the distance between the object and the receiver's end-effector. Demonstration actions are converted into soft labels by finding the non-negative mixture of candidate gradient directions that best reconstructs the demonstrated Cartesian direction. The supervised selector is trained with cross-entropy on those labels.

\paragraph{Supervised baselines.}
For comparison, we also train standalone supervised selectors on expert demonstration data collected offline (described in Section \ref{sec:results}).  Each $(robot, timestep)$ pair becomes one sample with input features (global context + per-target derivative features) and a label (expert's target assignment). All learned models are trained with cross-entropy loss, class-frequency weighting, Adam ($\text{lr}=10^{-3}$), cosine-annealing schedule, and an 80/20 episode-level train/validation split over 50 epochs.

\section{AICON graphs}\label{app:aicon_graphs}

Each graph contains five kinds of nodes:
\begin{description}
  \item[\textbf{Sensors} (yellow).]
        Compute noisy measurements of world quantities from raw simulator
        observations.
  \item[\textbf{Recursive estimators} (blue).]
        Maintain a recursive belief (mean + covariance) over a single world
        quantity. Their posteriors are the inputs to actions and goals.
  \item[\textbf{Active interconnections} (green).]
        Differentiable couplings that route quantities and cost gradients
        between estimators, sensors, and actions. They are the edges along
        which the world model propagates.
  \item[\textbf{Actions} (red).]
        Emit a low-level command. Their gradient flow back through the
        graph is what the learned task policy modulates.
  \item[\textbf{Specified goals} (pink).]
        Scalar cost functions over estimator means whose gradients are
        injected into the graph.
\end{description}

\subsection{Cooperative search}

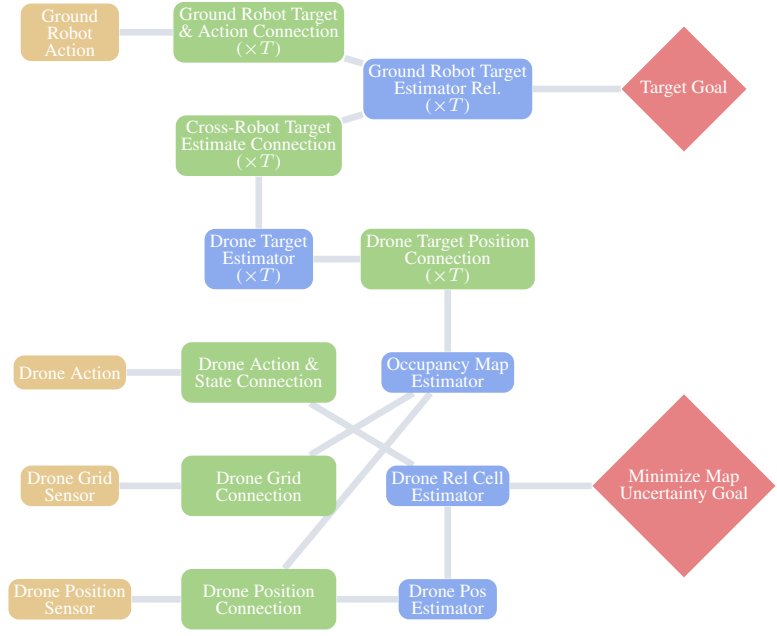
\begin{figure}[t]
\centering
\begin{tikzpicture}[x=2.5cm, y=1.5cm]
 
  \pgfdeclarelayer{bg}
  \pgfsetlayers{bg,main}
 
  \node[aicon/sensor]  at (0,0) (dpos_s)   {Drone Position\\Sensor};
  \node[aicon/sensor]  at (0,1) (dgrid_s)  {Drone Grid\\Sensor};
  \node[aicon/action]  at (0,2) (dr_a)     {Drone Action};
 
  \node[aicon/active]  at (1,0) (dpos_c)   {Drone Position\\Connection};
  \node[aicon/active]  at (1,1) (dgrid_c)  {Drone Grid\\Connection};
  \node[aicon/active]  at (2,3) (dtgt_c)   {Drone Target Position\\Connection\\($\times T$)};
  \node[aicon/active]  at (1,2) (dact_c)   {Drone Action \&\\State Connection};
  \node[aicon/active]  at (1,4) (cross_c)  {Cross-Robot Target\\Estimate Connection\\($\times T$)};
  \node[aicon/active]  at (1,5) (gr_act_c) {Ground Robot Target\\\& Action Connection\\($\times T$)};
 
  \node[aicon/state]   at (2,0) (dpos_e)   {Drone Pos\\Estimator};
  \node[aicon/state]   at (2,1) (drel_e)   {Drone Rel Cell\\Estimator};
  \node[aicon/state]   at (2,2) (omap)     {Occupancy Map\\Estimator};
  \node[aicon/state]   at (1,3) (dtgt_e)   {Drone Target\\Estimator\\($\times T$)};
  \node[aicon/state]   at (2,4.5) (gr_rel)   {Ground Robot Target\\Estimator Rel.\\($\times T$)};
 
  \node[aicon/action]  at (0,5) (gr_a)     {Ground\\Robot\\Action};
 
  \node[aicon/goal]    at (3.25,1)(gmap)    {Minimize Map\\Uncertainty Goal};
  \node[aicon/goal]    at (3.25,4.5)  (gtgt)    {Target Goal};
 
  \begin{pgfonlayer}{bg}
    \draw[aicon/link] (dpos_s)  -- (dpos_c);
    \draw[aicon/link] (dgrid_s) -- (dgrid_c);
    \draw[aicon/link] (dpos_c)  -- (dpos_e);
    \draw[aicon/link] (dpos_c)  -- (omap);
    \draw[aicon/link] (dgrid_c) -- (omap);
    \draw[aicon/link] (dpos_e)  -- (drel_e);
    \draw[aicon/link] (drel_e)  -- (gmap);
    \draw[aicon/link] (omap)    -- (dtgt_c);
    \draw[aicon/link] (dtgt_c)  -- (dtgt_e);
    \draw[aicon/link] (dr_a)    -- (dact_c);
    \draw[aicon/link] (dact_c)  -- (drel_e);
    \draw[aicon/link] (dtgt_e)  -- (cross_c);
    \draw[aicon/link] (cross_c) -- (gr_rel);
    \draw[aicon/link] (gr_rel)  -- (gtgt);
    \draw[aicon/link] (gr_rel)  -- (gr_act_c);
    \draw[aicon/link] (gr_act_c)-- (gr_a);
  \end{pgfonlayer}
 
\end{tikzpicture}
\caption{AICON graph for the cooperative search task, used by the RL, LD and AICON policies.}
\label{fig:aicon_graph_cs}
\end{figure}

The AICON graph used for the RL, LD and AICON policies of Section \ref{sec:results} is represented in Fig.~\ref{fig:aicon_graph_cs}. The specific implementation of each of the modules is the following:

\subsubsection{Sensors:}
\begin{description}
  \item[\textbf{Drone Position Sensor}.]
        Noisy absolute position of the drone.
  \item[\textbf{Drone Grid Sensor}.]
        Which occupancy cells the drone perceives this step
        (within \textit{drone\_obs\_range}).
  \item[\textbf{Ground Robot LiDAR}.]
        Per-ray distance-to-obstacle measurements from each ground robot.
\end{description}

\subsubsection{Estimators:}
\begin{description}
  \item[\textbf{Drone Pos Estimator}.]
        Recursive estimate of the drone's absolute position.
  \item[\textbf{Drone Rel Cell Estimator}.]
        The grid cell currently occupied by the drone (function of its
        position estimate).
  \item[\textbf{Occupancy Map Estimator}.]
        Global belief over the explored/unexplored occupancy grid, updated
        from drone observations.
  \item[\textbf{Drone Target Estimator}.]
        Drone-side estimate of each target's absolute position from
        accumulated grid observations.
  \item[\textbf{Ground Robot Target Estimator (Relative)}.]
        Per-robot estimate of each target's position in the robot's local
        frame.
\end{description}

\subsubsection{Active interconnections:}
\begin{description}
  \item[\textbf{Drone Action Connection / Drone State Connection}.]
        Propagate the drone's velocity action and resulting state into the
        world graph.
  \item[\textbf{Drone Position Connection}.]
        Binds drone-position sensor measurements to the drone position
        estimator.
  \item[\textbf{Drone Grid Connection / Drone Grid Estimate Connection}.]
        Feed raw grid observations and the maintained occupancy estimate
        between the drone sensor and the occupancy map estimator.
  \item[\textbf{Drone Target Position Connection}]
        Couples occupancy-map evidence to each drone-side target
        estimate.
  \item[\textbf{Cross Robot Target Estimate Connection}]
        Bridges each drone-side target estimate with the corresponding
        ground robot's relative target estimate. This is the channel
        through which cooperation is realized: information observed by the
        drone reaches each robot's local frame as a gradient.
  \item[\textbf{Ground Robot Abs2Rel Connection}]
        Differentiable change of frame, world $\rightarrow$ robot-local.
  \item[\textbf{Ground Robot Target Position Connection}]
        Routes the robot-local target estimate to the robot's action.
  \item[\textbf{Ground Robot Action Connection}]
        Binds the ground robot velocity action into the world graph.
\end{description}

\subsubsection{Goals:}
\begin{description}
  \item[\textbf{Minimize Map Uncertainty Goal}.]
        $-\sum_i p_i \log p_i$ over occupancy-map cells. Its gradient
        drives the drone toward unexplored regions.
  \item[\textbf{Fixed Distance To Target Goal}.]
        $\lVert \hat{t}_i - \mathbf{x}_r \rVert - d^\ast$ for each target;
        its gradient pushes the ground robot to a fixed approach distance
        from each target.
\end{description}

\subsection{Bimanual handover}
The handover graph is built around the two end-effector displacement actions and two gripper actions.  It estimates the object pose, each end-effector pose, the distance from each end-effector to the object, whether each gripper has grasped the object, whether the object is in reach of each arm, and the distance from each end-effector to the active obstacle AABBs.  These estimates define the candidate gradient paths used by the RL and LD policies.

\subsubsection{Sensors:}
\begin{description}
  \item[\textbf{End-Effector Pose Sensor}.]
        Measures the Cartesian end-effector position of each arm from robosuite proprioception.
  \item[\textbf{Object Pose Sensor}.]
        Measures the 3-D object position used by the object-pose estimator.
  \item[\textbf{Obstacle AABB Sensor}.]
        Publishes the center and half-size of each active obstacle box, padded to a maximum of four boxes.
\end{description}

\subsubsection{Estimators:}
\begin{description}
  \item[\textbf{End-Effector Pose Estimator}.]
        EKF estimate of each arm's 3-D end-effector position, updated from direct proprioception and the previous displacement action.
  \item[\textbf{Object Pose Estimator}.]
        Shared EKF estimate of the object position, updated from direct object-pose measurements and switching couplings to either end-effector when a grasp is likely.
  \item[\textbf{End-Effector--Object Distance Estimator}.]
        Per-arm estimate of distance to the object grasp point.
  \item[\textbf{Object Grasped Likelihood Estimator}.]
        Per-arm scalar likelihood that the object is grasped, driven by distance-to-object, gripper activation, and force evidence.
  \item[\textbf{Object In-Reach Likelihood Estimator}.]
        Per-arm reach likelihood from the end-effector--object distance.
  \item[\textbf{End-Effector--Obstacle Distance Estimator}.]
        Per-arm smoothed distance to the nearest obstacle AABB, computed by Gaussian-weighted samples around the end-effector.
  \item[\textbf{Obstacle Proximity Likelihood Estimator}.]
        Per-arm scalar likelihood that the end-effector is near the obstacle, used to gate action-gradient flow near obstacles.
\end{description}

\subsubsection{Active interconnections:}
\begin{description}
  \item[\textbf{Forward Kinematics / End-Effector Displacement Connection}.]
        Couples the Cartesian displacement action to the end-effector pose estimator and gates the action-gradient path by object reach and obstacle proximity.
  \item[\textbf{Direct Measurement Connection}.]
        Binds measured end-effector position to the corresponding end-effector pose estimator.
  \item[\textbf{Object Direct Measurement Connection}.]
        Binds measured object position to the shared object pose estimator.
  \item[\textbf{End-Effector--Object Distance Connection}.]
        Computes the distance from each end-effector to the corresponding object grasp point.
  \item[\textbf{Object Grasped Connection}.]
        Updates grasp likelihood and routes gripper gradients when the gripper is close enough to close on the object.
  \item[\textbf{End-Effector--Object Coupling Connection}.]
        Couples the object estimate to a gripper pose when that gripper is likely to hold the object.
  \item[\textbf{Obstacle Distance and Proximity Connections}.]
        Compute smoothed distances to all active AABBs and convert them into a proximity likelihood for action gating.
\end{description}

\subsubsection{Goals:}
\begin{description}
  \item[\textbf{GoalGraspedObject}.]
        Active on the receiver; minimizes $1-p_{\mathrm{grasp}}$ so the receiver closes on the transferred object.
  \item[\textbf{GoalAvoidObstacle}.]
        Active for both arms; penalizes small end-effector--obstacle distances.
\end{description}

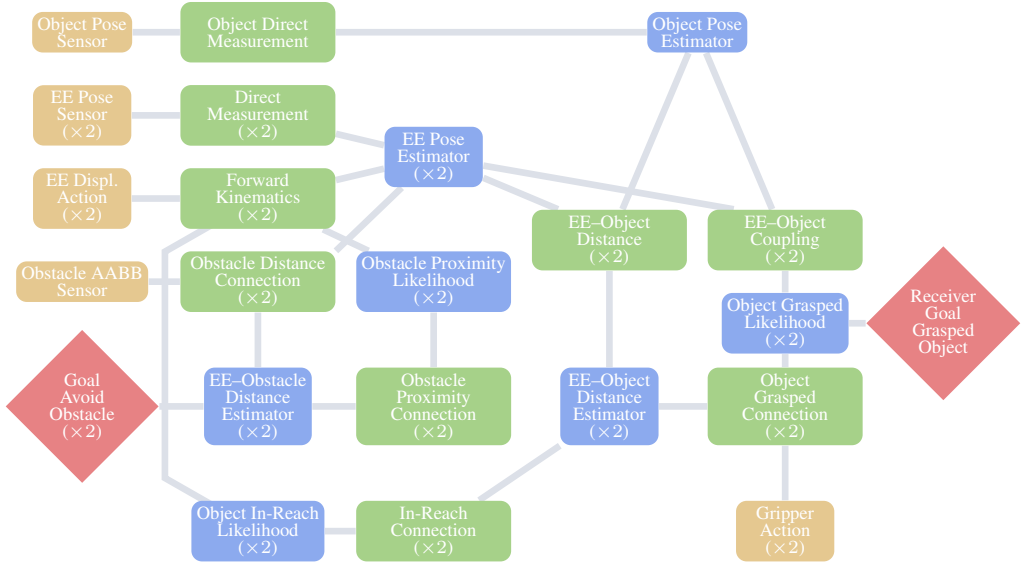
\begin{figure}[t]
\centering
\begin{tikzpicture}[x=1.55cm, y=1.10cm]

  \pgfdeclarelayer{bg}
  \pgfsetlayers{bg,main}

  \node[aicon/sensor] at (0,5.0) (ee_s)
    {EE Pose\\Sensor\\($\times 2$)};

  \node[aicon/action] at (0,4.0) (ee_a)
    {EE Displ.\\Action\\($\times 2$)};

  \node[aicon/sensor] at (0,3.0) (obs_s)
    {Obstacle AABB\\Sensor};

  \node[aicon/sensor] at (0,6) (obj_s)
    {Object Pose\\Sensor};

  \node[aicon/action] at (6,0) (grip_a)
    {Gripper\\Action\\($\times 2$)};

  \node[aicon/active] at (1.5,5.0) (ee_meas_c)
    {Direct\\Measurement\\($\times 2$)};

  \node[aicon/active] at (1.5,4.0) (fk_c)
    {Forward\\Kinematics\\($\times 2$)};

  \node[aicon/active] at (1.5,3) (obs_dist_c)
    {Obstacle Distance\\Connection\\($\times 2$)};

  \node[aicon/active] at (1.5,6) (obj_meas_c)
    {Object Direct\\Measurement};

  \node[aicon/active] at (6,1.5) (grasp_c)
    {Object\\Grasped\\Connection\\($\times 2$)};

  \node[aicon/state] at (3,4.5) (ee_e)
    {EE Pose\\Estimator\\($\times 2$)};

  \node[aicon/active] at (4.5,3.5) (ee_obj_c)
    {EE--Object\\Distance\\($\times 2$)};

  \node[aicon/state] at (5.25,6) (obj_e)
    {Object Pose\\Estimator};

  \node[aicon/state] at (4.5,1.5) (ee_obj_e)
    {EE--Object\\Distance\\Estimator\\($\times 2$)};

  \node[aicon/active] at (3,1.5) (prox_c)
    {Obstacle\\Proximity\\Connection\\($\times 2$)};

  \node[aicon/state] at (1.5,1.5) (obs_dist_e)
    {EE--Obstacle\\Distance\\Estimator\\($\times 2$)};

  \node[aicon/active] at (6,3.5) (obj_couple_c)
    {EE--Object\\Coupling\\($\times 2$)};

  \node[aicon/active] at (3,0) (reach_c)
    {In-Reach\\Connection\\($\times 2$)};

  \node[aicon/state] at (3,3) (prox_e)
    {Obstacle Proximity\\Likelihood\\($\times 2$)};

  \node[aicon/state] at (6,2.5) (grasp_e)
    {Object Grasped\\Likelihood\\($\times 2$)};

  \node[aicon/state] at (1.5,0) (reach_e)
    {Object In-Reach\\Likelihood\\($\times 2$)};

  \node[aicon/goal] at (0,1.5) (avoid_g)
    {Goal\\Avoid\\Obstacle\\($\times 2$)};

  \node[aicon/goal] at (7.35,2.5) (grasp_g)
    {Receiver\\Goal\\Grasped\\Object};

  \begin{pgfonlayer}{bg}

    \draw[aicon/link] (ee_s) -- (ee_meas_c);
    \draw[aicon/link] (ee_meas_c) -- (ee_e);

    \draw[aicon/link] (ee_a) -- (fk_c);
    \draw[aicon/link] (fk_c) -- (ee_e);

    \draw[aicon/link] (obj_s) -- (obj_meas_c);
    \draw[aicon/link] (obj_meas_c) -- (obj_e);

    \draw[aicon/link] (ee_e) -- ([xshift=10]ee_obj_c.north west);
    \draw[aicon/link] (obj_e) -- (ee_obj_c);
    \draw[aicon/link] (ee_obj_c) -- (ee_obj_e);

    \draw[aicon/link] (grip_a) -- (grasp_c);
    \draw[aicon/link] (ee_obj_e) -- (grasp_c);
    \draw[aicon/link] (grasp_c) -- (grasp_e);
    \draw[aicon/link] (grasp_e) -- (grasp_g);

    \draw[aicon/link] (ee_obj_e.south west) -- (reach_c);
    \draw[aicon/link] (reach_c) -- (reach_e);

    \draw[aicon/link]
      (reach_e.center) -- ([xshift=-35.0,yshift=20.0]reach_e.center) -- ([xshift=-35.0,yshift=-20.0]fk_c.center) -- (fk_c.center);

    \draw[aicon/link] (grasp_e) -- (obj_couple_c);
    \draw[aicon/link] (ee_e) -- ([xshift=10]obj_couple_c.north west);
    \draw[aicon/link] (obj_couple_c) -- (obj_e);

    \draw[aicon/link] (obs_s) -- (obs_dist_c);
    \draw[aicon/link] (ee_e) -- (obs_dist_c.north east);
    \draw[aicon/link] (obs_dist_c) -- (obs_dist_e);
    \draw[aicon/link] (obs_dist_e) -- (avoid_g);

    \draw[aicon/link] (obs_dist_e) -- (prox_c);
    \draw[aicon/link] (prox_c) -- (prox_e);

    \draw[aicon/link]
      (prox_e)
      --
      (fk_c);

  \end{pgfonlayer}

\end{tikzpicture}
\caption{AICON graph for the bimanual handover task, used by the RL, LD and AICON policies. Per-arm components are aggregated with $\times 2$.}
\label{fig:aicon_graph_handover}
\end{figure}

\subsection{Pressure plate}

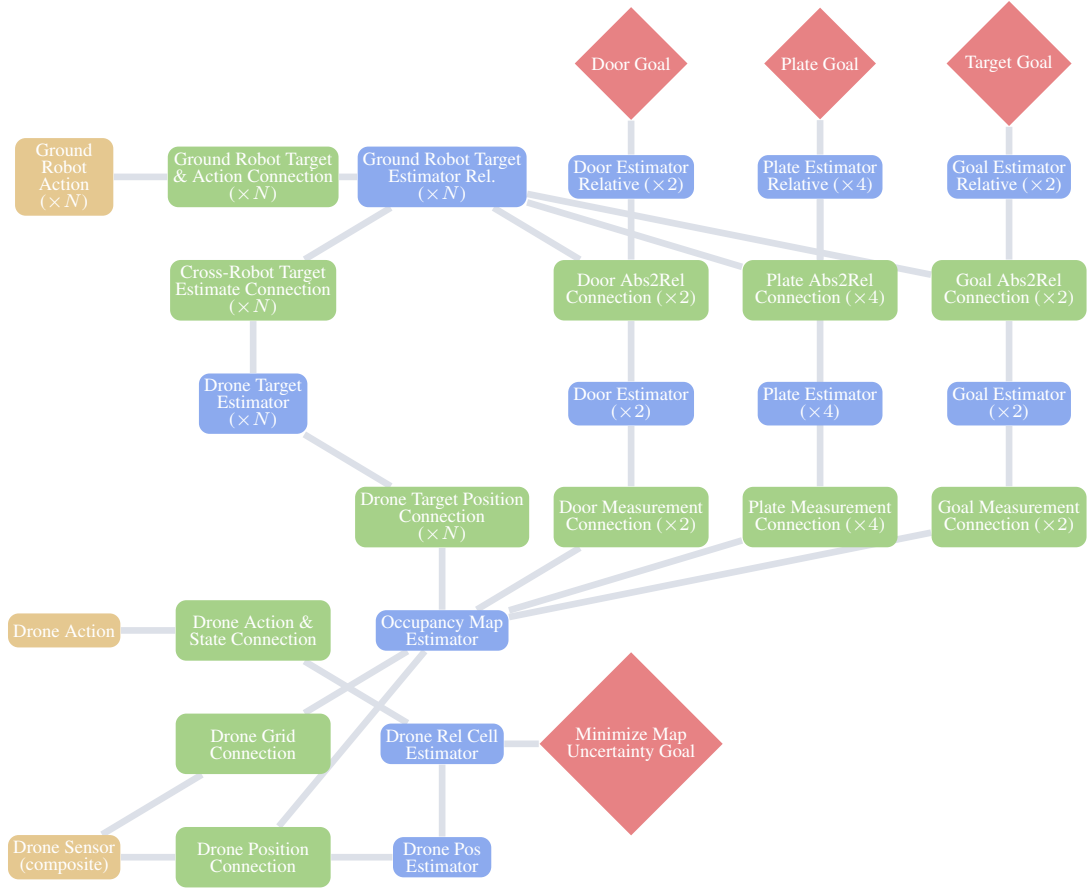
\begin{figure}[t]
\centering
\begin{tikzpicture}[x=2.5cm, y=1.5cm]
 
  \pgfdeclarelayer{bg}
  \pgfsetlayers{bg,main}
 
  \node[aicon/sensor]  at (0,-1) (dns)      {Drone Sensor\\(composite)};
  \node[aicon/action]  at (0,1) (dr_a)     {Drone Action};
 
  \node[aicon/active]  at (1,-1) (dpos_c)   {Drone Position\\Connection};
  \node[aicon/active]  at (1,0) (dgrid_c)  {Drone Grid\\Connection};
  \node[aicon/active]  at (2,2) (dtgt_c)   {Drone Target Position\\Connection\\($\times N$)};
  \node[aicon/active]  at (1,1) (dact_c)   {Drone Action \&\\State Connection};
  \node[aicon/active]  at (1,4) (cross_c)  {Cross-Robot Target\\Estimate Connection\\($\times N$)};
  \node[aicon/active]  at (1,5) (gr_act_c) {Ground Robot Target\\\& Action Connection\\($\times N$)};
 
  \node[aicon/state]   at (2,-1) (dpos_e)   {Drone Pos\\Estimator};
  \node[aicon/state]   at (2,0) (drel_e)   {Drone Rel Cell\\Estimator};
  \node[aicon/state]   at (2,1) (omap)     {Occupancy Map\\Estimator};
  \node[aicon/state]   at (1,3) (dtgt_e)   {Drone Target\\Estimator\\($\times N$)};
  \node[aicon/state]   at (2,5) (gr_rel)   {Ground Robot Target\\Estimator Rel.\\($\times N$)};
 
  \node[aicon/action]  at (0,5) (gr_a)     {Ground\\Robot\\Action\\ ($\times N$)};
 
  \node[aicon/active]  at (3,2) (door_mc)  {Door Measurement\\Connection ($\times 2$)};
  \node[aicon/active]  at (3,4) (door_a2r) {Door Abs2Rel\\Connection ($\times 2$)};
  \node[aicon/active]  at (5,2) (goal_mc)  {Goal Measurement\\Connection ($\times 2$)};
  \node[aicon/active]  at (5,4) (goal_a2r) {Goal Abs2Rel\\Connection ($\times 2$)};
  \node[aicon/active]  at (4,2) (plate_mc) {Plate Measurement\\Connection ($\times 4$)};
  \node[aicon/active]  at (4,4) (plate_a2r) {Plate Abs2Rel\\Connection ($\times 4$)};

  \node[aicon/state]   at (3,3) (door_e)   {Door Estimator\\($\times 2$)};
  \node[aicon/state]   at (3,5) (door_rel) {Door Estimator\\Relative ($\times 2$)};
  \node[aicon/state]   at (5,3) (goal_e)   {Goal Estimator\\($\times 2$)};
  \node[aicon/state]   at (5,5) (goal_rel) {Goal Estimator\\Relative ($\times 2$)};
    \node[aicon/state]   at (4,3) (plate_e_nonrel)  {Plate Estimator\\($\times 4$)};
  \node[aicon/state]   at (4,5) (plate_e)  {Plate Estimator\\Relative ($\times 4$)};
 
  \node[aicon/goal]    at (3,0) (gmap)     {Minimize Map\\Uncertainty Goal};
  \node[aicon/goal]    at (3,6) (gdoor)    {Door Goal};
  \node[aicon/goal]    at (5,6) (gtarget)  {Target Goal};
  \node[aicon/goal]    at (4,6) (gplate)   {Plate Goal};
 
  \begin{pgfonlayer}{bg}
    \draw[aicon/link] (dns)     -- (dpos_c);
    \draw[aicon/link] (dns)     -- (dgrid_c);
    \draw[aicon/link] (dpos_c)  -- (dpos_e);
    \draw[aicon/link] (dpos_c)  -- (omap);
    \draw[aicon/link] (dgrid_c) -- (omap);
    \draw[aicon/link] (dpos_e)  -- (drel_e);
    \draw[aicon/link] (drel_e)  -- (gmap);
    \draw[aicon/link] (omap)    -- (dtgt_c);
    \draw[aicon/link] (dtgt_c)  -- (dtgt_e);
    \draw[aicon/link] (dr_a)    -- (dact_c);
    \draw[aicon/link] (dact_c)  -- (drel_e);
    \draw[aicon/link] (dtgt_e)  -- (cross_c);
    \draw[aicon/link] (cross_c) -- (gr_rel);
    \draw[aicon/link] (gr_rel)  -- (gr_act_c);
    \draw[aicon/link] (gr_act_c)-- (gr_a);
 
 
    \draw[aicon/link] (omap)    -- (door_mc);
    \draw[aicon/link] (door_mc) -- (door_e);
    \draw[aicon/link] (door_e)  -- (door_a2r);
    \draw[aicon/link] (door_a2r)-- (door_rel);
    \draw[aicon/link] (door_rel)-- (gdoor);
    \draw[aicon/link] (door_a2r)-- (gr_rel);
 
    \draw[aicon/link] (omap)     -- (goal_mc);
    \draw[aicon/link] (goal_mc) -- (goal_e);
    \draw[aicon/link] (goal_e)  -- (goal_a2r);
    \draw[aicon/link] (goal_a2r)-- (goal_rel);
    \draw[aicon/link] (goal_rel)-- (gtarget);
    \draw[aicon/link] (goal_a2r)-- (gr_rel);
 
    \draw[aicon/link] (omap)    -- (plate_mc);
    \draw[aicon/link] (plate_mc)-- (plate_e);
    \draw[aicon/link] (plate_e) -- (gplate);
    \draw[aicon/link] (plate_e) -- (plate_a2r);
    \draw[aicon/link] (plate_a2r) -- (gr_rel);
  \end{pgfonlayer}
 
\end{tikzpicture}
\caption{AICON graph for the pressure plate task, used by the RL, LD and AICON policies.}
\label{fig:aicon_graph_pp}
\end{figure}

The pressure plate graph instantiates the compositionality claim directly: it reuses the cooperative search modules unchanged---drone self-localization, occupancy mapping, cross-robot target estimation, and Abs2Rel framing---and extends the graph with task-specific components for the plates, door, and final goal. No existing component is modified; the new regularities are simply added as additional nodes and edges, consistent with the structural compositionality described in Section~\ref{sec:instantiation}. The AICON graph used for the RL, LD and AICON policies of Section \ref{sec:results} is represented in Fig.~\ref{fig:aicon_graph_pp}. The additional task-specific components for the plates, door, and final goal are:

\subsubsection{Task-specific sensors:}
\begin{description}
  \item[\textbf{Drone Sensor}.]
        Composite drone-side sensor that exposes plate-, door- and
        goal-cell observations alongside drone position and the
        occupancy grid.
\end{description}

\subsubsection{Task-specific estimators:}
\begin{description}
  \item[\textbf{Goal Estimator}.]
        Absolute position of the final goal cell.
  \item[\textbf{Door Estimator}.]
        Absolute position of the door that opens when both plates are
        pressed.
  \item[\textbf{Ground Robot Plate Estimator (Relative)}.]
        Each robot's local-frame estimate of its assigned pressure
        plate.
  \item[\textbf{Goal Estimator (Relative)}.]
        Per-robot relative-frame estimate of the goal cell.
\end{description}

\subsubsection{Task-specific interconnections:}
\begin{description}
  \item[\textbf{Door / Goal Measurement Connection}.]
        Feed drone-side observations of the door and goal to their
        respective estimators.
  \item[\textbf{Door / Goal Abs2Rel Connection}.]
        Change of frame for door and goal estimates from world frame to
        each ground robot's local frame.
  \item[\textbf{Drone Door Connection / Drone Goal Connection}.]
        Bind the drone-side door- and goal-observation channels into the
        Drone Sensor and corresponding estimators.
\end{description}

\subsubsection{Goals:}
\begin{description}
  \item[\textbf{Target Goal}.]
        \textsl{Fixed-distance to plate} cost for the ground robot
        assigned to each pressure plate.
  \item[\textbf{Plate Goal}.]
        Plate-specific approach goal (active while the plate is still
        unpressed).
  \item[\textbf{Door Goal}.]
        Approach the door once the plates have been pressed and the
        door has opened.
  \item[\textbf{Minimize Map Uncertainty Goal}.]
        Same exploration prior as in cooperative search; remains active
        throughout the task so the drone keeps charting the map.
\end{description}

\section{Further experimental results}\label{app:experiments}
This appendix presents the real-robot deployments summarized in \textbf{Q5}. In all panels the quadrotor and ground-robot trajectories are overlaid on the arena and colored by time, so that coordination and re-targeting are visible directly. The learned task factor is identical to simulation; only the sensor drivers in the world factor modeled with AICON are swapped for their physical counterparts, and no policy is retrained.

\begin{figure}
    \centering
    \begin{tabular}{cc}
         \includegraphics[width=0.4\linewidth]{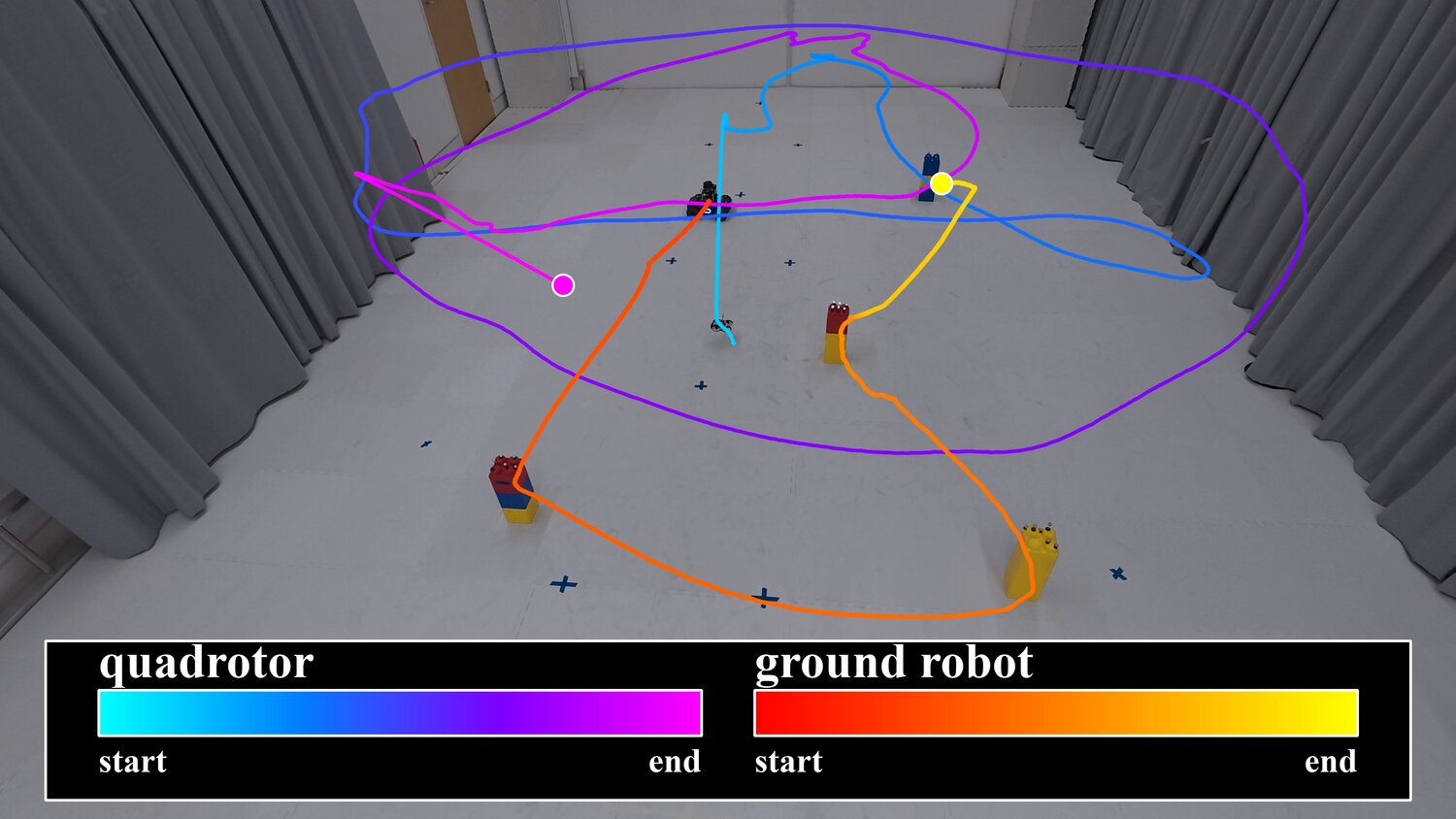}
         &  
         \includegraphics[width=0.4\linewidth]{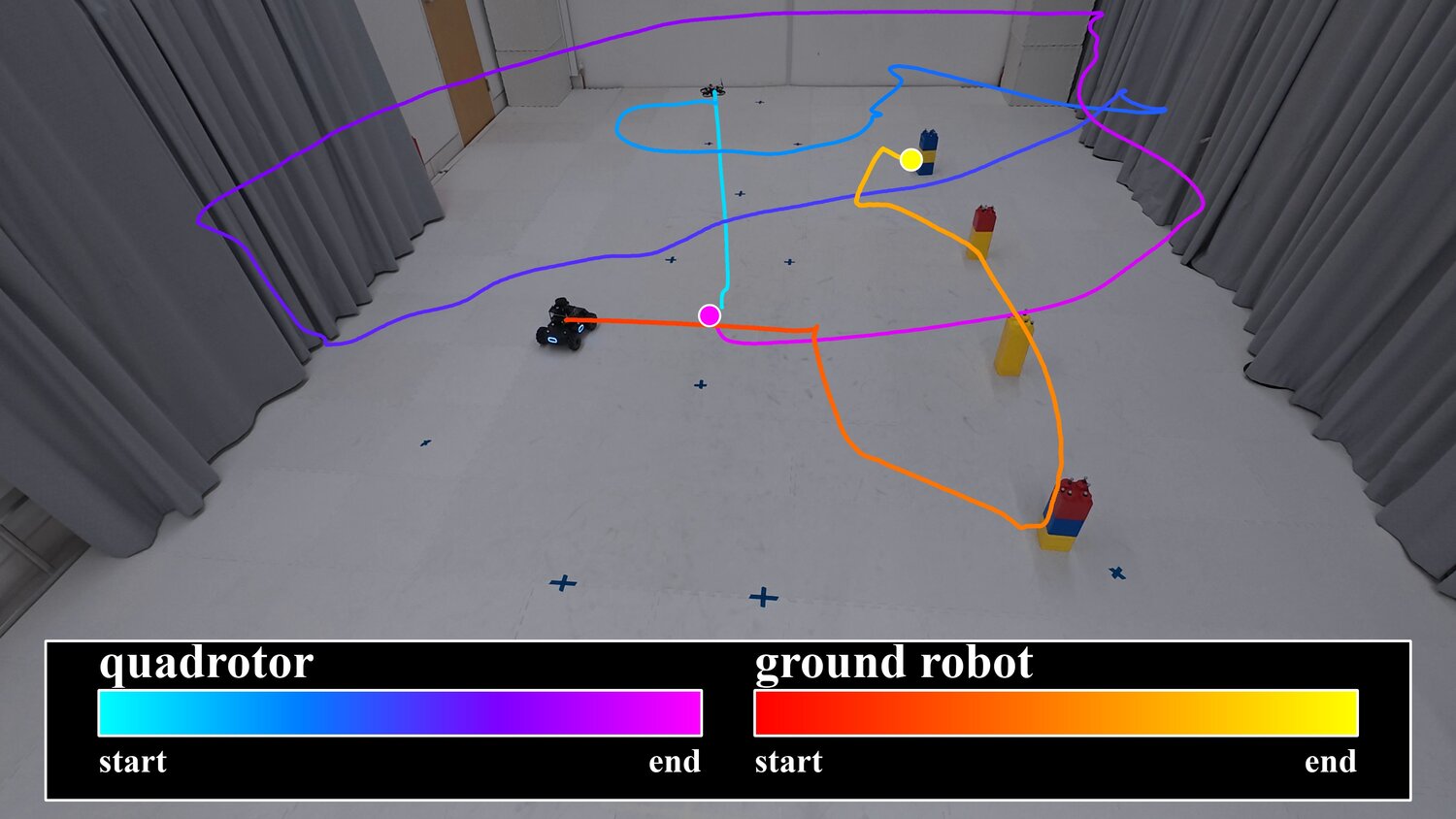}
         \\
         \includegraphics[width=0.4\linewidth]{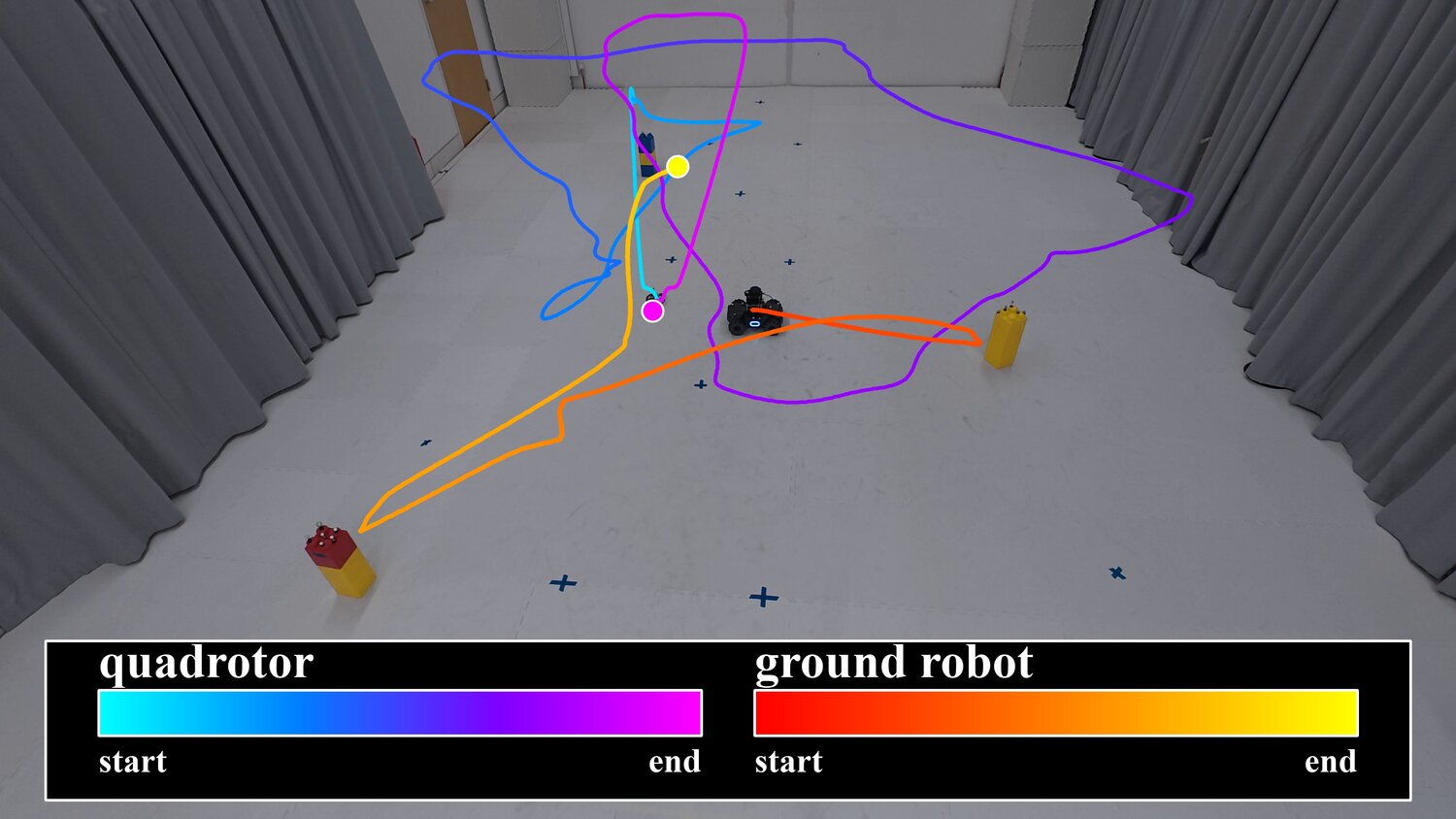}
         &  
         \includegraphics[width=0.4\linewidth]{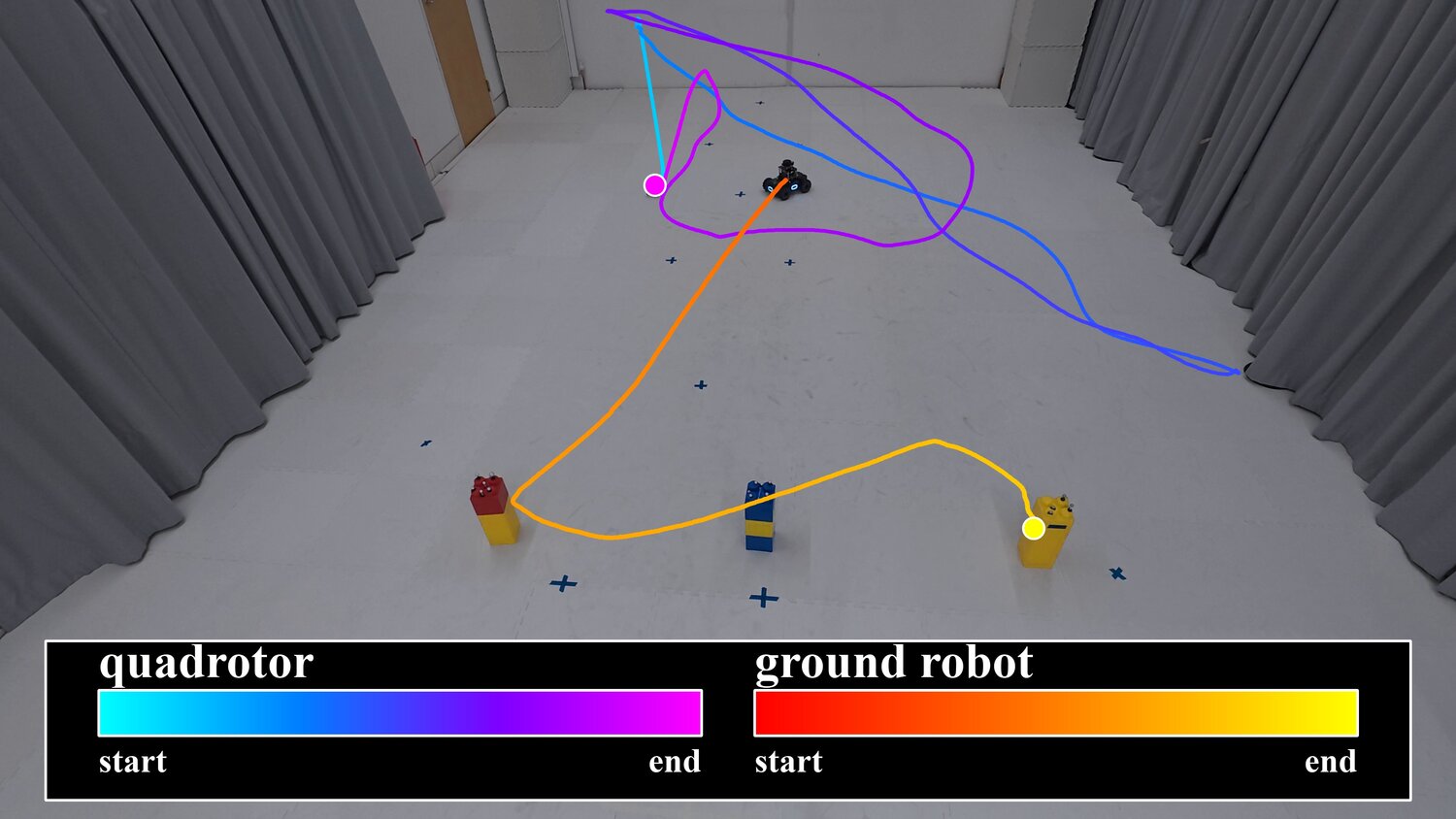}
         \\
         \includegraphics[width=0.4\linewidth]{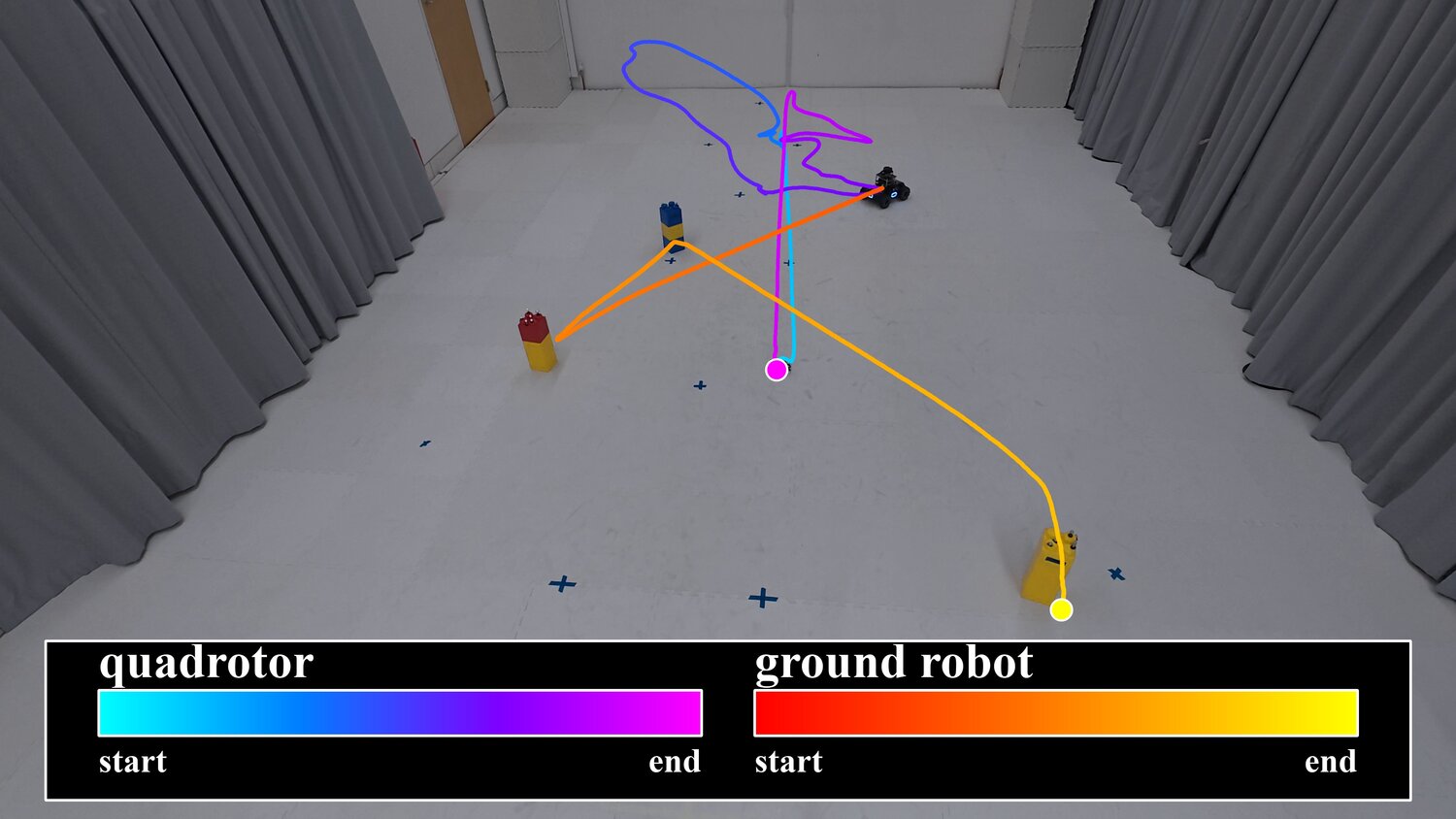}
         &  
         \includegraphics[width=0.4\linewidth]{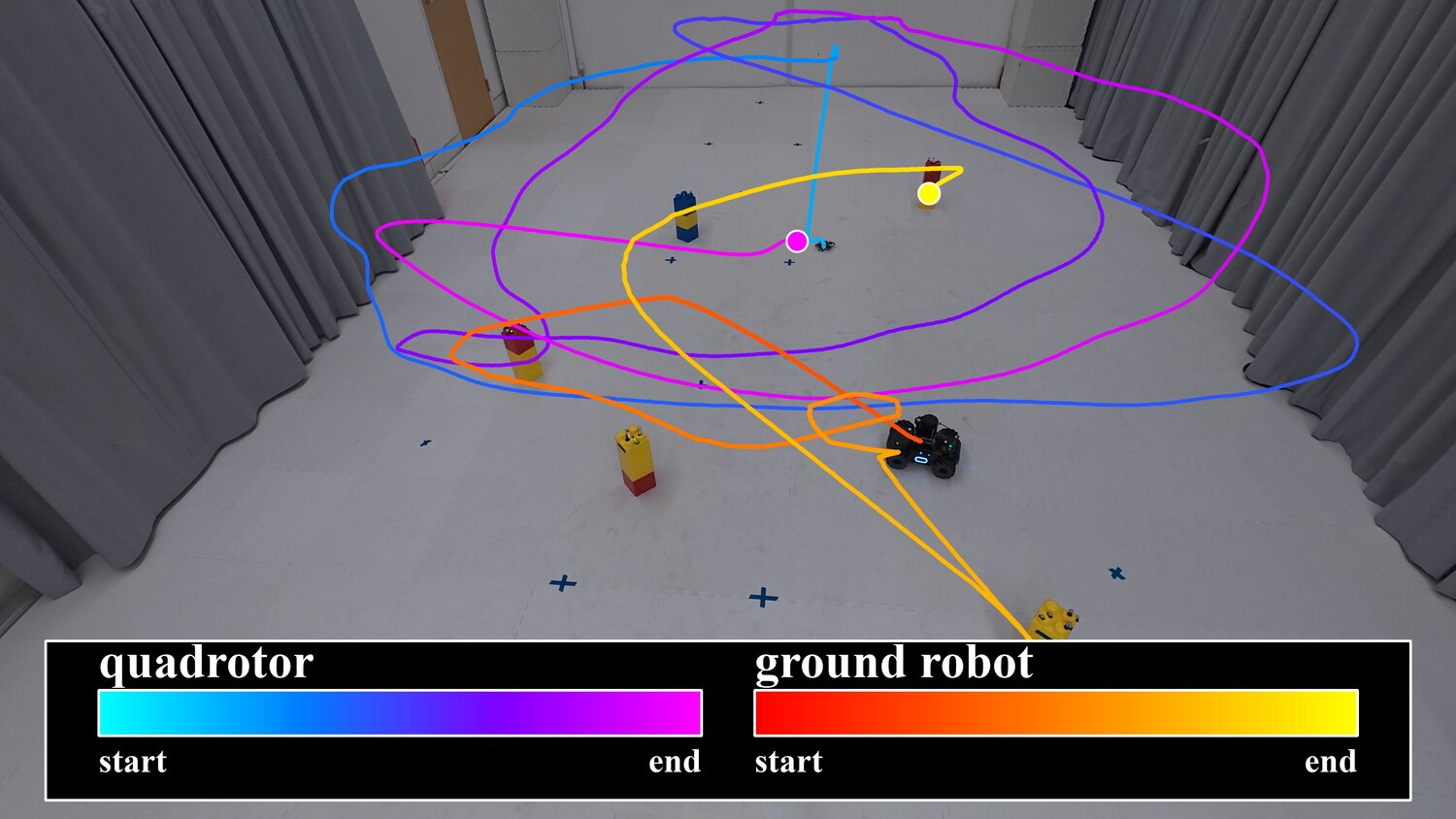}
         \\
         \includegraphics[width=0.4\linewidth]{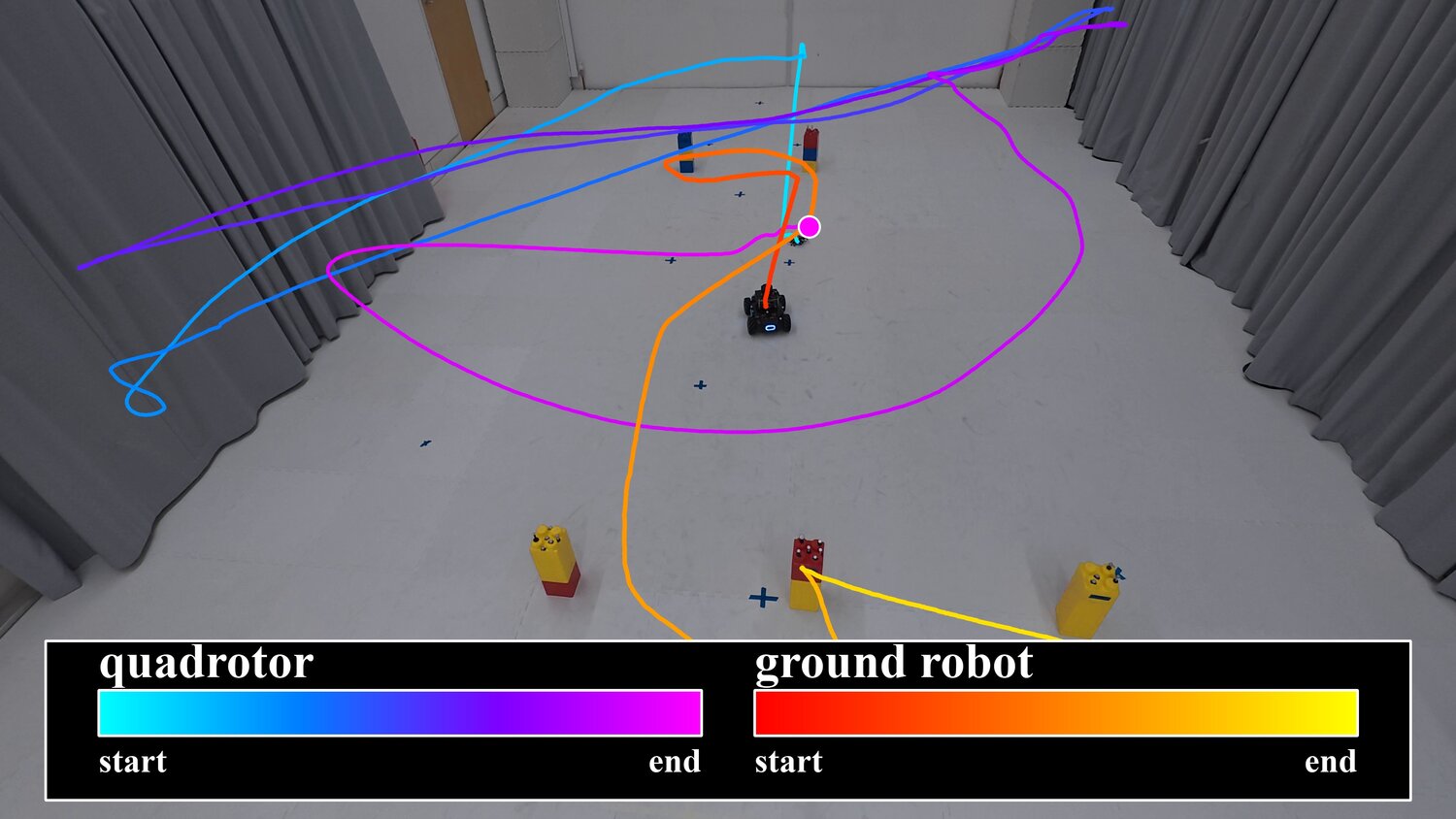}
         &  
         \includegraphics[width=0.4\linewidth]{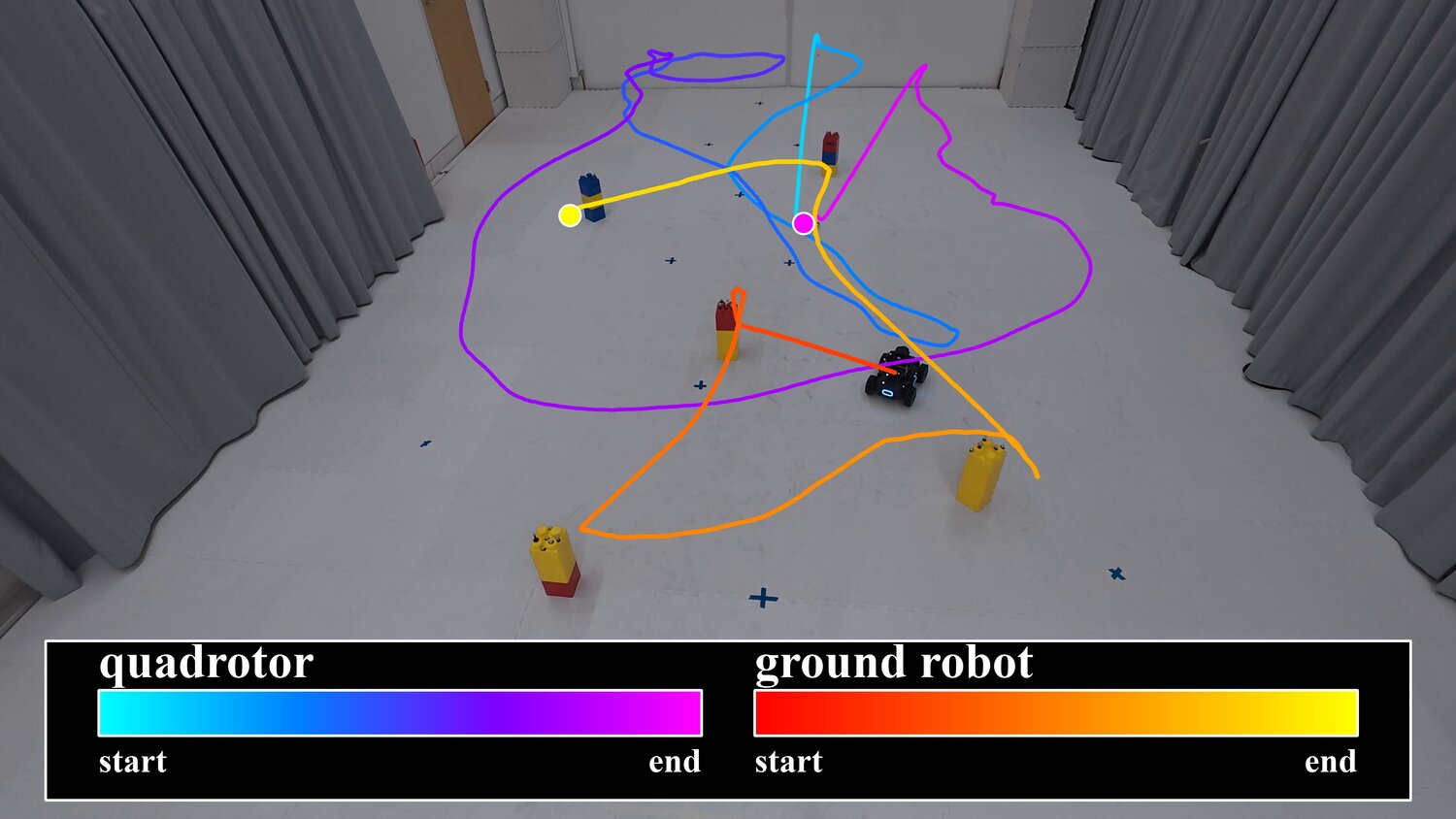} 
    \end{tabular}
    \caption{Real-robot search deployments, RL policy (8/8 success). Trajectories colored by time.}
    \label{fig:exps_cs_rl}
\end{figure}

\begin{figure}
    \centering
    \begin{tabular}{cc}
         \includegraphics[width=0.4\linewidth]{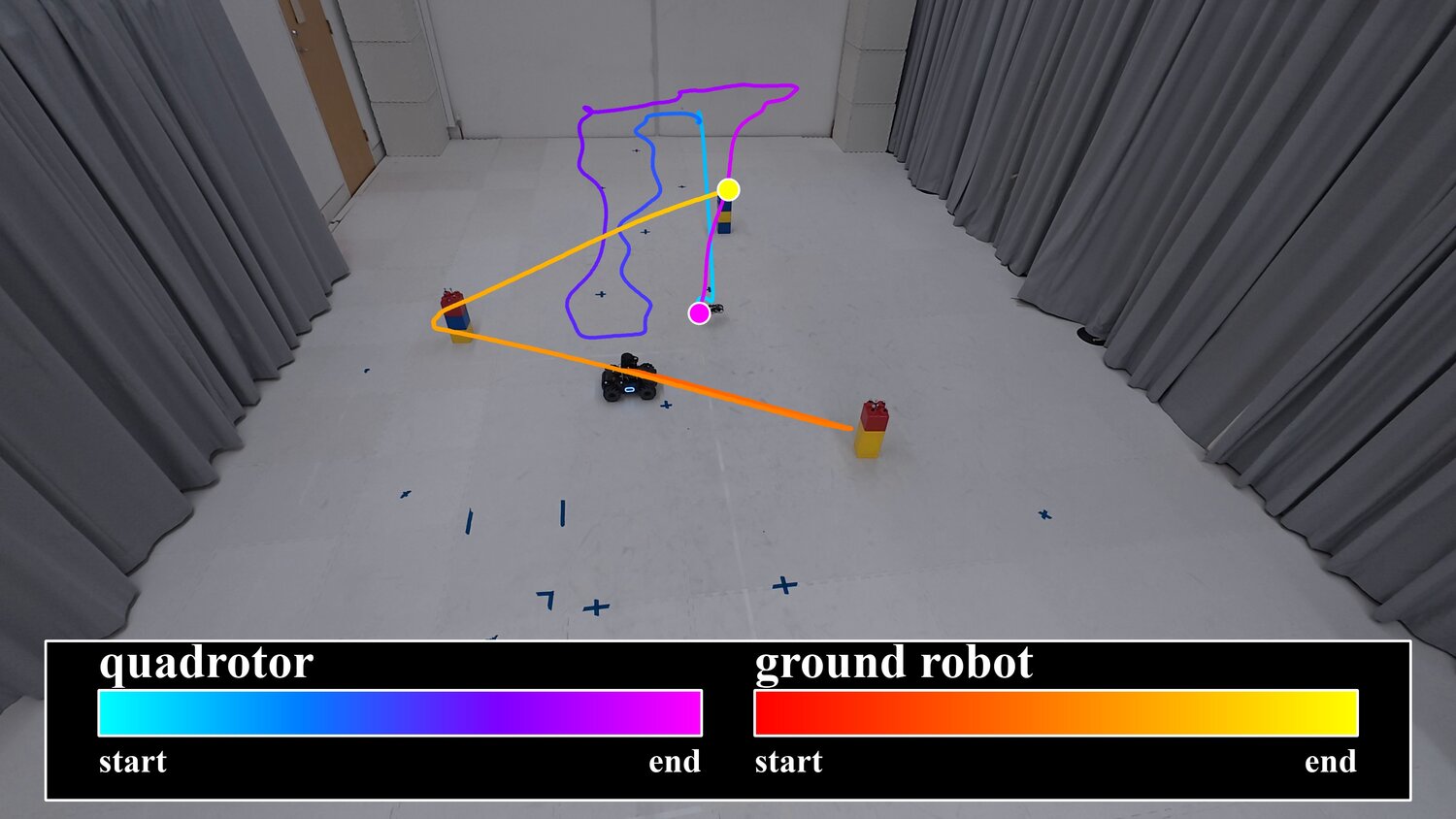}
         &  
         \includegraphics[width=0.4\linewidth]{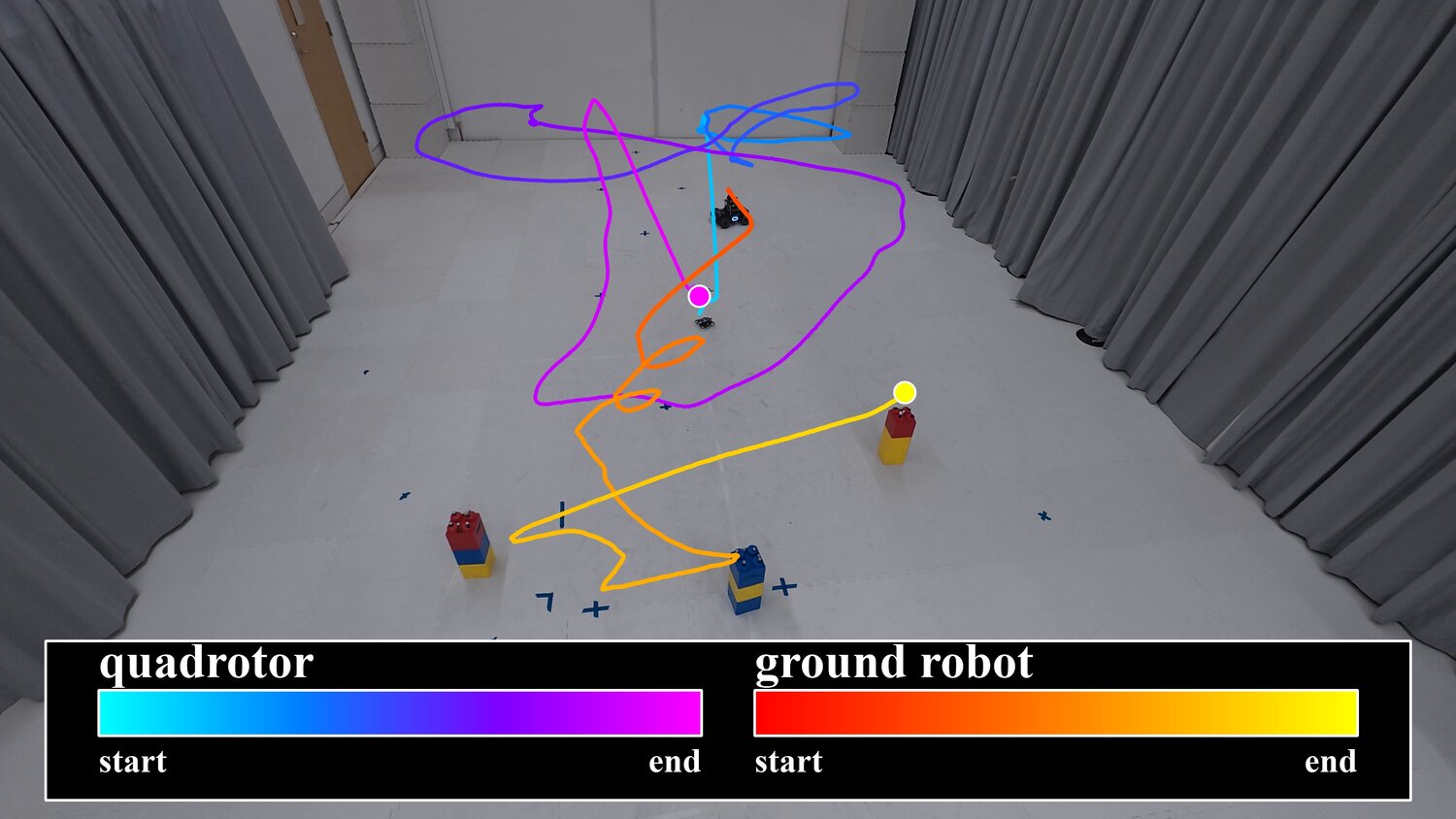}
         \\
         \includegraphics[width=0.4\linewidth]{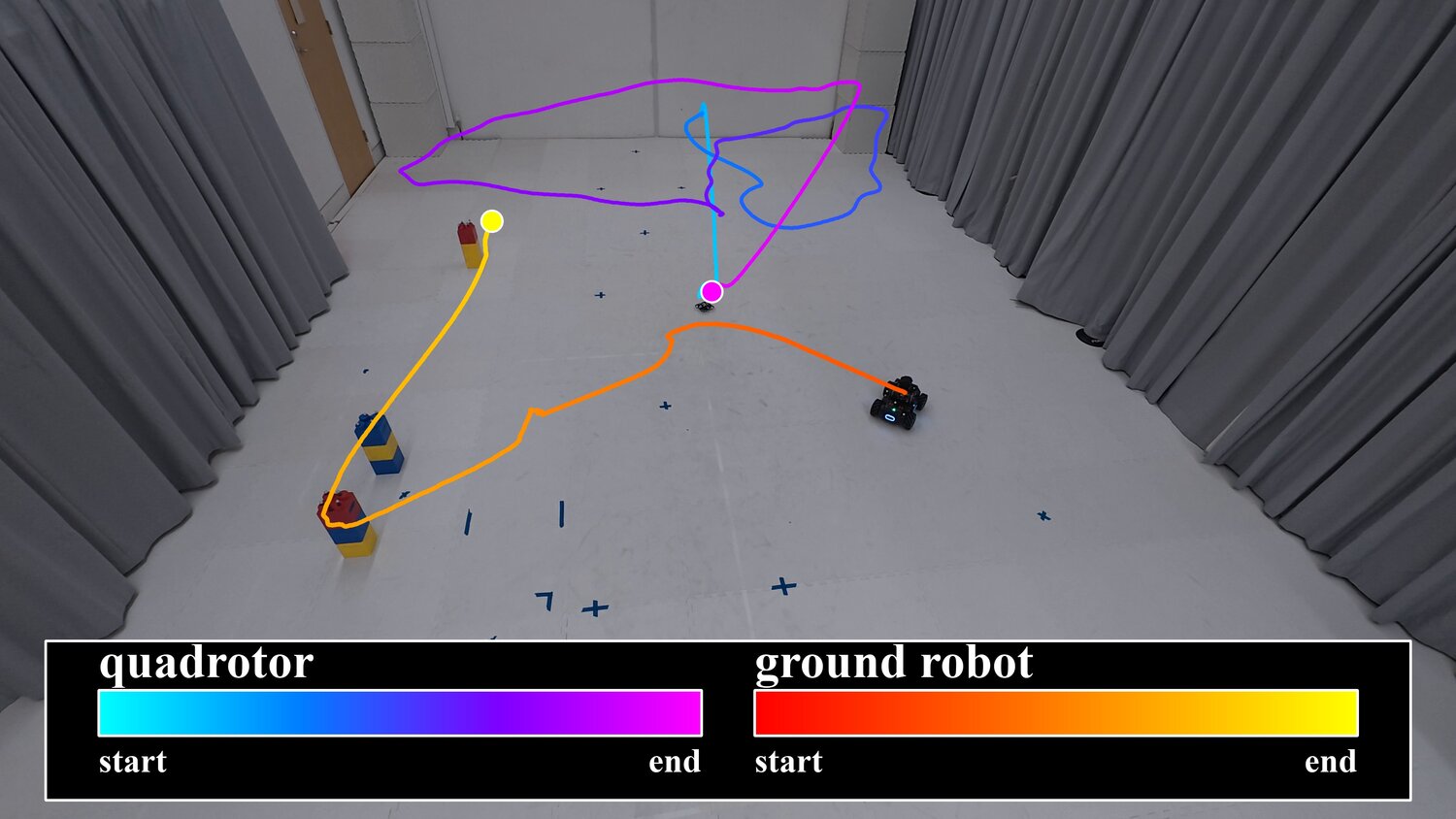}
         &  
         \includegraphics[width=0.4\linewidth]{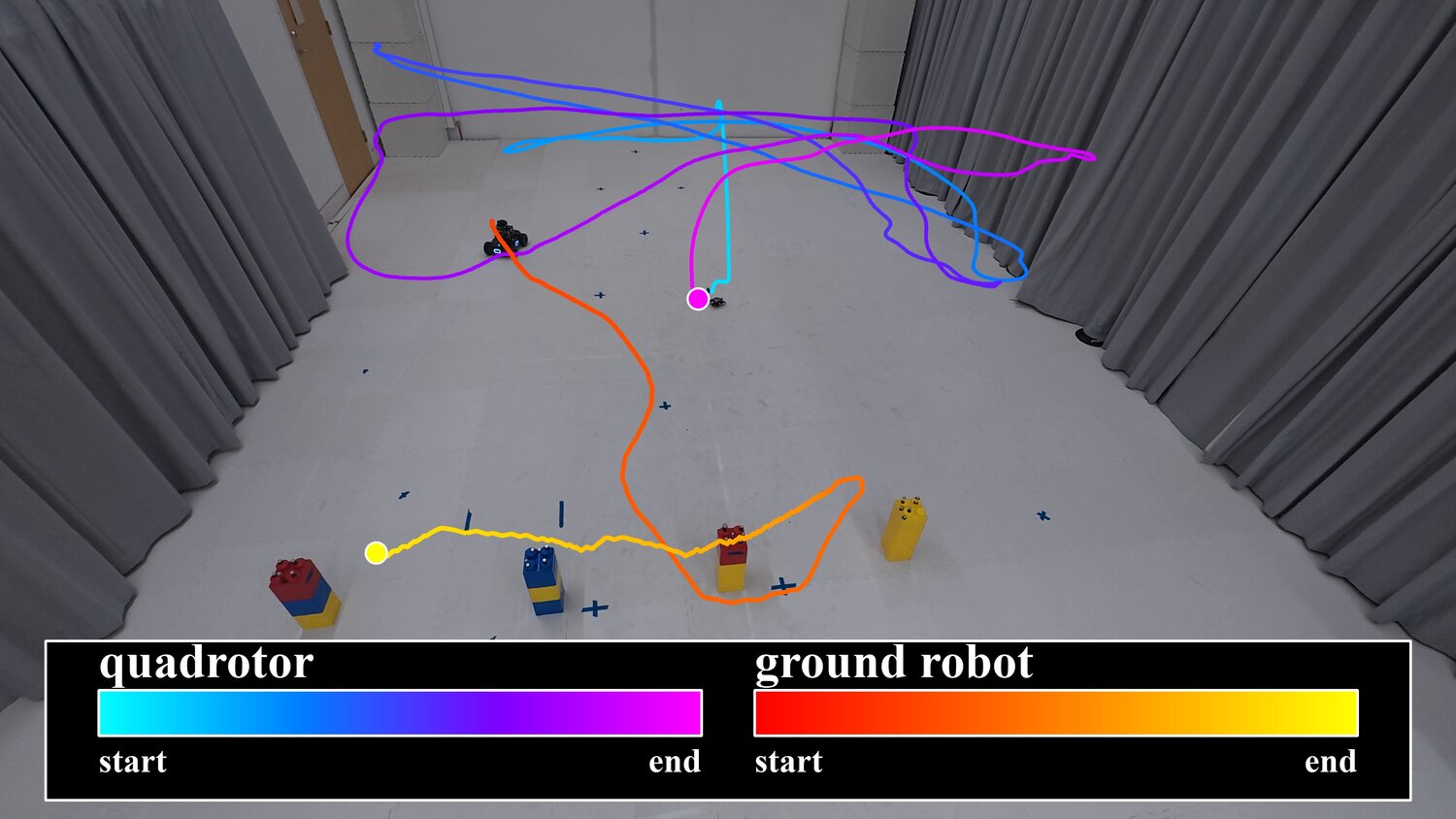}
         \\
         \includegraphics[width=0.4\linewidth]{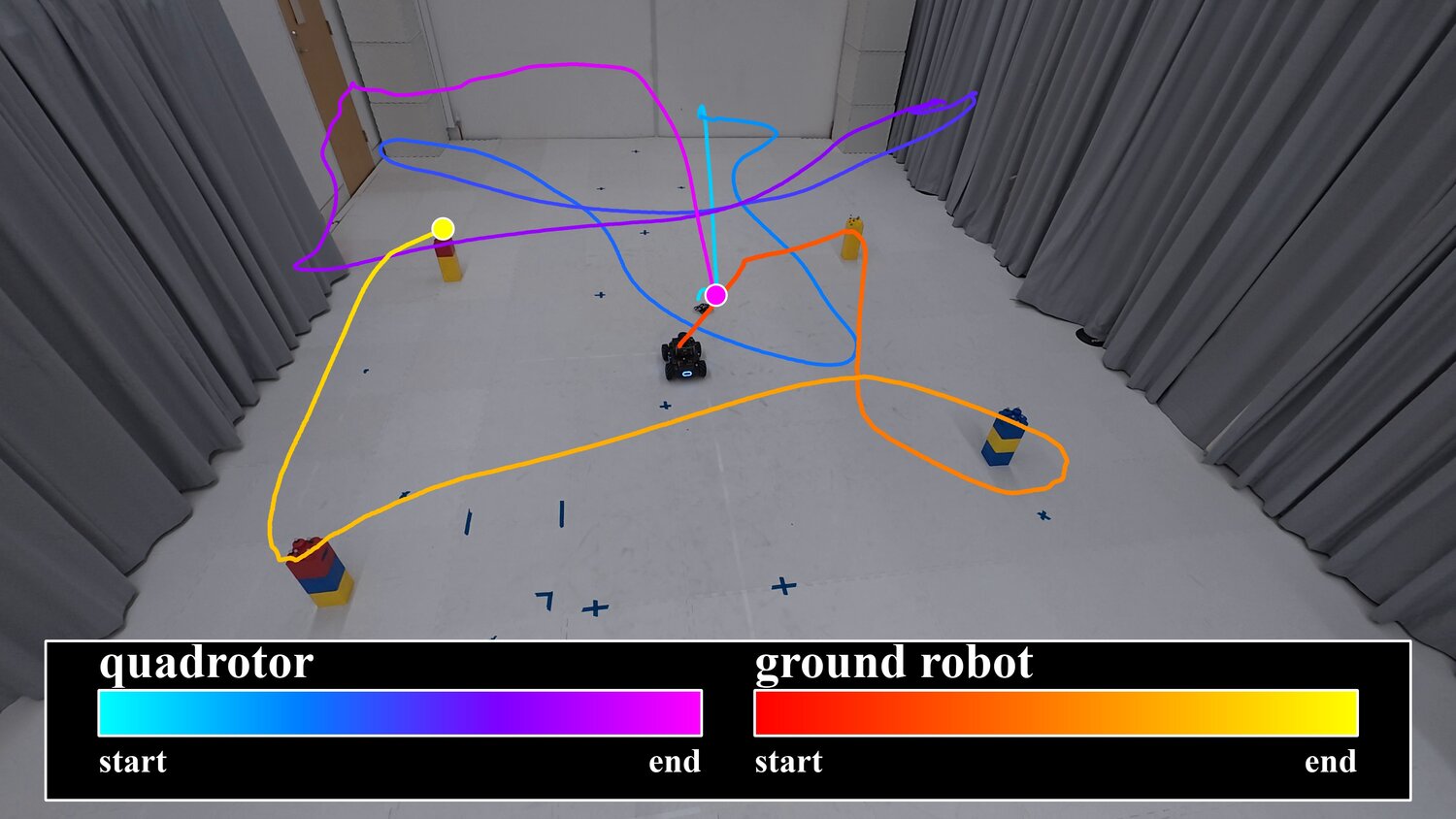}
         &  
         \includegraphics[width=0.4\linewidth]{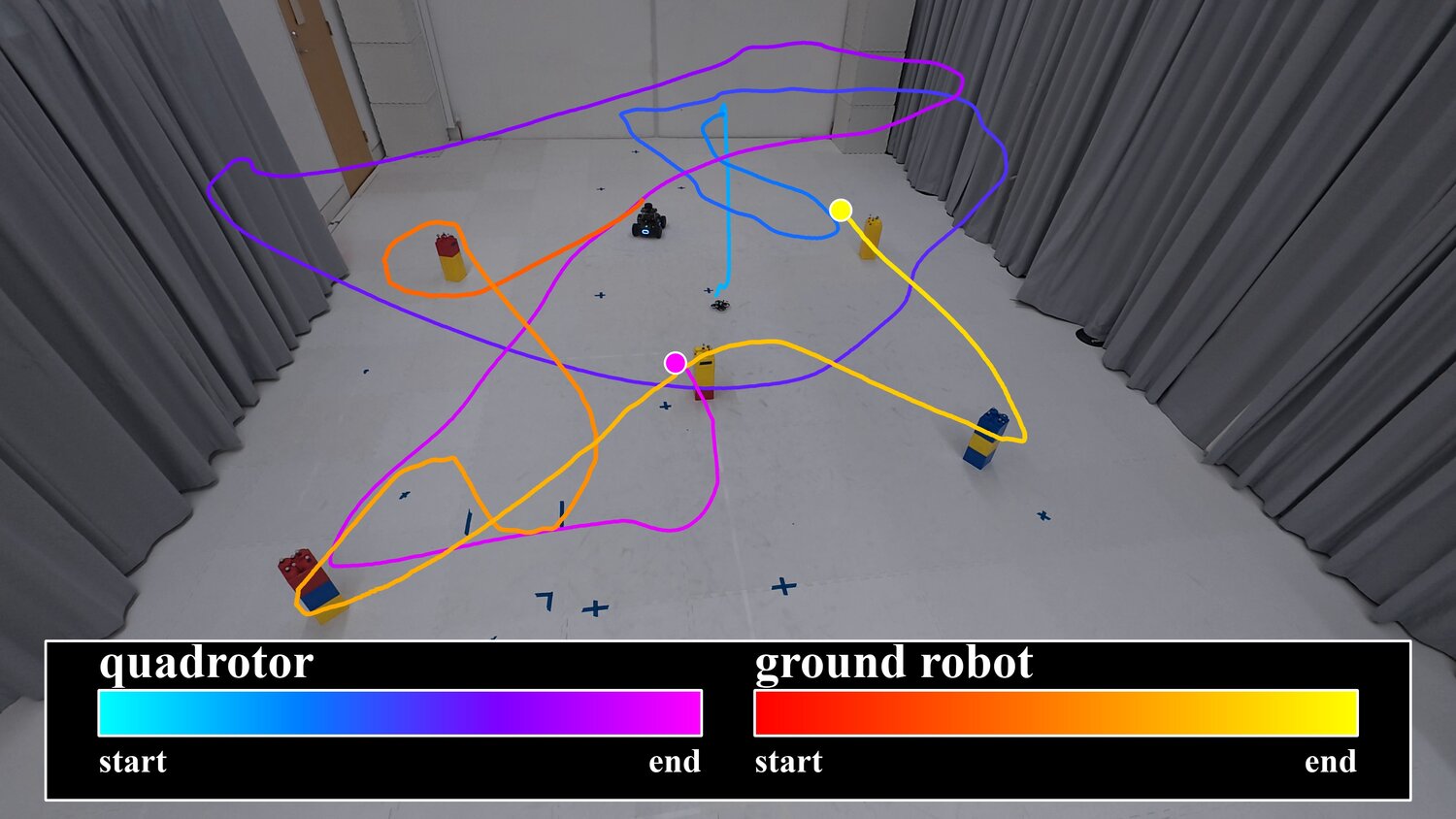}
         \\
         \includegraphics[width=0.4\linewidth]{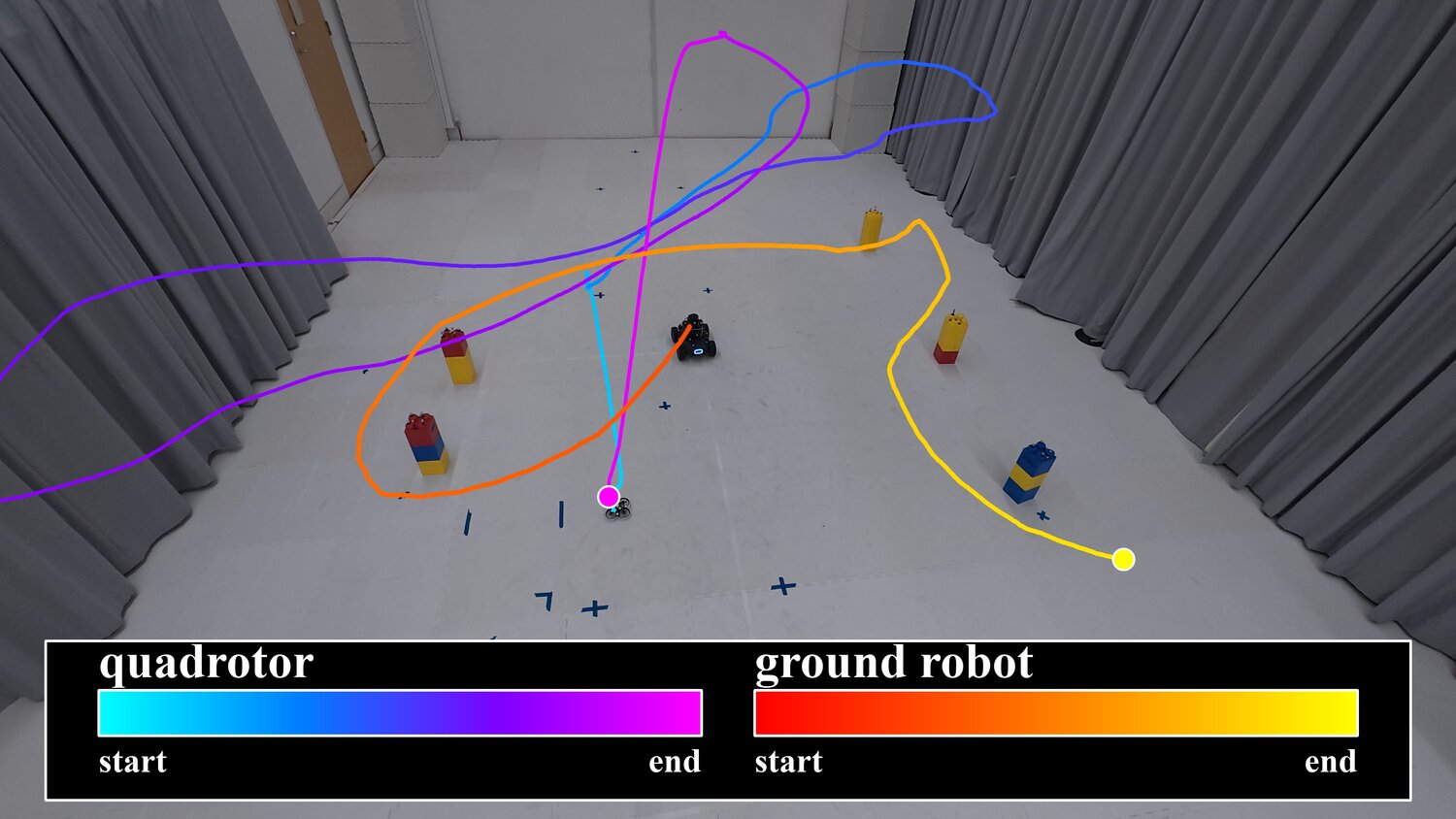}
         &  
         \includegraphics[width=0.4\linewidth]{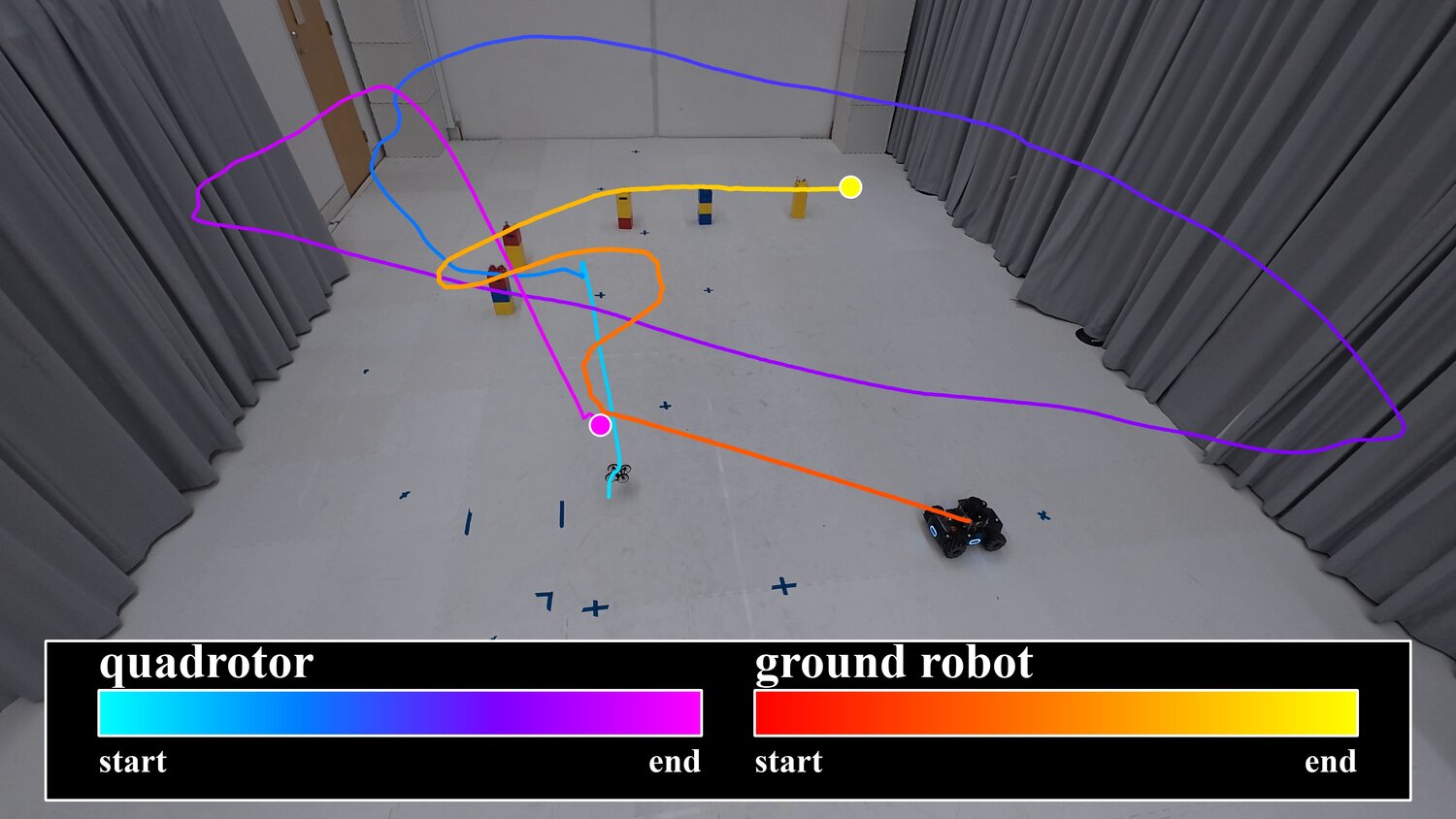} 
    \end{tabular}
    \caption{Real-robot search deployments, LD policy (8/8 success).}
    \label{fig:exps_cs_ld}
\end{figure}

\textbf{Search (Figs. \ref{fig:exps_cs_rl}-\ref{fig:exps_cs_ld})}. We ran 8 deployments of the RL policy (Fig.~\ref{fig:exps_cs_rl}) and 8 of the LD policy (Fig.~\ref{fig:exps_cs_ld}), each with a real quadrotor under true flight dynamics and localization drift, observation noise that reshapes the exploration pattern, a LiDAR of different range than in simulation, and a variable number of targets. Both policies achieve a 100\% success rate (8/8 each): every target is located by the quadrotor's evolving occupancy map and retrieved by the ground robot. The trajectories show the same reactive re-targeting observed in simulation—the ground robot's goal shifts as new targets enter the map.

\begin{figure}
    \centering
         \includegraphics[width=0.8\linewidth]{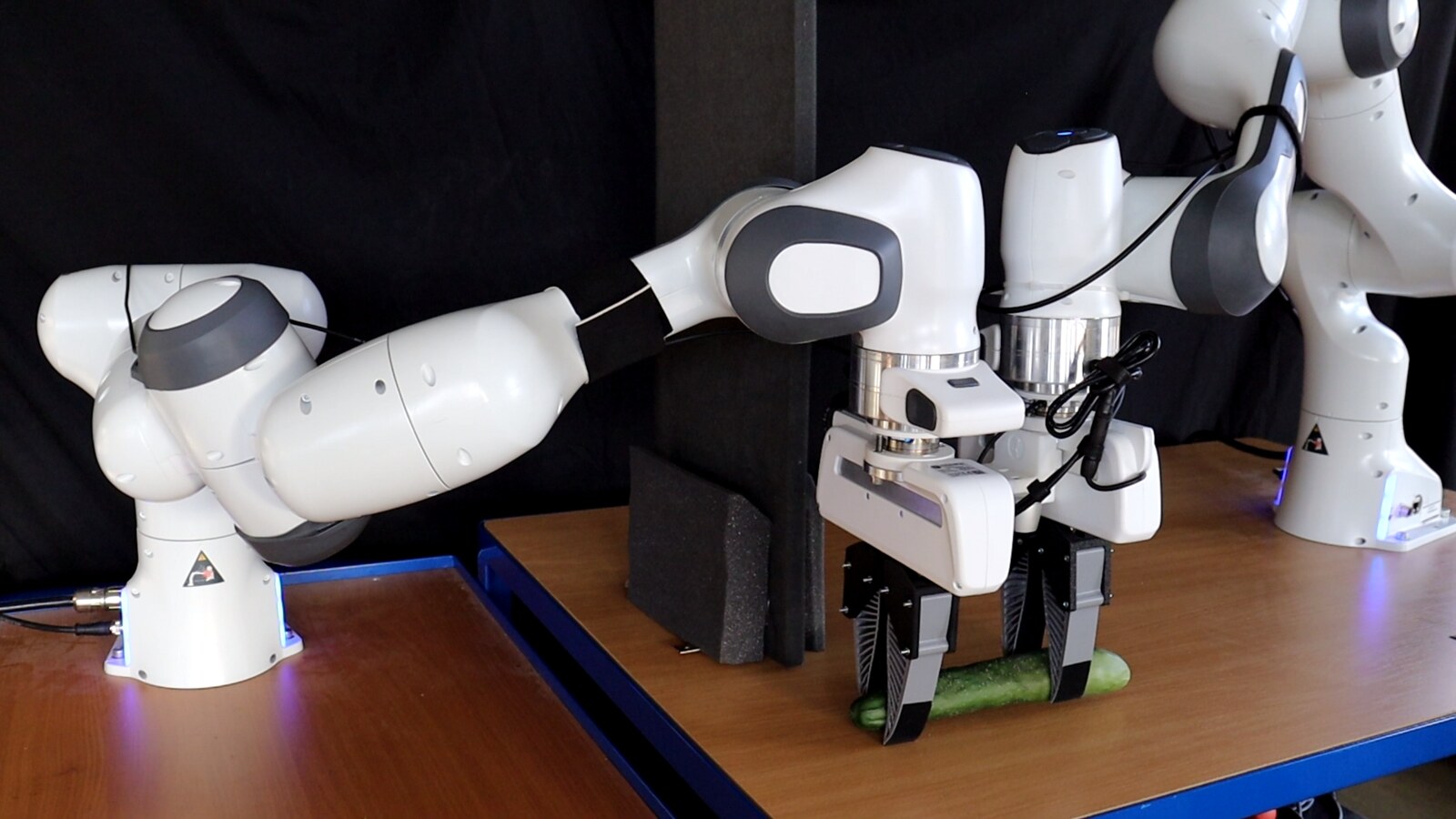}
    \caption{Successful handover in the real-world experiments.}
    \label{fig:exp_handover_real}
\end{figure}

\textbf{Handover.} Both bimanual policy RL and LD succeed in 9/10 real deployments (90\%). The real setup uses two live Franka Panda arms; simulated end-effector, object, and obstacle are replaced by calibrated robot-state and scene measurements, and commands are sent as Cartesian pose targets with binary gripper commands. Success holds even though the object's shape and weight and the arms' placement and orientation are out of distribution. Figure~\ref{fig:exp_handover_real} illustrates a representative example of a successful real-world handover. The only failure arises when the commanded motion exceeds the arm's unmodeled actuator limits, a physically unrealizable regime. \\
Details about OOD and scaling experiments are listed in Table~\ref{tab:handover_eval_conditions}.

\begin{figure}
    \centering
    \begin{tabular}{cc}
         \includegraphics[width=0.4\linewidth]{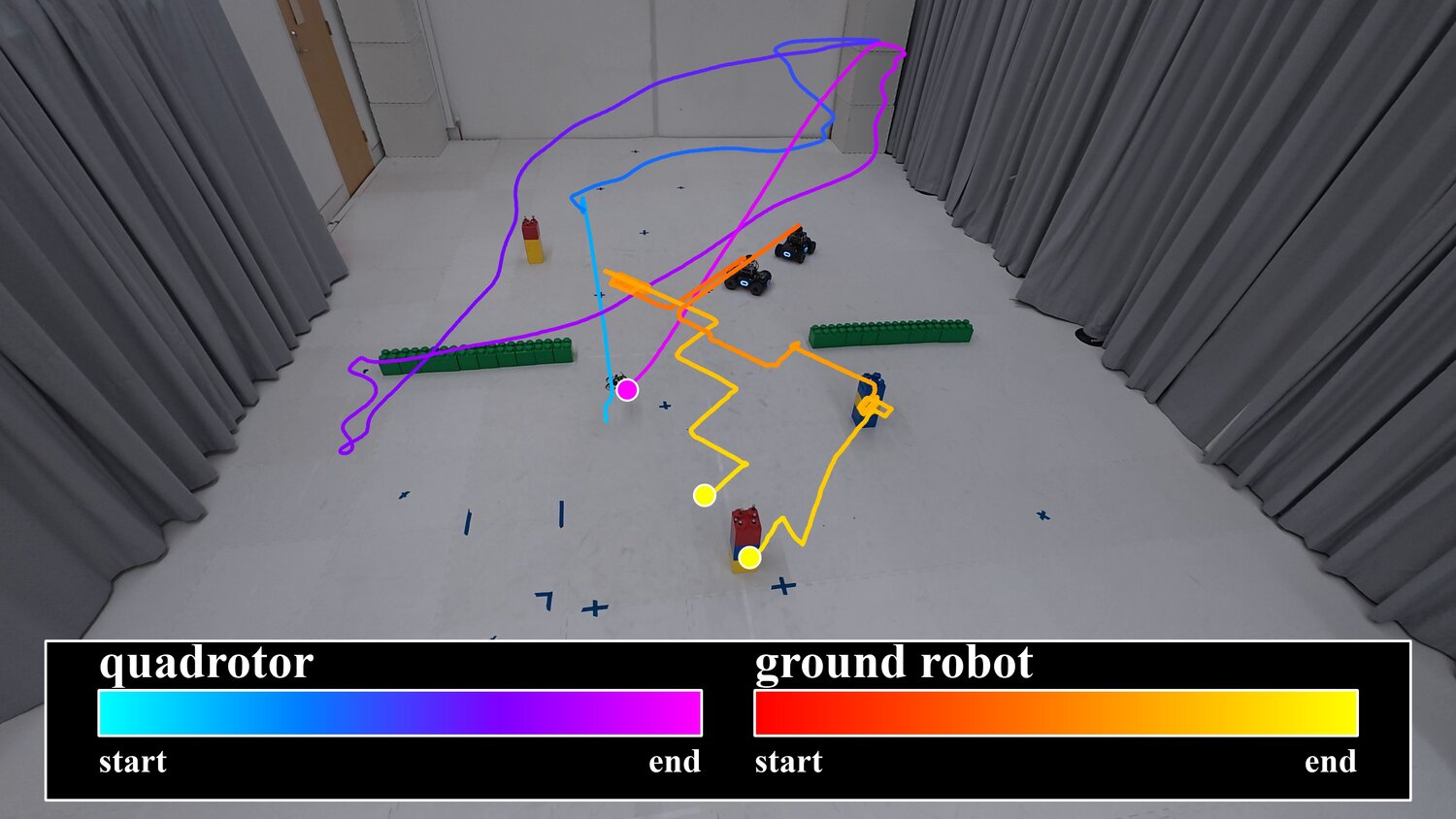}
         &  
         \includegraphics[width=0.4\linewidth]{exp_pp_rl_2_trajectory.jpg}
         \\
         \includegraphics[width=0.4\linewidth]{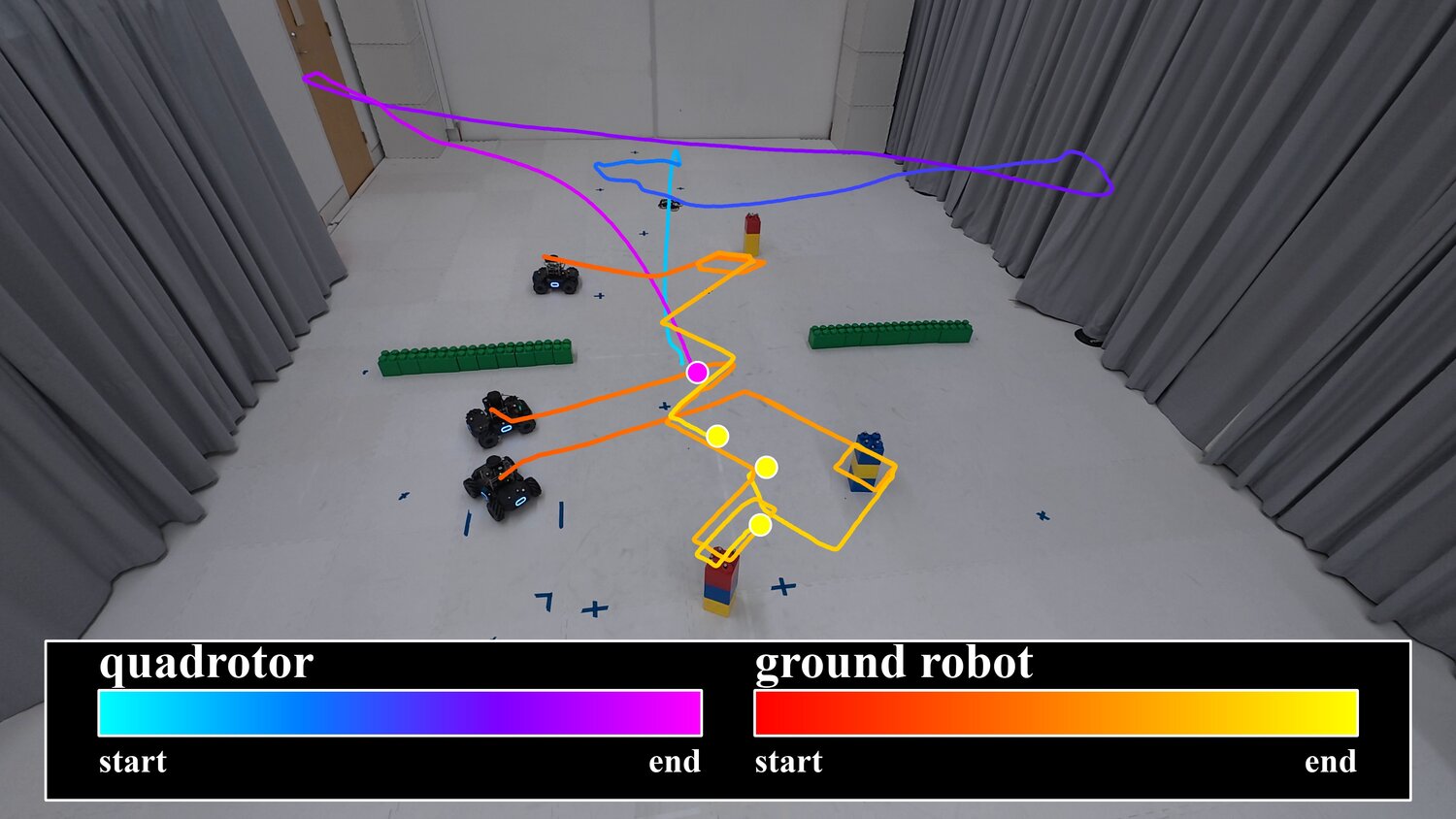}
         &  
         \includegraphics[width=0.4\linewidth]{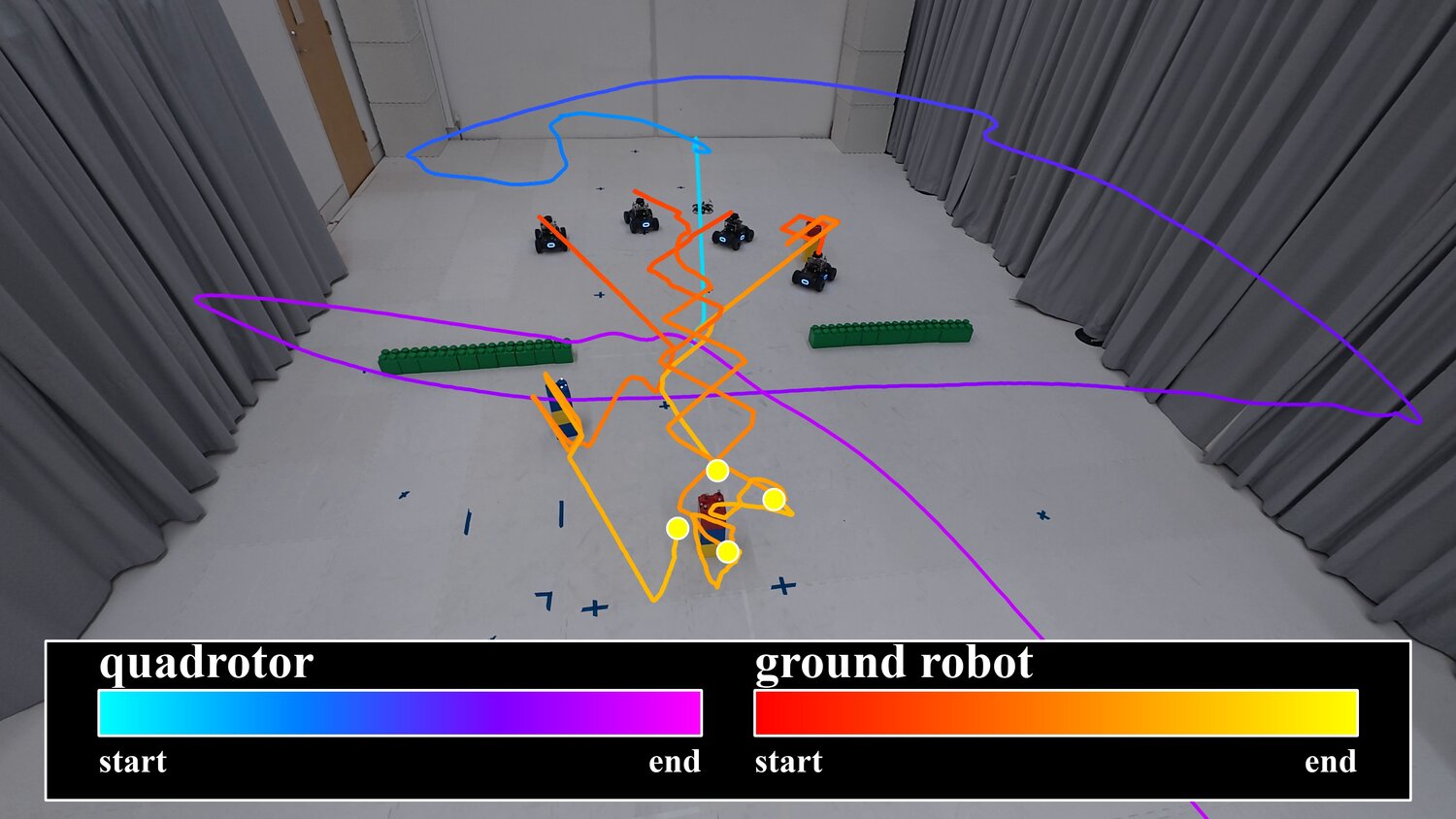}
    \end{tabular}
    \caption{Real-robot pressure-plate deployments, RL policy (4/4 success).}
    \label{fig:exps_pp_rl}
\end{figure}

\begin{figure}
    \centering
    \begin{tabular}{cc}
         \includegraphics[width=0.4\linewidth]{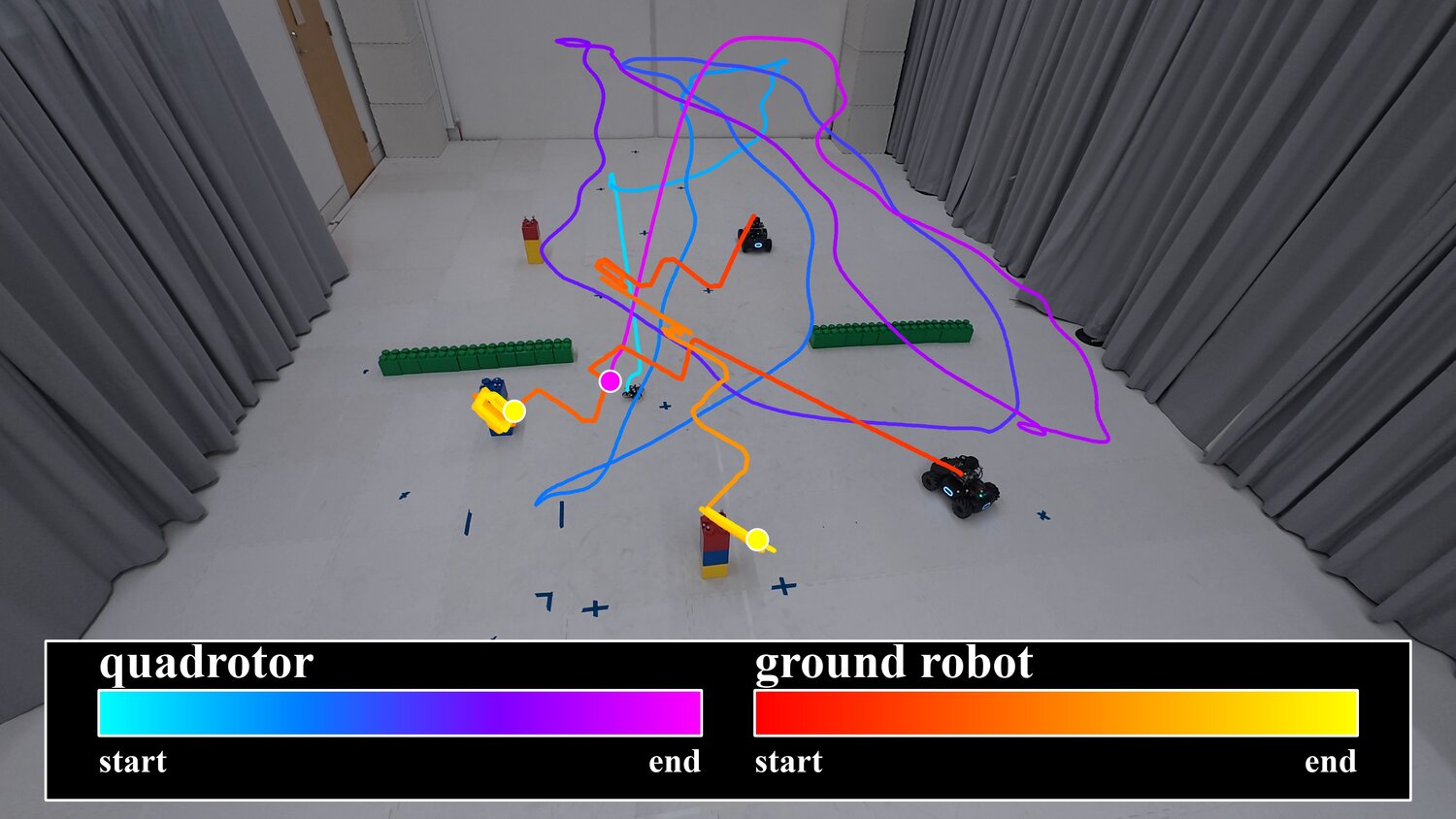}
         &  
         \includegraphics[width=0.4\linewidth]{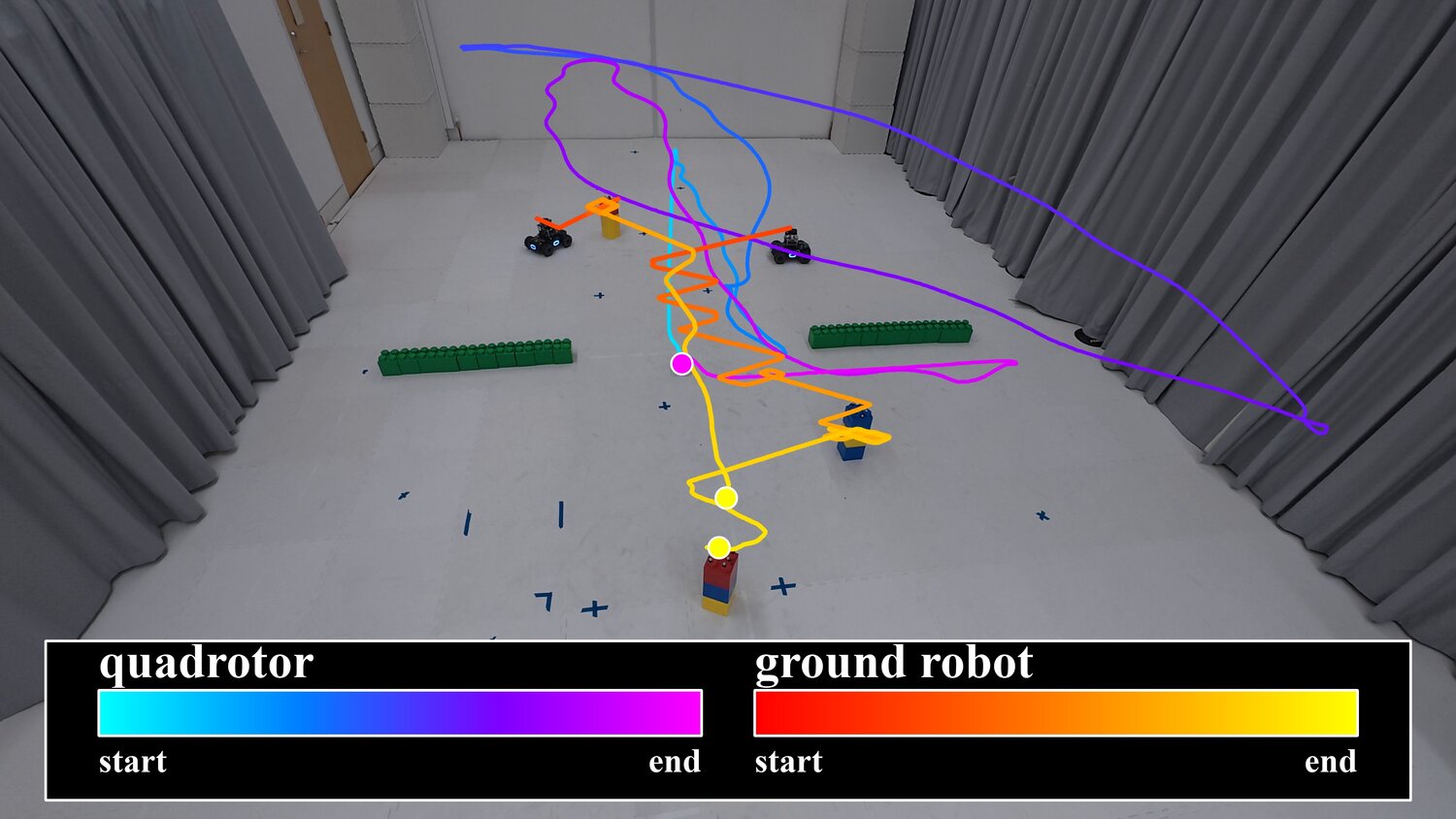}
         \\
         \includegraphics[width=0.4\linewidth]{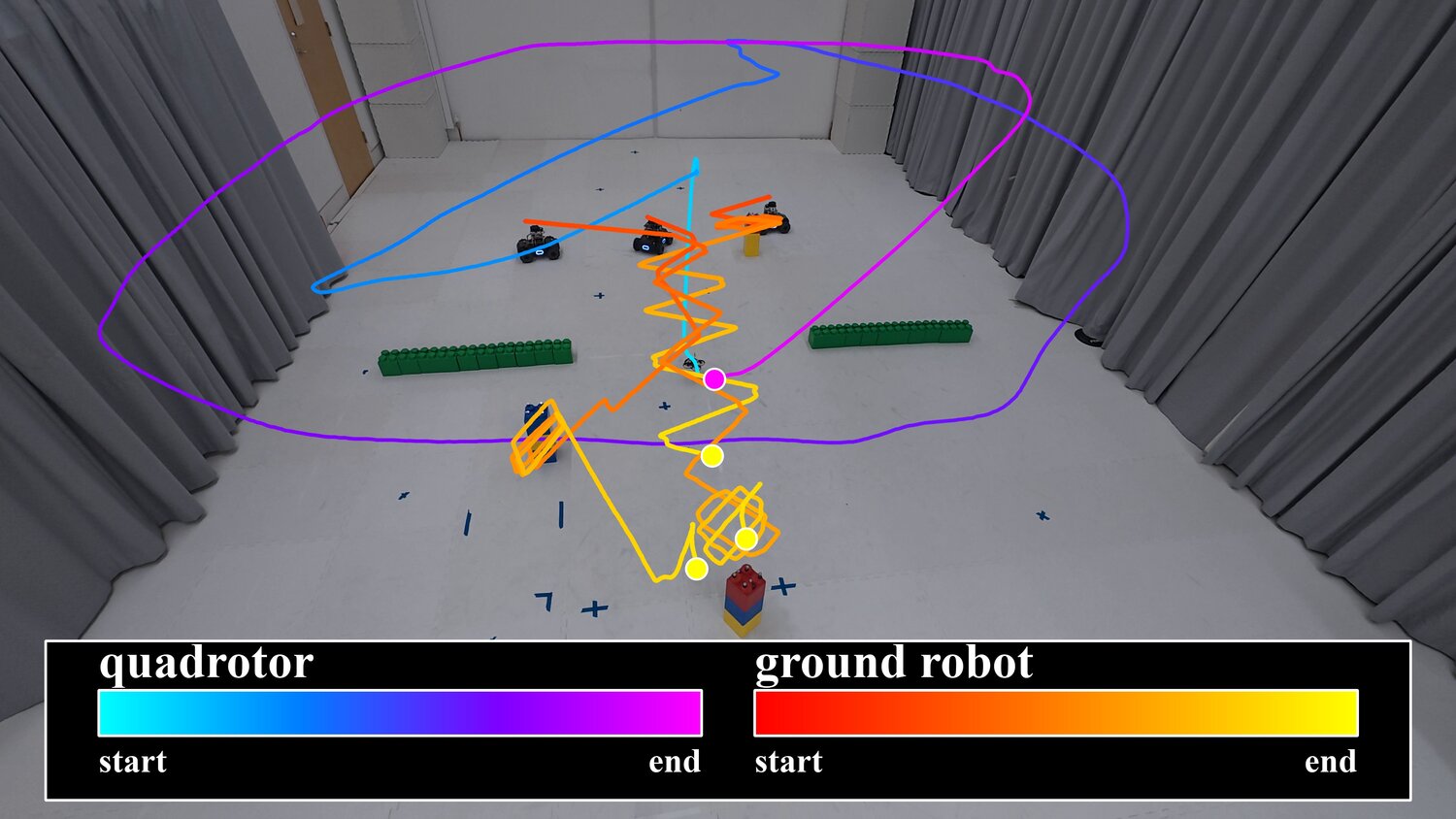}
         &  
         \includegraphics[width=0.4\linewidth]{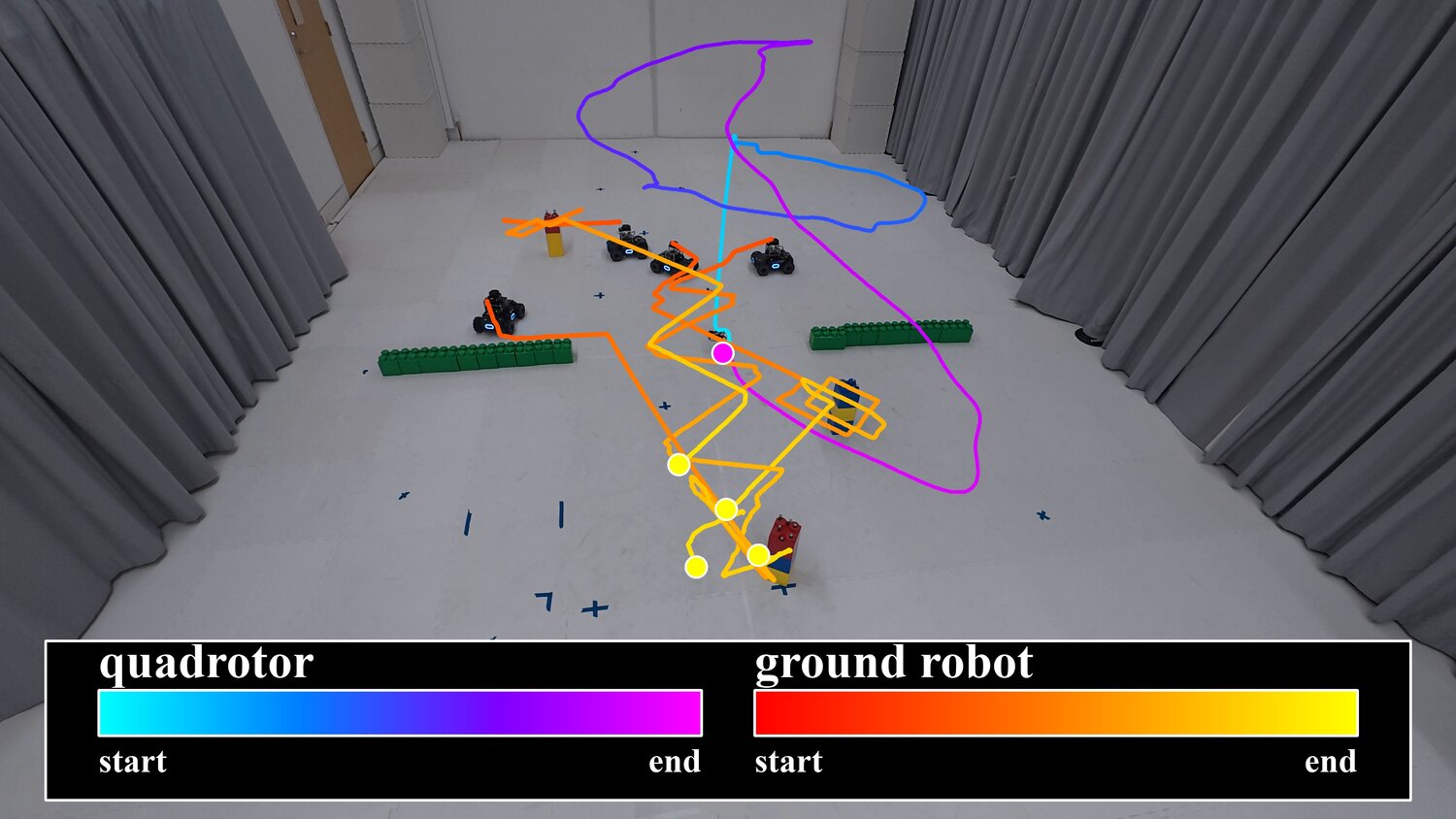}
    \end{tabular}
    \caption{Real-robot pressure-plate deployments, LD policy (3/4 success, failed one on the top right panel); the failed run shows a stalled stage transition, matching the simulation failure mode (Fig.~\ref{fig:main_performance}b).}
    \label{fig:exps_pp_ld}
\end{figure}

\textbf{Pressure plate (Figs. \ref{fig:exps_pp_rl}–\ref{fig:exps_pp_ld})}. We ran 4 deployments of RL (Fig.~\ref{fig:exps_pp_rl}) and 4 of LD (Fig.~\ref{fig:exps_pp_ld}), under the same quadrotor and LiDAR changes and a variable number of ground robots. RL succeeds in 4/4 (100\%) and LD in 3/4 (75\%). The single LD failure is a failed transition between task stages---the same behavior seen in simulation (Fig.~\ref{fig:main_performance}b)---and is not caused by the reality gap; the world estimates remain accurate throughout. Besides, this happens in a setting with only two ground robots, which is out of the training distribution. Successful runs show robots implicitly differentiating roles, holding plates and crossing in sequence to open the door and reach the goal.

\end{document}